\documentclass[runningheads]{llncs}
\usepackage{graphicx}

\usepackage{tikz}
\usepackage{comment} 
\usepackage{amsmath,amssymb} 
\usepackage{color}

\usepackage{makecell}
\usepackage{subfigure}
\usepackage{multirow}
\usepackage{array}
\usepackage{listings}

\usepackage{epsfig}
\usepackage{graphicx}
\usepackage{amsmath}
\usepackage{amssymb}

\usepackage{cases}
\usepackage[linesnumbered,ruled,vlined,lined,boxed,commentsnumbered]{algorithm2e}
\usepackage{indentfirst}
\usepackage{boldline}
\usepackage{enumitem}
\setitemize{noitemsep,topsep=0pt,parsep=0pt,partopsep=0pt}

\usepackage{longtable}
\usepackage{float}

\usepackage[pagebackref=true,breaklinks=true,letterpaper=true,colorlinks,bookmarks=false]{hyperref}
\usepackage{caption}
\usepackage{xspace}
\makeatletter
\DeclareRobustCommand\onedot{\futurelet\@let@token\@onedot}
\def\@onedot{\ifx\@let@token.\else.\null\fi\xspace}

\def\eg{\emph{e.g}\onedot}

\def\etc{\emph{etc}\onedot} 
 
\def\etal{\emph{et al}\onedot}

\newcommand{\printfnsymbol}[1]{%
  \textsuperscript{\@fnsymbol{#1}}%
}

\makeatother

\begin{document}
\pagestyle{headings}
\mainmatter
\def\ECCVSubNumber{1203}  

\title{Fashionpedia: Ontology, Segmentation, and an Attribute Localization Dataset} 

\titlerunning{Fashionpedia}

\author{
Menglin Jia\thanks{equal contribution.}$^{1}$, 
Mengyun Shi\printfnsymbol{1}$^{1,4}$, 
Mikhail Sirotenko\printfnsymbol{1}$^{3}$, 
Yin Cui\printfnsymbol{1}$^{3}$, 
\\ 
Claire Cardie$^{1}$, 
Bharath Hariharan$^{1}$, 
Hartwig Adam$^{3}$, 
Serge Belongie$^{1,2}$  \\}
\institute{$^{1}$Cornell University \qquad $^{2}$Cornell Tech \qquad $^{3}$Google Research \qquad $^{4}$Hearst Magazines}
\authorrunning{M. Jia et al.}

\maketitle

\begin{abstract}
In this work we explore the task of \emph{instance segmentation with attribute localization}, which unifies instance segmentation (detect and segment each object instance) and fine-grained visual attribute categorization (recognize one or multiple attributes).
The proposed task requires both localizing an object and describing its properties.
To illustrate the various aspects of this task, we focus on the domain of fashion and introduce \emph{Fashionpedia} as a step toward mapping out the visual aspects of the fashion world.
Fashionpedia consists of two parts: 
(1) an ontology built by fashion experts containing 27 main apparel categories, 19 apparel parts, 294 fine-grained attributes and their relationships; 
(2) a dataset with everyday and celebrity event fashion images annotated with segmentation masks and their associated per-mask fine-grained attributes, built upon the Fashionpedia ontology. 
In order to solve this challenging task, we propose a novel Attribute-Mask R-CNN model to jointly perform instance segmentation and localized attribute recognition, and provide a novel evaluation metric for the task.
Fashionpedia is available at: \footnotesize\url{https://fashionpedia.github.io/home/}.

\keywords{Dataset, Ontology, Instance Segmentation, Fine-Grained, Attribute, Fashion}
\end{abstract}

\begin{figure}[t]
  \centering
  \includegraphics[width=\columnwidth]{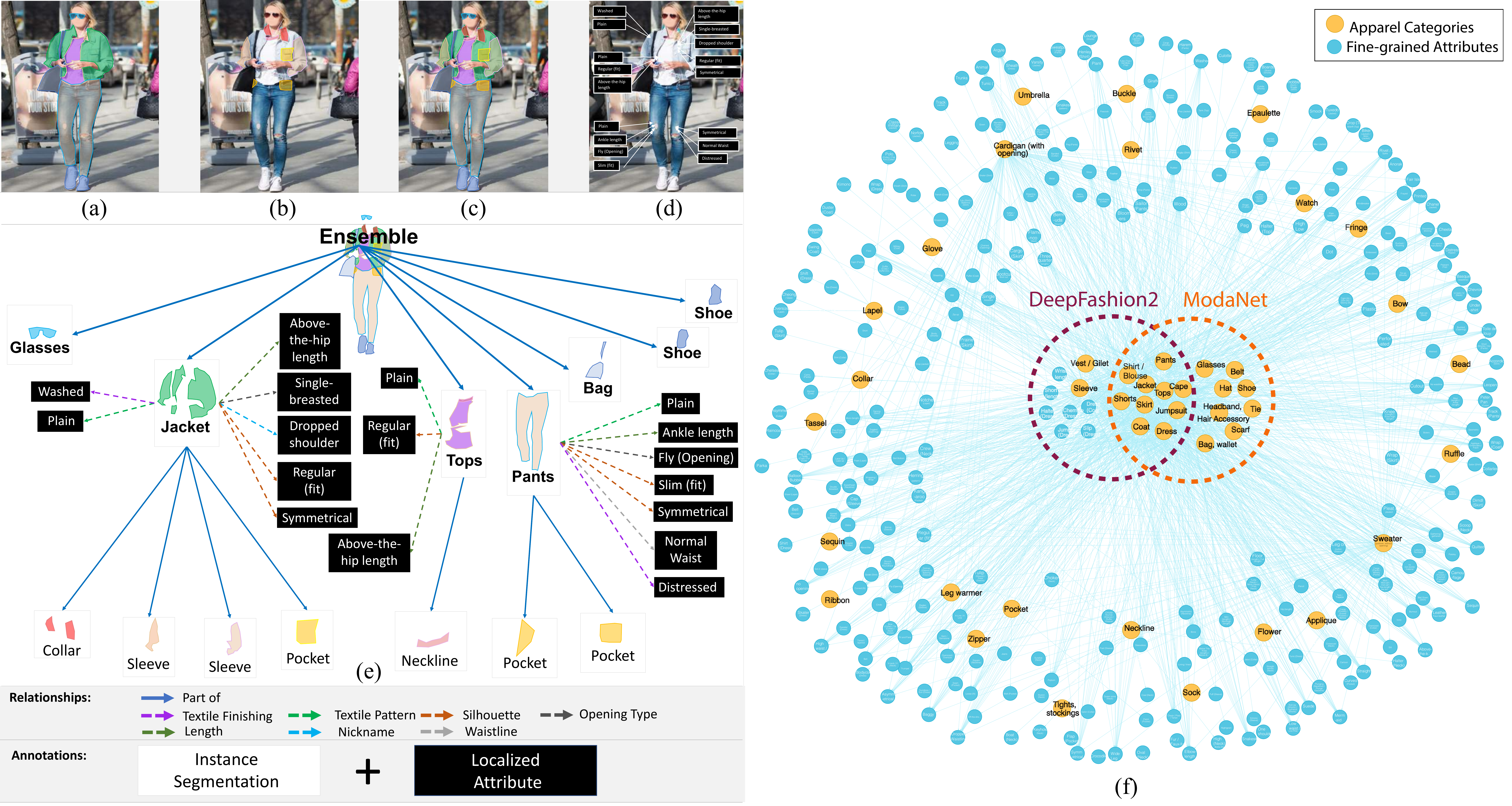}
  \caption{\textbf{An illustration of the Fashionpedia dataset and ontology:} (a) main garment masks; (b) garment part masks; (c) both main garment and garment part masks; (d) fine-grained apparel attributes; (e) an exploded view of the annotation diagram: the image is annotated with both instance segmentation masks \textit{(white boxes)} and per-mask fine-grained attributes \textit{(black boxes)}; (f) visualization of the Fashionpedia ontology: we created Fashionpedia ontology and separate the concept of categories \textit{(yellow nodes)} and attributes~\cite{rosch1975cognitive} \textit{(blue nodes)} in fashion. It covers pre-defined garment categories used by both Deepfashion2~\cite{ge_deepfashion2:_2019} and ModaNet~\cite{zheng_modanet:_2018}. Mapping with DeepFashion2 also shows the versatility of using attributes and categories. We are able to present all 13 garment classes in DeepFashion2 with 11 main garment categories, 1 garment part, and 7 attributes}
  \label{fig:overview}
\end{figure}

\section{Introduction}
Recent progress in the field of computer vision has advanced machines' ability to recognize and understand our visual world, showing significant impacts in fields including autonomous driving~\cite{yu2018bdd100k}, product recognition~\cite{liu_deepfashion:_2016,guo2019imaterialist}, \etc. These real-world applications are fueled by various visual understanding tasks with the goals of \textit{naming}, \textit{describing (attribute recognition)}, or \textit{localizing} objects within an image.

\textit{Naming and localizing} objects is formulated as an object detection task (Figure~\ref{fig:overview}(a-c)).
As a hallmark for computer recognition, this task is to identify and indicate the boundaries of objects in the form of bounding boxes or segmentation masks~\cite{girshick2014rich,ren2015faster,he2017mask}. 
\textit{Attribute recognition}~\cite{farhadi2009describing,kumar2009attribute,ferrari2008learning,parikh2011relative} (Figure~\ref{fig:overview}(d)) instead focuses on describing and comparing objects, since an object also has many other properties or attributes in addition to its category.
Attributes not only provide a compact and scalable way to represent objects in the world, as pointed out by Ferrari and Zisserman~\cite{ferrari2008learning}, attribute learning also enables transfer of existing knowledge to novel classes.
This is particularly useful for fine-grained visual recognition, with the goal of distinguishing subordinate visual categories such as birds~\cite{wah_caltech-ucsd_nodate} or natural species~\cite{van_horn_inaturalist_2017}.

In the spirit of mapping the visual world, we propose a new task, \textit{instance segmentation with attribute localization}, which unifies object detection and fine-grained attribute recognition.
As illustrated in Figure~\ref{fig:overview}(e), this task offers a structured representation of an image.
Automatic recognition of a rich set of attributes for each segmented object instance complements category-level object detection and therefore advance the degree of complexity of images and scenes we can make understandable to machines.
In this work, we focus on the fashion domain as an example to illustrate this task.
Fashion comes with rich and complex apparel with attributes, influences many aspects of modern societies, and has a strong financial and cultural impact.
We anticipate that the proposed task is also suitable for other man-made product domains such as automobile and home interior.

Structured representations of images often rely on structured vocabularies~\cite{krishna_visual_2016}.
With this in mind, we construct the Fashionpedia ontology (Figure~\ref{fig:overview}(f)) and image dataset (Figure~\ref{fig:overview}(a-e)), annotating fashion images with detailed segmentation masks for apparel categories, parts, and their attributes.
Our proposed ontology provides a rich schema for interpretation and organization of individuals' garments, styles, or fashion collections~\cite{kendall2019ontology}. 
For example, we can create a knowledge graph (see supplementary material for more details) by aggregating structured information within each image and exploiting relationships between garments and garment parts, categories, and attributes in the Fashionpedia ontology. 
Our insight is that a large-scale fashion segmentation and attribute localization dataset built with a fashion ontology can help computer vision models achieve better performance on fine-grained image understanding and reasoning tasks.

The contributions of this work are as follows:

A \textit{novel task} of fine-grained instance segmentation with attribute localization.
The proposed task unifies instance segmentation and visual attribute recognition, which is an important step toward structural understanding of visual content in real-world applications.

A \textit{unified fashion ontology} informed by product descriptions from the internet and built by fashion experts. 
Our ontology captures the complex structure of fashion objects and ambiguity in descriptions obtained from the web, containing 46 apparel objects (27 main apparels and 19 apparel parts), and 294 fine-grained attributes (spanning 9 super categories) in total.
To facilitate the development of related efforts, we also provide a mapping with categories from existing fashion segmentation datasets, see Figure~\ref{fig:overview}(f).

A \textit{dataset} with a total of 48,825 clothing images in daily-life, street-style, celebrity events, runway, and online shopping annotated both by crowd workers for segmentation masks and fashion experts for localized attributes, with the goal of developing and benchmarking computer vision models for comprehensive understanding of fashion.

A new \textit{model}, Attribute-Mask R-CNN, is proposed with the aim of jointly performing instance segmentation and localized attribute recognition; a novel \textit{evaluation metric} for this task is also provided.

\begin{table*}[t]
\begin{center}
\caption{Comparison of fashion-related datasets: (Cls. = Classification, Segm. = Segmentation, MG = Main Garment, GP = Garment Part, A = Accessory, S = Style, FGC = Fine-Grained Categorization). To the best of our knowledge, we include all fashion-related datasets focusing on visual recognition}
\label{table:1}
\resizebox{1\textwidth}{!}{%
\begin{tabular}{ l  c c c  c c c }
\Xhline{1.0pt}\noalign{\smallskip}
\textbf{Name} & \multicolumn{3}{ c }{\textbf{Category Annotation Type}} & \multicolumn{3}{ c }{\textbf{Attribute Annotation Type}} \\ 
\cline{2-7}\noalign{\smallskip}
& \textbf{Cls.} & \textbf{BBox} & \textbf{Segm.} & \textbf{Unlocalized} & \textbf{Localized} & \textbf{FGC}\\
\Xhline{1.0pt}\noalign{\smallskip}
Clothing Parsing~\cite{yamaguchi_parsing_2012}  &MG, A   &-   &- &-  &-  &-\\
Chic or Social~\cite{yamaguchi_chic_2014}  &MG, A   &-   &-  &-  &-   &-\\
Hipster~\cite{fleet_hipster_2014}  &MG, A, S  &-   &-   &-  &-   &-\\
Ups and Downs~\cite{he_ups_2016}  & MG &-   &-  &-  &-   &- \\
Fashion550k~\cite{inoue_multi-label_2017}  &MG, A  &-   &-  &-  &-   &- \\
Fashion-MNIST~\cite{xiao_fashion-mnist:_2017} &MG  &-   & - &-  &-   &- \\
\hline
Runway2Realway~\cite{vittayakorn_runway_2015}  &-  &-   &MG, A   &-  &-  &-\\
ModaNet~\cite{zheng_modanet:_2018} &-  &MG, A   &MG, A  &-   &-   &-\\
Deepfashion2~\cite{ge_deepfashion2:_2019} &-  &MG   &MG  &-  &-   &-\\
\hline\noalign{\smallskip}
Fashion144k~\cite{simo-serra_neuroaesthetics_2015}  & MG, A &-   &-  &\checkmark   &-   &-\\
Fashion Style-128 Floats~\cite{simo-serra_fashion_2016}  & S &-   &-  &\checkmark  &-   &-\\
UT Zappos50K~\cite{yu2017semantic}  &A  &-   &-  &\checkmark  &-   &-\\
Fashion200K~\cite{han_automatic_2017}  &MG  &-   &-  &\checkmark   &-  &-\\
FashionStyle14~\cite{takagi_what_2017}  & S  &-   &-  &\checkmark  &-   &-\\ \hline
Main Product Detection~\cite{yu_multi-modal_2017} &-  &MG   &-  &\checkmark  &-   &-\\ 
\hline\noalign{\smallskip}
StreetStyle-27K~\cite{matzen_streetstyle:_2017}  &-  &-   &-  &\checkmark   &-  &\checkmark \\ 
\hline\noalign{\smallskip}
UT-latent look~\cite{hsiao_learning_2017}  & MG, S  &-   &-  &\checkmark  &-   &\checkmark\\
FashionAI~\cite{FashionAI}  &MG, GP, A  &-   &-  &\checkmark   &-  &\checkmark \\
iMat-Fashion Attribute~\cite{guo2019imaterialist} & MG, GP, A, S  &-   & - &\checkmark   & - &\checkmark \\
\hline\noalign{\smallskip}
Apparel classification-Style~\cite{hutchison_apparel_2013}  &-  & MG  &-  &\checkmark  &-   &\checkmark \\
DARN~\cite{huang_cross-domain_2015}  &-  & MG  &-  &\checkmark   &-  &\checkmark \\
WTBI~\cite{kiapour_where_2015}  &-  & MG, A &-  &\checkmark   &-   &\checkmark  \\
Deepfashion~\cite{liu_deepfashion:_2016} & S  &MG   & - &\checkmark   &-  &\checkmark \\
\Xhline{1.0pt}\noalign{\smallskip}
\textbf{Fashionpedia} &-  &\textbf{MG, GP, A}  & \textbf{MG, GP, A} &-  &\textbf{\checkmark}  & \textbf{\checkmark}  \\
\Xhline{1.0pt}\noalign{\smallskip}
\end{tabular}
}
\end{center}
\end{table*}


\section{Related Work}

The combined task of fine-grained instance segmentation and attribute localization has not received a great deal of attention in the literature.
On one hand, COCO~\cite{lin_microsoft_2014} and LVIS~\cite{Gupta_2019_CVPR} represent the benchmarks of object detection for common objects. Panoptic segmentation is proposed to unify both semantic and instance segmentation, addressing both stuff and thing classes~\cite{kirillov2019panoptic}. In spite of the domain differences, Fashionpedia has comparable mask qualities with LVIS and the similar total number of segmentation masks as COCO.
On the other hand, we have also observed an increasing effort to curate datasets for fine-grained visual recognition, evolved from CUB-200 Birds~\cite{wah_caltech-ucsd_nodate} to the recent iNaturalist dataset~\cite{van_horn_inaturalist_2017}. The goal of this line of work is to advance the state-of-the-art in automatic image classification for large numbers of real world, fine-grained categories. 
A rather unexplored area of these datasets, however, is to provide a structured representation of an image.
Visual Genome~\cite{krishna_visual_2016} provides dense annotations of object bounding boxes, attributes, and relationships in the general domain, enabling a structured representation of the image.
In our work, we instead focus on fine-grained attributes and provide segmentation masks in the fashion domain to advance the clothing recognition task.

Clothing recognition has received increasing attention in the computer vision community recently. 
A number of works provide valuable apparel-related datasets~\cite{yamaguchi_parsing_2012,hutchison_apparel_2013,yamaguchi_chic_2014,fleet_hipster_2014,vittayakorn_runway_2015,simo-serra_neuroaesthetics_2015,huang_cross-domain_2015,kiapour_where_2015,he_ups_2016,simo-serra_fashion_2016,liu_deepfashion:_2016,inoue_multi-label_2017,xiao_fashion-mnist:_2017,yu2017semantic,han_automatic_2017,takagi_what_2017,yu_multi-modal_2017,matzen_streetstyle:_2017,hsiao_learning_2017,zheng_modanet:_2018,FashionAI,ge_deepfashion2:_2019}. 
These pioneering works enabled several recent advances in clothing-related recognition and knowledge discovery~\cite{fu2019imp,Mall_2019_ICCV}.
Table~\ref{table:1} summarizes the comparison among different fashion datasets regarding annotation types of clothing categories and attributes. 
Our dataset distinguishes itself in the following three aspects.

\textbf{Exhaustive annotation of segmentation masks}: 
Existing fashion datasets ~\cite{vittayakorn_runway_2015,zheng_modanet:_2018,ge_deepfashion2:_2019} offer segmentation masks for the main garment (e.g., jacket, coat, dress) and the accessory categories (e.g., bag, shoe).
Smaller garment objects such as collars and pockets are not annotated. 
However, these small objects could be valuable for real-world applications (e.g., searching for a specific collar shape during online-shopping). 
Our dataset is not only annotated with the segmentation masks for a total of 27 main garments and accessory categories but also 19 garment parts (e.g., collar, sleeve, pocket, zipper, embroidery).

\textbf{Localized attributes}: 
The fine-grained attributes from existing datasets ~\cite{huang_cross-domain_2015,liu_deepfashion:_2016,yu_multi-modal_2017,guo2019imaterialist} tend to be noisy, mainly because most of the annotations are collected by crawling fashion product attribute-level descriptions directly from large online shopping websites. 
Unlike these datasets, fine-grained attributes in our dataset are annotated manually by fashion experts. 
To the best of our knowledge, ours is the only dataset to annotate localized attributes: fashion experts are asked to annotate attributes associated with the segmentation masks labeled by crowd workers.
Localized attributes could potentially help computational models detect and understand attributes more accurately.

\textbf{Fine-grained categorization}: 
Previous studies on fine-grained attribute categorization suffer from several issues including: (1) repeated attributes belonging to the same category (e.g., zip,zipped and zipper)~\cite{liu_deepfashion:_2016,hsiao_learning_2017}; (2) basic level categorization only (object recognition) and lack of fine-grained categorization~\cite{yamaguchi_parsing_2012,hutchison_apparel_2013,yamaguchi_chic_2014,fleet_hipster_2014,kiapour_where_2015,vittayakorn_runway_2015,simo-serra_fashion_2016,takagi_what_2017,inoue_multi-label_2017,han_automatic_2017,xiao_fashion-mnist:_2017,zheng_modanet:_2018,ge_deepfashion2:_2019}; (3) lack of fashion taxonomies with the needs of real-world applications for the fashion industry, possibly due to the research gap in fashion design and computer vision;
(4) diverse taxonomy structures from different sources in fashion domain.
To facilitate research in the areas of fashion and computer vision, our proposed ontology is built and verified by fashion experts based on their own design experiences and informed by the following four sources: (1) world-leading e-commerce fashion websites (e.g., ZARA, H\&M, Gap, Uniqlo, Forever21); (2) luxury fashion brands (e.g., Prada, Chanel, Gucci); (3) trend forecasting companies (e.g., WGSN); (4) academic resources~\cite{fashionpedianodate,bloomsburycom}.


\section{Dataset Specification and Collection}

\subsection{Ontology specification}

We propose a unified fashion ontology (Figure~\ref{fig:overview}(f)), a structured vocabulary that utilizes the basic level categories and fine-grained attributes~\cite{rosch1975cognitive}.
The Fashionpedia ontology relies on similar definitions of object and attributes as previous well-known image datasets. For example, a Fashionpedia object is similar to ``item'' in Wikidata~\cite{vrandevcic2014wikidata}, or ``object'' in COCO~\cite{lin_microsoft_2014} and Visual Genome~\cite{krishna_visual_2016}).
In the context of Fashionpedia, objects represent common items in apparel (e.g., jacket, shirt, dress).
In this section, we break down each component of the Fashionpedia ontology and illustrate the construction process. With this ontology and our image dataset, a large-scale fashion knowledge graph can be built as an extended application of our dataset (more details can be found in the supplementary material).

\begin{figure*}[t]
\centering
\includegraphics[width=1\columnwidth]{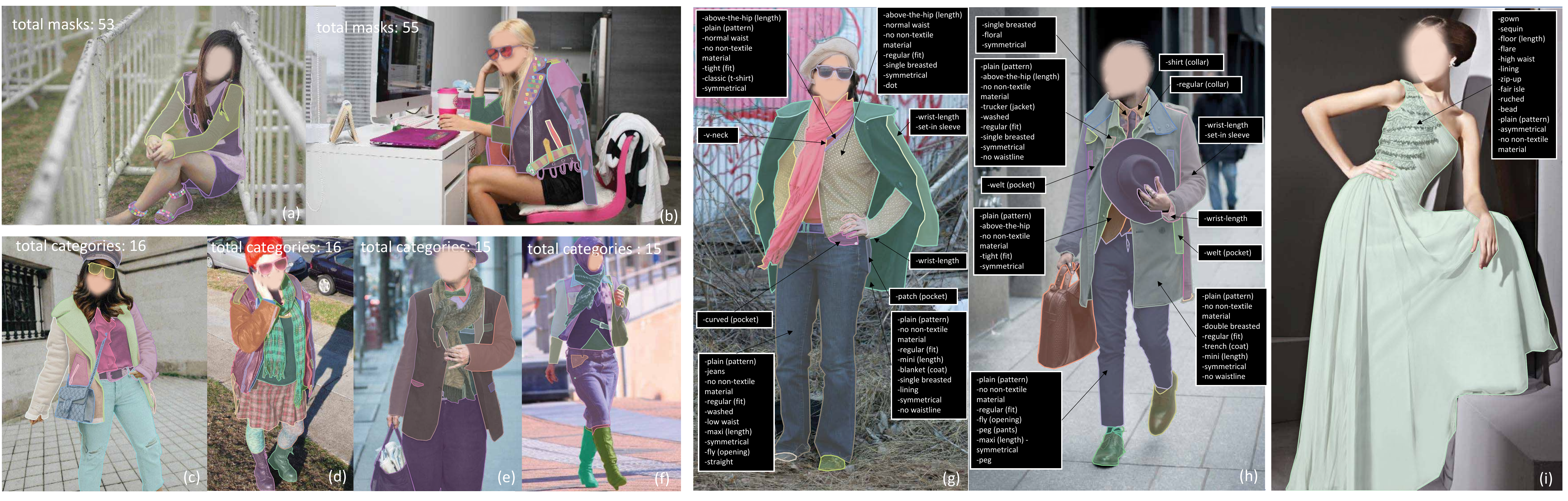}
\caption{Image examples with annotated segmentation masks (a-f) and fine-grained attributes (g-i)}
\label{fig:img_examples}
\end{figure*}

\textbf{Apparel categories.}
In the Fashionpedia dataset, all images are annotated with one or multiple main garments. Each main garment is also annotated with its garment parts. 
For example, general garment types such as jacket, dress, pants are considered as main garments. These garments also consist of several garment parts such as collars, sleeves, pockets, buttons, and embroideries. Main garments are divided into three main categories: outerwear, intimate and accessories. Garment parts also have different types: garment main parts (e.g., collars, sleeves), bra parts, closures (e.g., button, zipper) and decorations (e.g., embroidery, ruffle). On average, each image consists of 1 person, 3 main garments, 3 accessories, and 12 garment parts, each delineated by a tight segmentation mask (Figure~\ref{fig:overview}(a-c)).
Furthermore, each object is assigned to a synset ID in our Fashionpedia ontology.

\textbf{Fine-grained attributes.}
Main garments and garment parts can be associated with apparel attributes (Figure~\ref{fig:overview}(e)). 
For example, ``button'' is part of the main garment ``jacket''; ``jacket'' can be linked with the silhouette attribute ``symmetrical''; the garment part ``button'' could contain the attribute ``metal'' with a relationship of material. 
The Fashionpedia ontology provide attributes for 13 main outerwear garments categories, and 5 out of 19 garments parts (``sleeve'', ``neckline'', ``pocket'', ``lapel'', and ``collar'').
Each image has 16.7 attributes on average (max 57 attributes).
As with the main garments and garment parts, we canonicalize all attributes to our Fashionpedia ontology. 

\textbf{Relationships.}
Relationships can be formed between categories and attributes.
There are three main types of relationships (Figure~\ref{fig:overview}(e)): 
(1) outfits to main garments, main garments to garment parts: meronymy (part-of) relationship; 
(2) main garments to attributes or garment parts to attributes: these relationship types can be garment silhouette (e.g., peplum), collar nickname (e.g., peter pan collars), garment length (e.g., knee-length), textile finishing (e.g., distressed), or textile-fabric patterns (e.g., paisley), etc.; 
(3) within garments, garment parts or attributes: there is a maximum of four levels of hyponymy (is-an-instance-of) relationships. For example, weft knit is an instance of knit fabric, and the fleece is an instance of weft knit.

\subsection{Image Collection and Annotation Pipeline}
\textbf{Image collection.}
A total of 50,527 images were harvested from Flickr and free license photo websites, which included Unsplash, Burst by Shopify, Freestocks, Kaboompics, and Pexels.
Two fashion experts verified the quality of the collected images manually. 
Specifically, the experts checked a scenes' diversity and made sure clothing items were visible in the images.
Fdupes~\cite{fdupes} is used to remove duplicated images. 
After filtering, 48,825 images were left and used to build our Fashionpedia dataset.

\textbf{Annotation Pipeline.}
Expert annotation is often a time-consuming process.
In order to accelerate the annotation process, we decoupled the work between crowd workers and experts. We divided the annotation process into the following two phases.

First, segmentation masks with apparel objects are annotated by 28 crowd workers, who were trained for 10 days before the annotation process (with prepared annotation tutorials of each apparel object, see supplementary material for details).
We collected high-quality annotations by having the annotators follow the contours of garments in the image as closely as possible
(See section~\ref{sec: mask_ana} for annotation analysis).
This polygon annotation process was monitored daily and verified weekly by a supervisor and by the authors.

Second, 15 fashion experts (graduate students in the apparel domain) were recruited to annotate the fine-grained attributes for the annotated segmentation masks.
Annotators were given one mask and one attribute super-category (``textile pattern'', ``garment silhouette'' for example) at a time.
An additional two options, ``not sure'' and ``not on the list'' were added during the annotation.
The option ``not on the list'' indicates that the expert found an attribute that is not on the proposed ontology.
If ``not sure'' is selected, it means the expert can not identify the attribute of one mask. Common reasons for this selection include occlusion of the masks and viewing angles of the image; (for example, a top underneath a closed jacket).
More details can be found in Figure~\ref{fig:img_examples}. 
Each attribute supercategory is assigned to one or two fashion experts, depending on the number of masks.
The annotations are also checked by another expert annotator before delivery.

We split the data into training, validation and test sets, with 45,623, 1158, 2044 images respectively. More details of the dataset creation can be found in the supplementary material.


\section{Dataset Analysis}
\label{sec: dataset_analysis}
This section will discuss a detailed analysis of our dataset using the training images.
We begin by discussing general image statistics, followed by an analysis of segmentation masks, categories, and attributes. 
We compare Fashionpedia with four other segmentation datasets, including two recent fashion datasets DeepFashion2~\cite{ge_deepfashion2:_2019} and ModaNet~\cite{zheng_modanet:_2018}, and two general domain datasets COCO~\cite{lin_microsoft_2014} and LVIS~\cite{Gupta_2019_CVPR}.

\subsection{Image Analysis}
\label{sec: img_ana}

We choose to use images with high resolutions during the curating process since Fashionpedia ontology includes diverse fine-grained attributes for both garments and garment parts. 
The Fashionpedia training images have an average dimension of $1710$ (width) $\times$ $2151$ (height).
Images with high resolutions are able to show apparel objects in detail, leading to more accurate and faster annotations for both segmentation masks and attributes. These high resolution images can also benefit downstream tasks such as detection and image generation. 
Examples of detailed annotations can be found in Figure~\ref{fig:img_examples}.

\subsection{Mask Analysis}
\label{sec: mask_ana}
We define ``masks'' as one apparel instance that may have more than one separate components (the jacket in Figure~\ref{fig:overview} for example), ``polygon'' as a disjoint area.

\begin{figure}[t]
\centering
\subfigure[\scriptsize{Mask count per image across datasets}]{
    \includegraphics[scale=0.2]{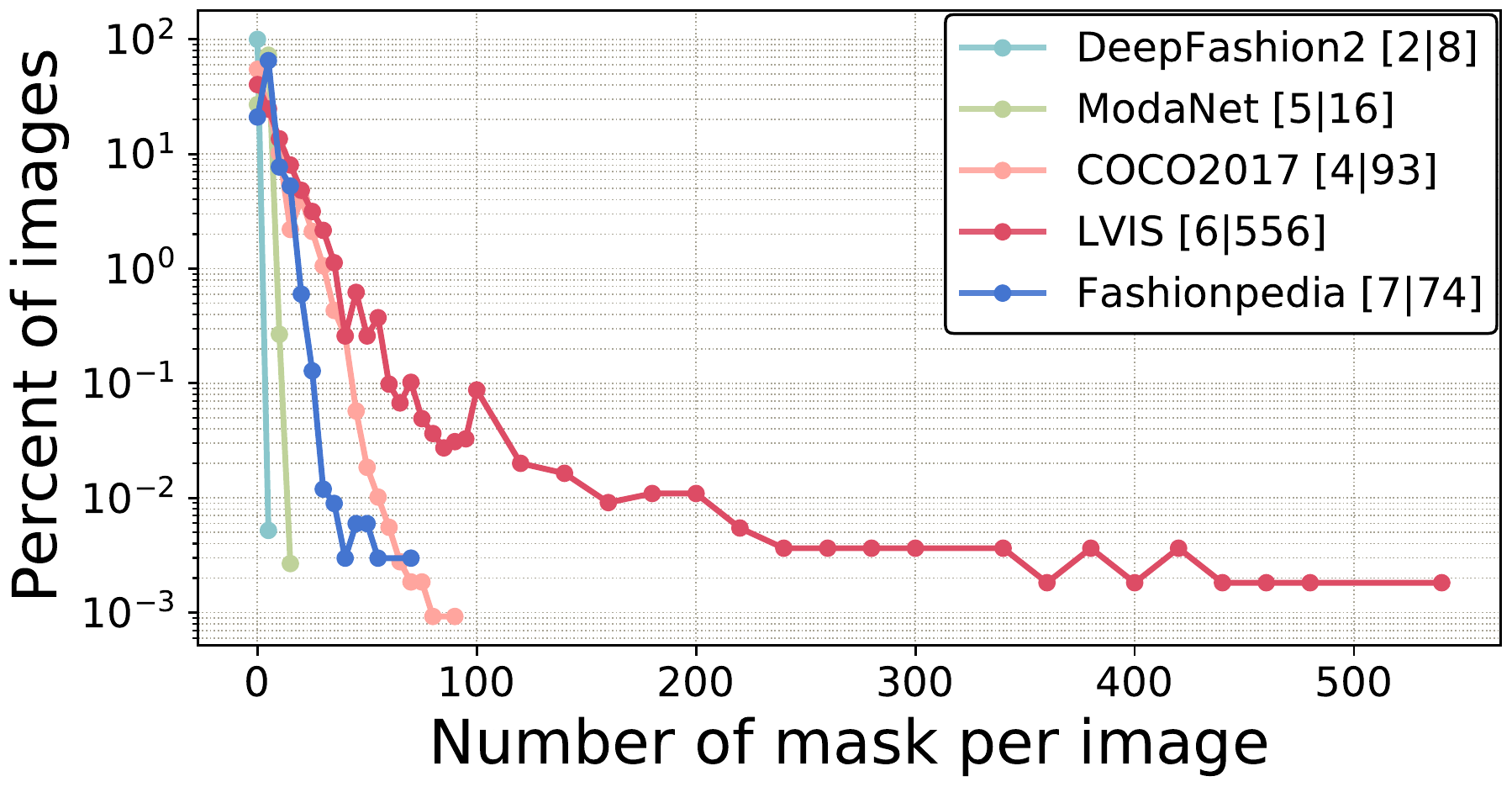}
    \label{fig:mask_per_img}
}
\hfill
\subfigure[\scriptsize{Relative mask size across datasets}]{
    \includegraphics[scale=0.2]{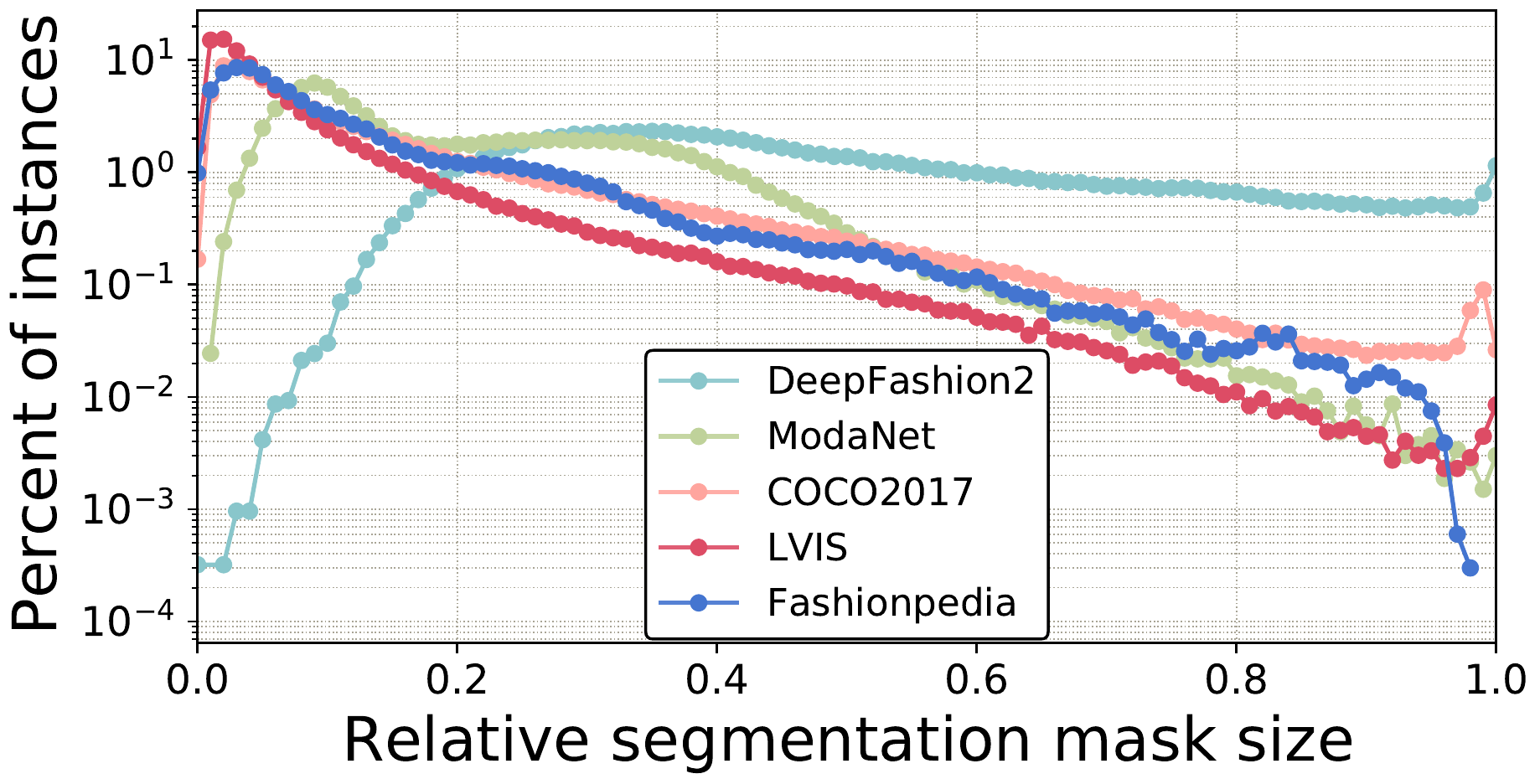}
    \label{fig:relative_mask_size}
}
\hfill
\subfigure[\scriptsize{Category diversity per image across datasets}]{
    \includegraphics[scale=0.2]{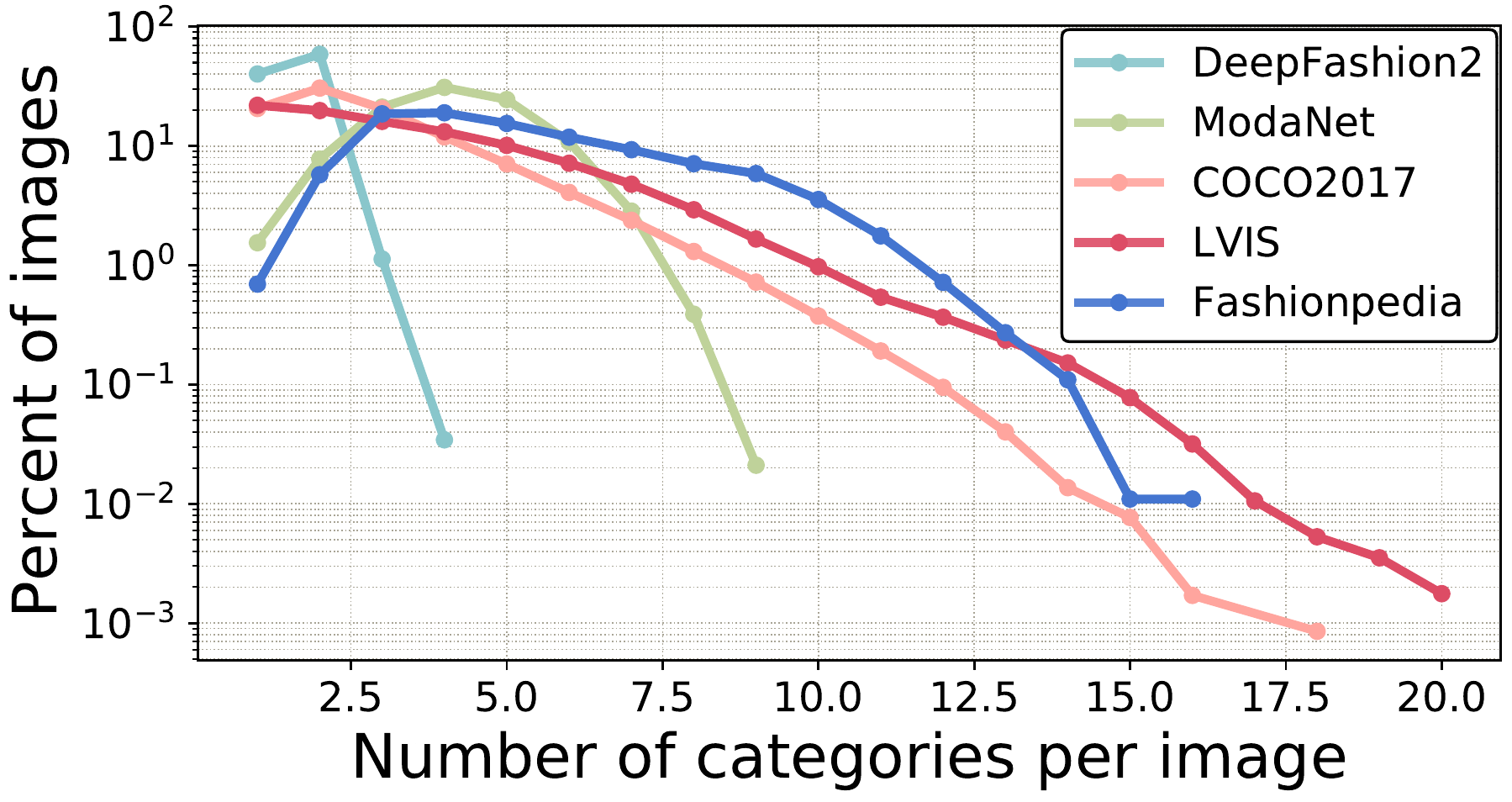}
    \label{fig:cat_per_image}
}
\subfigure[\scriptsize{Mask count per image in Fashionpedia}]{
    \includegraphics[scale=0.2]{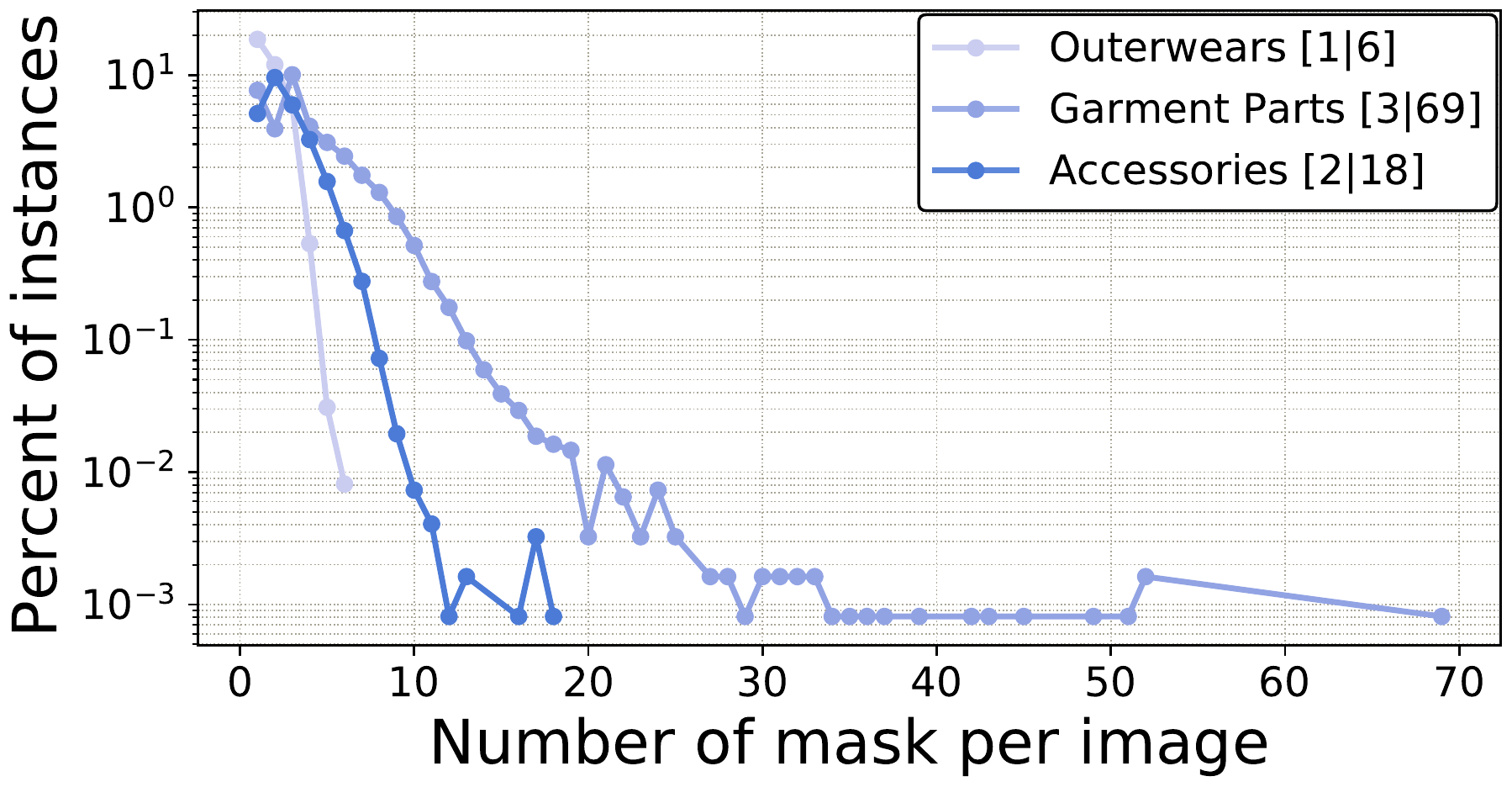}
    \label{fig:mask_per_img_per_cat}
}
\hfill
\subfigure[\scriptsize{Relative mask size in Fashionpedia}]{
    \includegraphics[scale=0.2]{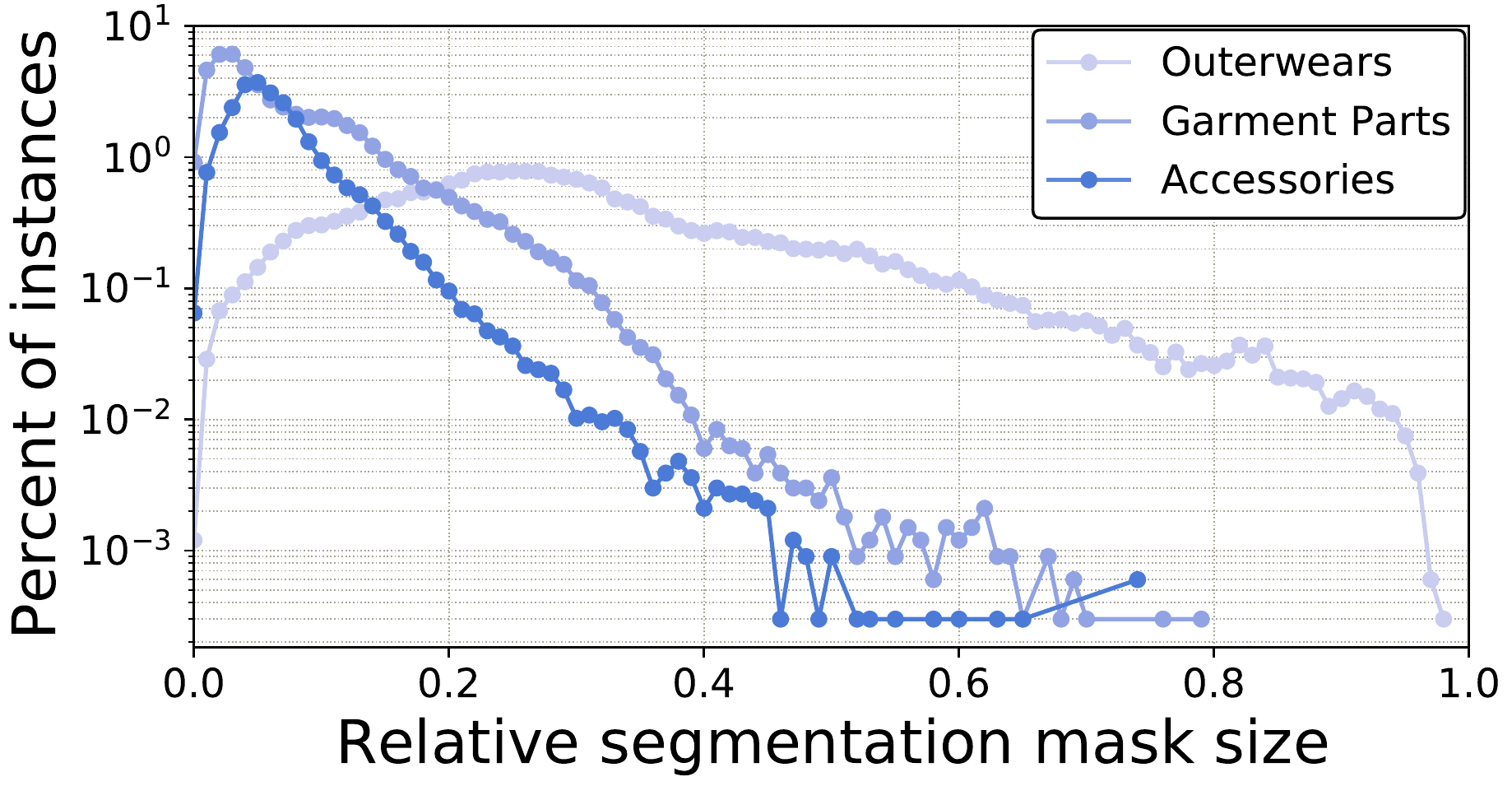}
    \label{fig:relative_mask_size_per_cat}
}
\hfill
\subfigure[\scriptsize{Categories and attributes distribution in Fashionpedia}]{
    \includegraphics[scale=0.2]{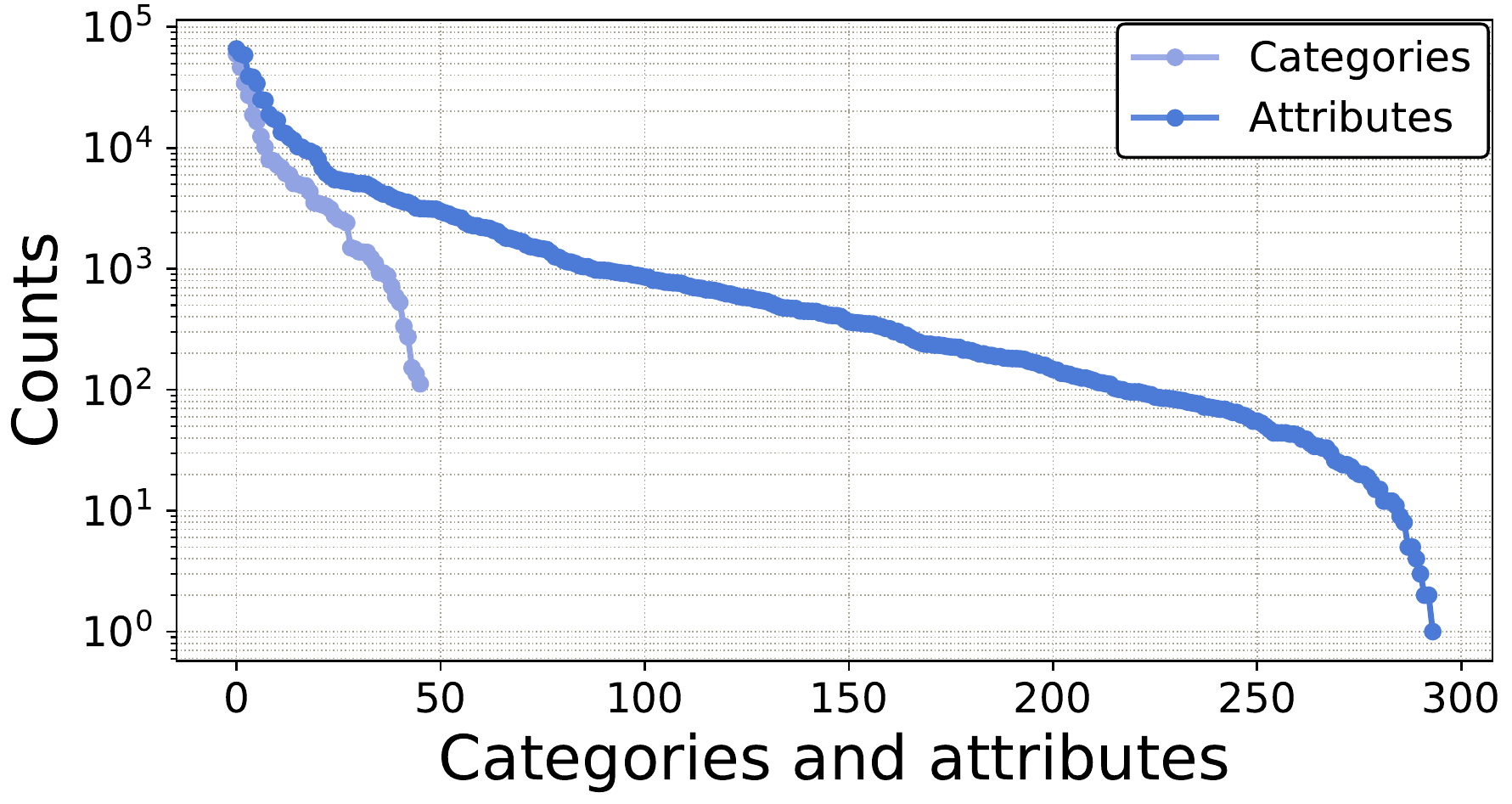}
    \label{fig:cat_att_dis}
}

\caption{\textbf{Dataset statistics:} First row presents comparison among datasets. Second row presents comparison within Fashionpedia. Y-axes are in log scale. Relative segmentation mask size were calculated same as~\cite{Gupta_2019_CVPR}. Relative mask size was rounded to precision of 2. For mask count per image comparisons (Figure~\ref{fig:mask_per_img} and~\ref{fig:mask_per_img_per_cat}), legends follow [\textit{median $\vert$ max}] format. Values in X-axis for Figure~\ref{fig:mask_per_img} was discretized for better visual effect}
\label{fig:mask_area_comp}
\end{figure}

\textbf{Mask quantity.} On average, there are 7.3 (median 7, max 74) number of masks per image in the Fashionpedia training set. 
Figure~\ref{fig:mask_per_img} shows that the Fashionpedia has the largest median value in the 5 datasets used for comparison.
Fashionpedia also has the widest range among three fashion datasets, and a comparable range with COCO, which is a dataset in a general domain.
Compared to ModaNet and Deepfashion2 datasets, Fashionpedia has the widest range of mask count distribution. However, COCO and LVIS maintain a wider distribution over Fashionpedia owing to their more common objects in their dataset. 
Figure~\ref{fig:mask_per_img_per_cat} illustrates the distribution within the Fashionpedia dataset. 
One image usually contains more garment parts and accessories than outerwears.

\textbf{Mask sizes.} Figure~\ref{fig:relative_mask_size} and~\ref{fig:relative_mask_size_per_cat} compares relative mask sizes within Fashionpedia and against other datasets. Ours has a similar distribution as COCO and LVIS, except for a lack of larger masks (area $>0.95$). DeepFashion2 has a heavier tail, meaning it contains a larger portion of garments with a zoomed-in view. Unlike DeepFashion2, our images mainly focused on the whole ensemble of clothing. Since ModaNet focuses on outwears and accessories, it has more masks with relative area between $0.2$ and $0.4$. whereas ours has an additional 19 apparel parts categories. As illustrated in Figure~\ref{fig:relative_mask_size_per_cat}, garment parts and accessories are relatively small compared to the outerwear (e.g., ``dress'', ``coat'').

\textbf{Mask quality.} Apparel categories also tend to have complex silhouettes. Table~\ref{tab:mask_comp} shows that the Fashionpedia masks have the most complex boundaries amongst the five datasets (according to the measurement used in~\cite{Gupta_2019_CVPR}). This suggests that our masks are better able to represent complex silhouettes of apparel categories more accurately than ModaNet and DeepFashion2. We also report the number of vertices per polygons, which is a measurement representing how granularity of the masks produced. Table~\ref{tab:mask_comp} shows that we have the second-highest average number of vertices among five datasets, next to LVIS.

\setlength{\tabcolsep}{4pt}
\begin{table}
\begin{center}
\caption{Comparison of segmentation mask complexity amongst segmentation datasets in both fashion and general domain (COCO 2017 instance training data was used). Each statistic (mean and median) represents a bootstrapped 95$\%$ confidence interval following~\cite{Gupta_2019_CVPR}. Boundary complexity was calculated according to~\cite{Gupta_2019_CVPR,attneave1956quantitative}. Reported mask boundary complexity for COCO and LVIS was different compared with~\cite{Gupta_2019_CVPR} due to different image resolution and image sets. The number of vertices per polygon is calculated as the number of vertices in one polygon. Polygon is defined as one disjoint area. Masks (and polygons) with zero area were ignored}
\label{tab:mask_comp}
\resizebox{1\textwidth}{!}{%
\begin{tabular}{ l  cc  cc  c }
\Xhline{1.0pt}\noalign{\smallskip}
\textbf{Dataset} & \multicolumn{2}{ c }{\textbf{Boundary complexity}} & \multicolumn{2}{ c }{\textbf{No. of vertices per polygon}} & \textbf{Images count}\\
& mean &median              & mean&median     &\\
\Xhline{1.0pt}\noalign{\smallskip}
COCO~\cite{lin_microsoft_2014} &6.65 - 6.66  &6.07 - 6.08   & 21.14 - 21.21 &15.96 - 16.04 &  118,287
\\
LVIS~\cite{Gupta_2019_CVPR} &6.78 - 6.80  & 5.89 - 5.91   &\textbf{35.77 - 35.95} &\textbf{22.91 - 23.09} & 57,263
\\
ModaNet~\cite{zheng_modanet:_2018}  & 5.87 - 5.89  & 5.26 - 5.27  &22.50 - 22.60 &18.95 - 19.05 & 
52,377 \\
Deepfashion2~\cite{ge_deepfashion2:_2019} &4.63 - 4.64  &4.45 - 4.46 & 14.68 - 14.75 & 8.96 - 9.04 & 
\textbf{191,960} \\
\Xhline{1.0pt}\noalign{\smallskip}
Fashionpedia  & \textbf{8.36 - 8.39}  & \textbf{7.35 - 7.37} &31.82 - 32.01 &20.90 - 21.10 & 
45, 623 \\
\Xhline{1.0pt}\noalign{\smallskip}
\end{tabular}
}
\end{center}
\end{table}
\setlength{\tabcolsep}{1.4pt}

\subsection{Category and Attributes Analysis}
There are 46 apparel categories and 294 attributes presented in the Fashionpedia dataset. 
On average, each image was annotated with 7.3 instances, 5.4 categories, and 16.7 attributes.
Of all the masks with categories and attributes, each mask has 3.7 attributes on average (max 14 attributes).
Fashionpedia has the most diverse number of categories within one image among three fashion datasets, while comparable to COCO (Figure~\ref{fig:cat_per_image}), since we provide a comprehensive ontology for the annotation.
In addition, Figure~\ref{fig:cat_att_dis} shows the distributions of categories and attributes in the training set, and highlights the long-tailed nature of our data. 

During the fine-grained attributes annotation process, we also ask the experts to choose ``not sure'' if they are uncertain to make a decision, ``not on the list'' if they find another attribute that not provided.
the majority of ``not sure'' comes from three attributes superclasses, namely ``Opening Type'', ``Waistline'', ``Length''.
Since there are masks only show a limited portion of apparel (a top inside a jacket for example), the annotators are not sure how to identify those attributes due to occlusion or viewpoint discrepancies. 
Less than $15\%$ of masks for each attribute superclasses account for ``not on the list'', which illustrates the comprehensiveness of our proposed ontology (see supplementary material for more details of the extra dataset analysis).


\section{Evaluation Protocol and Baselines}
\label{sec: evaluation_baseline}

\subsection{Evaluation metric}
In object detection, a true positive (TP) for each category $c$ is defined as a single detected object that matches a ground truth object with a Intersection over Union (IoU) over a threshold $\tau_{\text{IoU}}$.
COCO's main evaluation metric uses average precision averaged across all $10$ IoU thresholds $\tau_{\text{IoU}} \in [0.5 : 0.05 : 0.95]$ and all 80 categories.
We denote such metric as AP$_{\text{IoU}}$.

In the case of instance segmentation and attribute localization, we extend standard COCO metric by adding one more constraint: the macro F$_1$ score for predicted attributes of single detected object with category $c$ (see supplementary material for the average choice of f1-score).
We denote the F$_1$ threshold as $\tau_{\text{F}_1}$, and it has the same range as $\tau_{\text{IoU}}$ ($\tau_{\text{F}_1} \in [0.5 : 0.05 : 0.95]$).
The main metric AP$_{\text{IoU+F}_1}$ reports averaged precision score across all $10$ IoU thresholds, all $10$ macro F$_1$ scores, and all the categories.
Our evaluation API, code and trained models are available at: \footnotesize\url{https://fashionpedia.github.io/home/Model_and_API.html}.

\begin{figure}[t]
  \centering
  \includegraphics[width=\columnwidth]{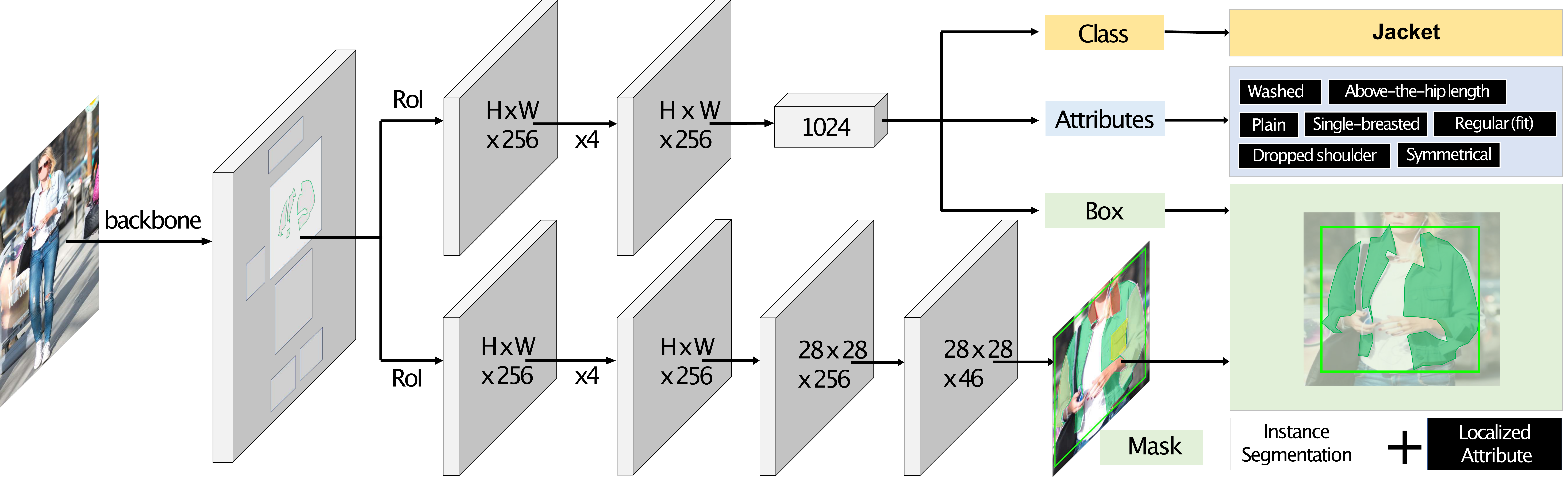}
  \caption{Attribute-Mask R-CNN adds a multi-label attribute prediction head upon Mask R-CNN for instance segmentation with attribute localization}
  \label{fig:baselinemodel}
\end{figure}

\subsection{Attribute-Mask R-CNN}
We perform two tasks on Fashionpedia: (1) apparel instance segmentation (ignoring attributes); (2) instance segmentation with attribute localization.
To better facilitate research on Fashionpedia, we design a strong baseline model named Attribute-Mask R-CNN that is built upon Mask R-CNN~\cite{he2017mask}.
As illustrated in Figure~\ref{fig:baselinemodel}, we extend Mask R-CNN heads to include an additional multi-label attribute prediction head, which is trained with sigmoid cross-entropy loss.
Attribute-Mask R-CNN can be trained end-to-end for jointly performing instance segmentation and localized attribute recognition.

We leverage different backbones including ResNet-50/101 (R50/101)~\cite{resnet} with feature pyramid network (FPN)~\cite{fpn} and SpineNet-49/96/143~\cite{spinenet}.
The input image is resized to $1024$ of the longer edge to feed all networks except SpineNet-143. For SpineNet-143, we instead use the input size of $1280$.
During training, other than standard random horizontal flipping and cropping, we use large scale jittering that resizes an image to a random ratio between $(0.5, 2.0)$ of the target input image size, following Zoph~\etal~\cite{zoph2020rethinking}.
We use an open-sourced Tensorflow~\cite{abadi2016tensorflow} code base\footnote{\url{https://github.com/tensorflow/tpu/tree/master/models/official/detection}} for implementation and all models are trained with a batch size of $256$.
We follow the standard training schedule of 1$\times$ (5625 iterations), 2$\times$ (11250 iterations), 3$\times$ (16875 iterations) and 6$\times$ (33750 iterations) in Detectron2~\cite{wu2019detectron2}, with linear learning rate scaling suggested by Goyal~\etal~\cite{goyal2017accurate}.

\begin{table}[t]
\small
\begin{center}
\caption{Baseline results of Mask R-CNN and Attribute-Mask R-CNN on Fashionpedia. The big performance gap between AP$_{\text{IoU}}$ and AP$_{\text{IoU+F}_1}$ suggests the challenging nature of our proposed task}
\label{tab:baseline}
\begin{tabular}{ l | c | c | c | c | c }
\Xhline{1.0pt}
\textbf{Backbone} & \textbf{Schedule} & \textbf{FLOPs(B)} & \textbf{params(M)} & \textbf{AP$^{\text{box}}_{\text{IoU} / \text{IoU+F}_1}$} & \textbf{AP$^{\text{mask}}_{\text{IoU} / \text{IoU+F}_1}$} \\
\Xhline{1.0pt}
\multirow{4}{*}{R50-FPN}  & 1$\times$ & \multirow{4}{*}{296.7} & \multirow{4}{*}{46.4} & 38.7 / 26.6 & 34.3 / 25.5 \\
 & 2$\times$ & & & ~41.6 / 29.3~ & ~38.1 / 28.5~ \\
 & 3$\times$ & & & ~43.4 / 30.7~ & ~39.2 / 29.5~ \\
 & 6$\times$ & & & ~42.9 / 31.2~ & ~38.9 / 30.2~ \\
\hline
\multirow{4}{*}{R101-FPN}  & 1$\times$ & \multirow{4}{*}{374.3} & \multirow{4}{*}{65.4} & 41.0 / 28.6 & 36.7 / 27.6 \\
 & 2$\times$ & & & ~43.5 / 31.0~ & ~39.2 / 29.8~ \\
 & 3$\times$ & & & ~44.9 / 32.8~ & ~40.7 / 31.4~ \\
 & 6$\times$ & & & ~44.3 / 32.9~ & ~39.7 / 31.3~ \\
\hline
SpineNet-49  & \multirow{3}{*}{\textbf{6$\times$}} & 267.2 & 40.8 & 43.7 / 32.4 & 39.6 / 31.4 \\
SpineNet-96  & & 314.0 & 55.2 & 46.4 / 34.0 & 41.2 / 31.8 \\
\textbf{SpineNet-143} & & \textbf{498.0} & \textbf{79.2} & \textbf{48.7 / 35.7} & \textbf{43.1 / 33.3} \\
\Xhline{1.0pt}
\end{tabular}%
\end{center}
\end{table}

\begin{table}
\begin{center}
\caption{Per super-category results (for masks) using Attribute-Mask R-CNN with SpineNet-143 backbone. We follow the same COCO sub-metrics for overall and three super-categories for apparel objects. 
Result format follows [AP$_{\text{IoU}}$ / AP$_{\text{IoU}+\text{F}_1}$] or [AR$_{\text{IoU}}$ / AR$_{\text{IoU}+\text{F}_1}$] (see supplementary material for per-class results)}
\label{tab:per_cls_results}
\begin{tabular}{ l | c | c | c | c | c | c }
\hline
\textbf{Category}
&\textbf{AP} &\textbf{AP50} &\textbf{AP75} &\textbf{APl} & \textbf{APm} & \textbf{APs}
\\
\hline
overall
& 43.1 / 33.3  &60.3 / 42.3 &47.6 / 37.6 &50.0 / 35.4 &40.2 / 27.0 &17.3 / 9.4\\
\hline
outerwear
& 64.1 / 40.7 &77.4 / 49.0 &72.9 / 46.2 &67.1 / 43.0 &44.4 / 29.3 &19.0 / 4.4 \\
parts
& 19.3 / 13.4 & 35.5 / 20.8 & 18.4 / 14.4 & 28.3 / 14.5 & 23.9 / 16.4 & 12.5 / 9.8 \\
accessory
&56.1 / ~~-~~   & 77.9 / ~~-~~   & 63.9 / ~~-~~   & 57.5 / ~~-~~   & 60.5 / ~~-~~   & 25.0 / ~~-~~   \\
\hline
\end{tabular}
\end{center}
\end{table}

\begin{figure*}[th]
\centering
\subfigure[Fashionpedia]{
    \includegraphics[scale=0.31, clip=true, trim = 13mm 60mm 25mm 60mm]{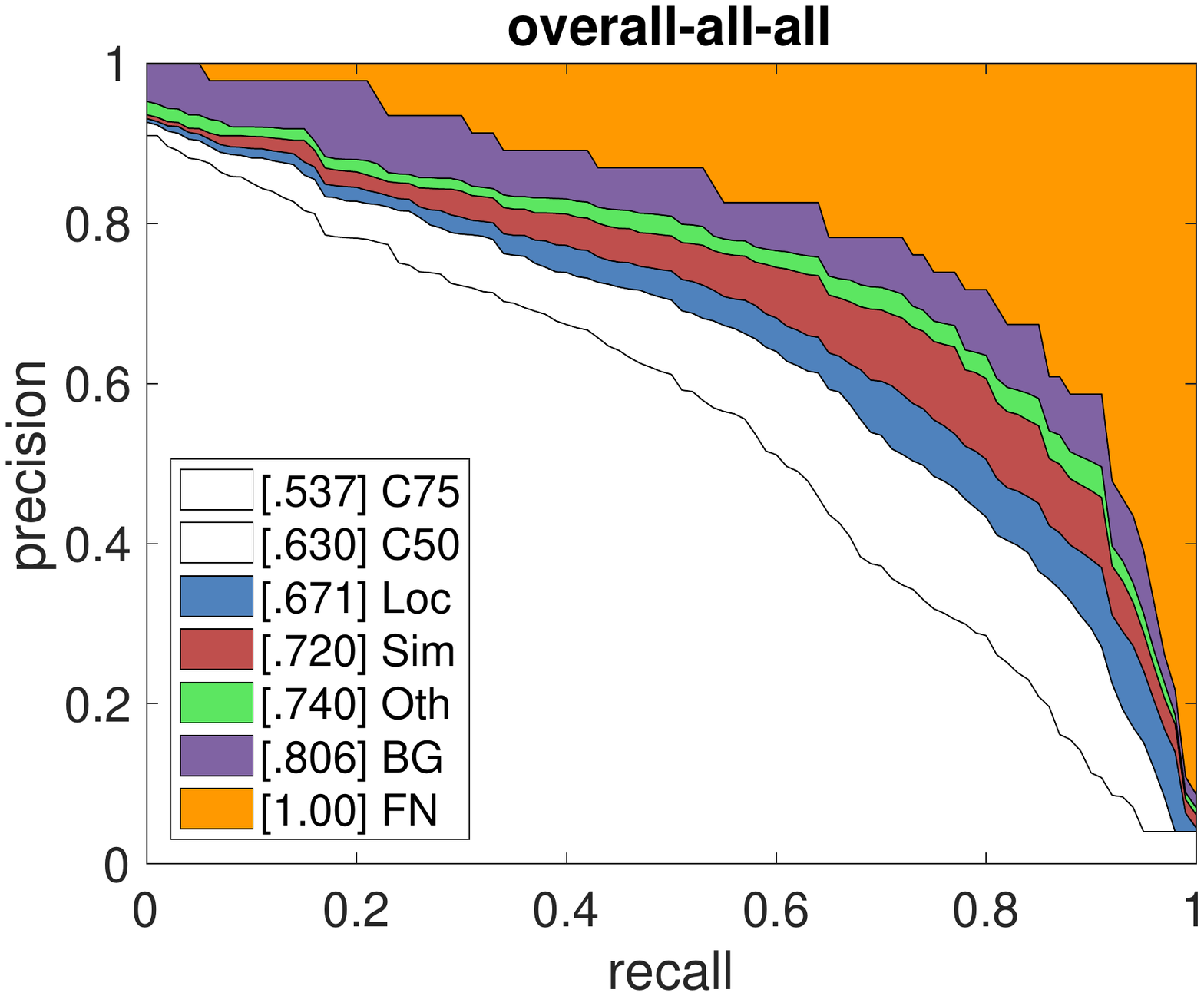}
    \label{fig:fp_overall}
}
\hfill
\subfigure[COCO (2015 challenge winner)]{
    \includegraphics[scale=0.3, clip=true, trim = 13mm 60mm 25mm 60mm]{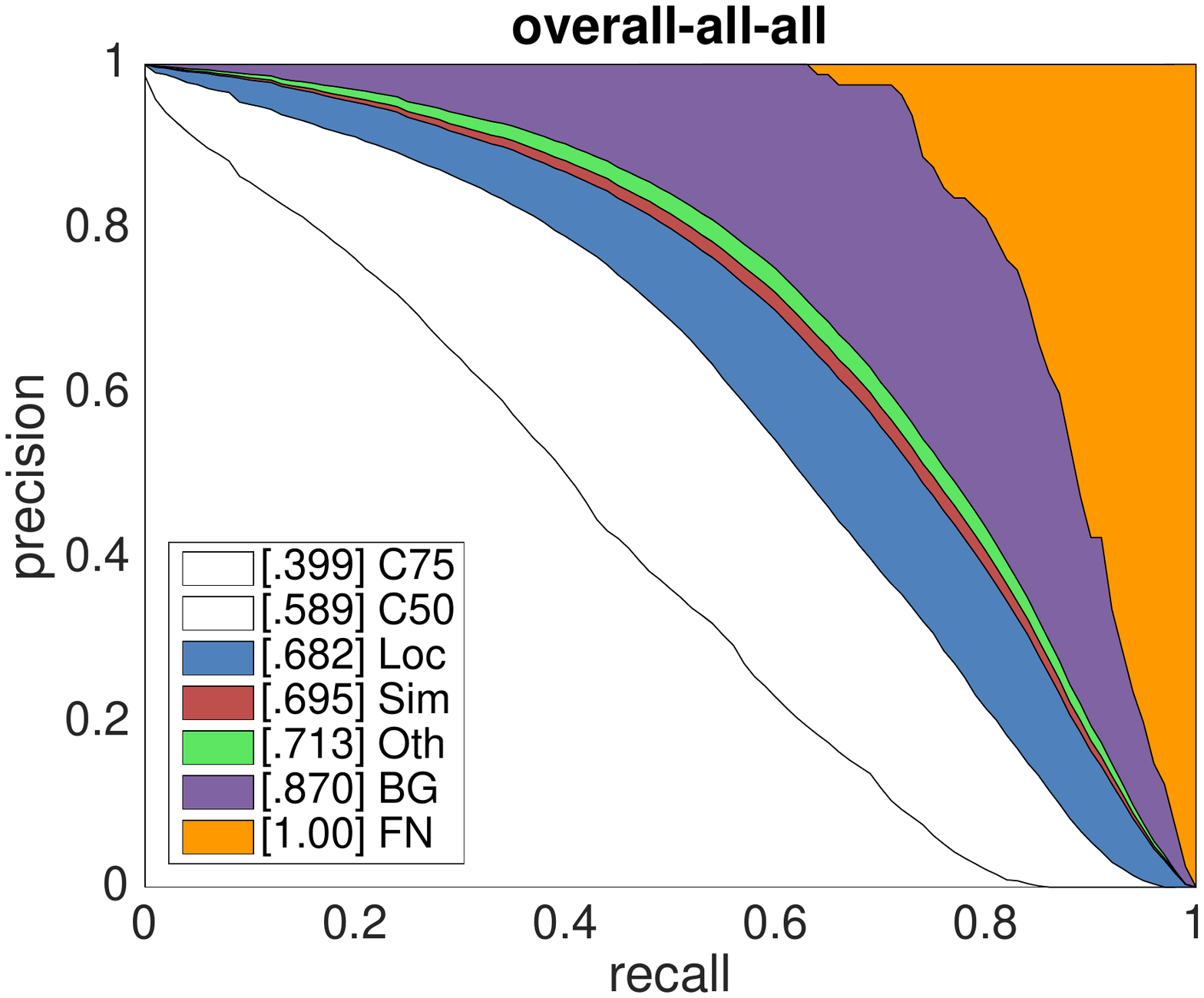}
    \label{fig:coco_overall}
}
\caption{
Main apparel detectors analysis.
Each plot shows 7 precision recall curves where each evaluation setting is more permissive than the previous.
Specifically,
\textbf{C75}: strict IoU ($\tau_{\text{IoU}}=0.75$);
\textbf{C50}: PASCAL IoU ($\tau_{\text{IoU}}=0.5$);
\textbf{Loc}: localization errors ignored ($\tau_{\text{IoU}}=0.1$);
\textbf{Sim}: supercategory False Positives (FPs) removed;
\textbf{Oth}: category FPs removed;
\textbf{BG}: background (and class confusion) FPs removed;
\textbf{FN}: False Negatives are removed.
Two plots are a comparison between two detectors trained on Fashionpedia and COCO respectively. The results are averaged over all categories.
Legends present the area under each curve (corresponds to AP metric) in brackets as well}
\label{supfig:analysis}
\end{figure*}

\subsection{Results Discussion}
\textbf{Attribute-Mask R-CNN.} 
From results in Table~\ref{tab:baseline}, we have the following observations:
(1) Our baseline models achieve promising performance on challenging Fashionpedia dataset. 
(2) There is a significant drop (\eg, from 48.7 to 35.7 for SpineNet-143) in box AP if we add $\tau_{\text{F}_1}$ as another constraint for true positive. 
This is further verified by per super-category mask results in Table~\ref{tab:per_cls_results}.
This suggests that joint instance segmentation and attribute localization is a significantly more difficult task than instance segmentation, leaving much more room for future improvements.

\textbf{Main apparel detection analysis.}
We also provide in-depth detector analysis following COCO detection challenge evaluation~\cite{lin_microsoft_2014} inspired by Hoiem~\etal~\cite{hoiem2012diagnosing}.
Figure~\ref{supfig:analysis} illustrates a detailed breakdown of bounding boxes false positives produced by the detectors.

Figure~\ref{fig:fp_overall} and ~\ref{fig:coco_overall} compare two detectors trained on Fashionpedia and COCO.
Errors of the COCO detector are dominated by imperfect localization (AP is increased by $28.3$ from overall AP at $\tau_{\text{IoU}}=0.75$) and background confusion ($+15.7$) (\ref{fig:coco_overall}).
Unlike the COCO detector, no mistake in particular dominates the errors produced by the Fashionpedia detector.
Figure~\ref{fig:fp_overall} shows that there are errors from localization ($+13.4$), classification ($+6.9$), background confusions ($+6.6$).
Due to the space constraint, we leave super-category analysis in the supplementary material.

\begin{figure*}
  \begin{center}
  \includegraphics[width=0.95\textwidth]{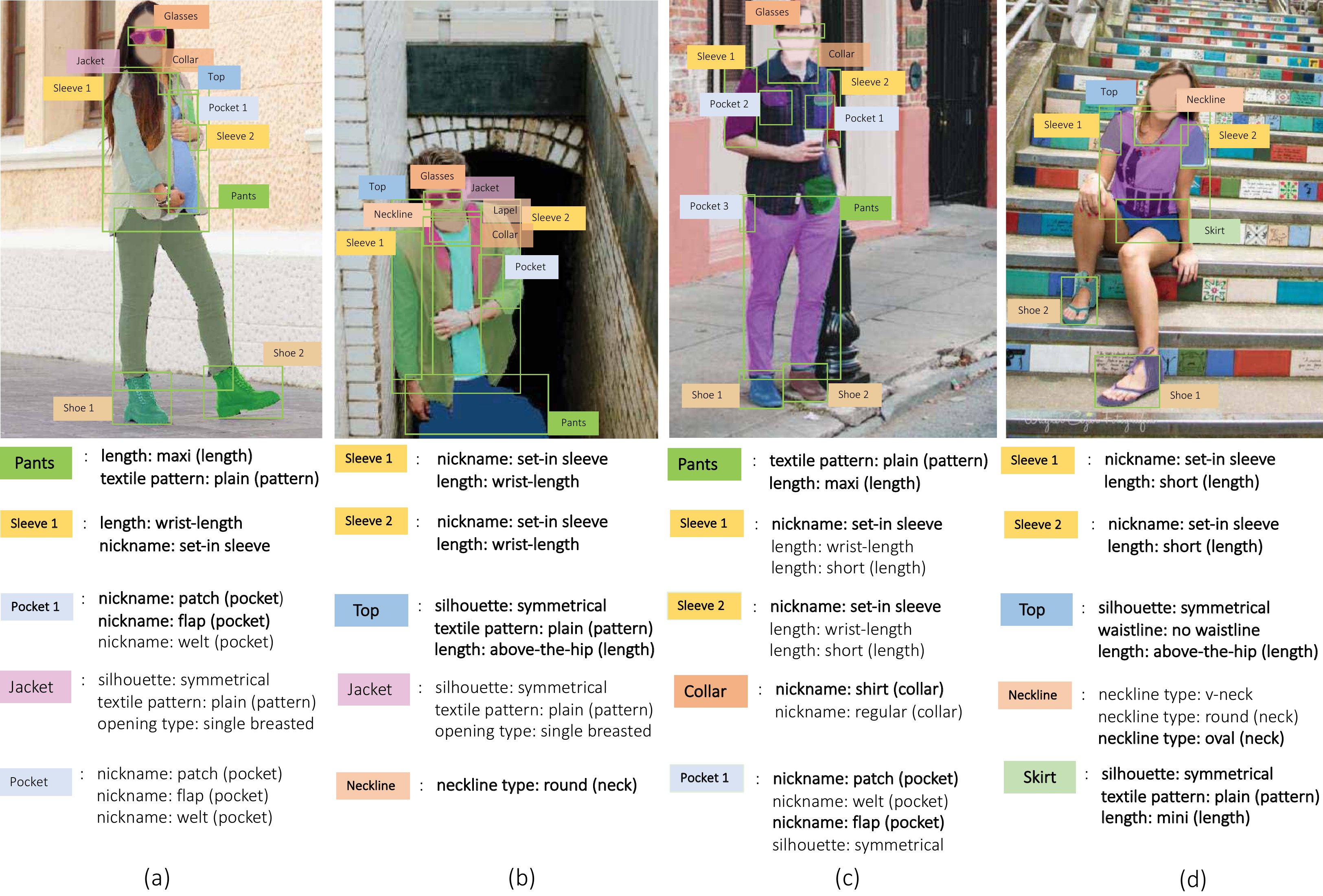}
  \includegraphics[width=0.95\textwidth]{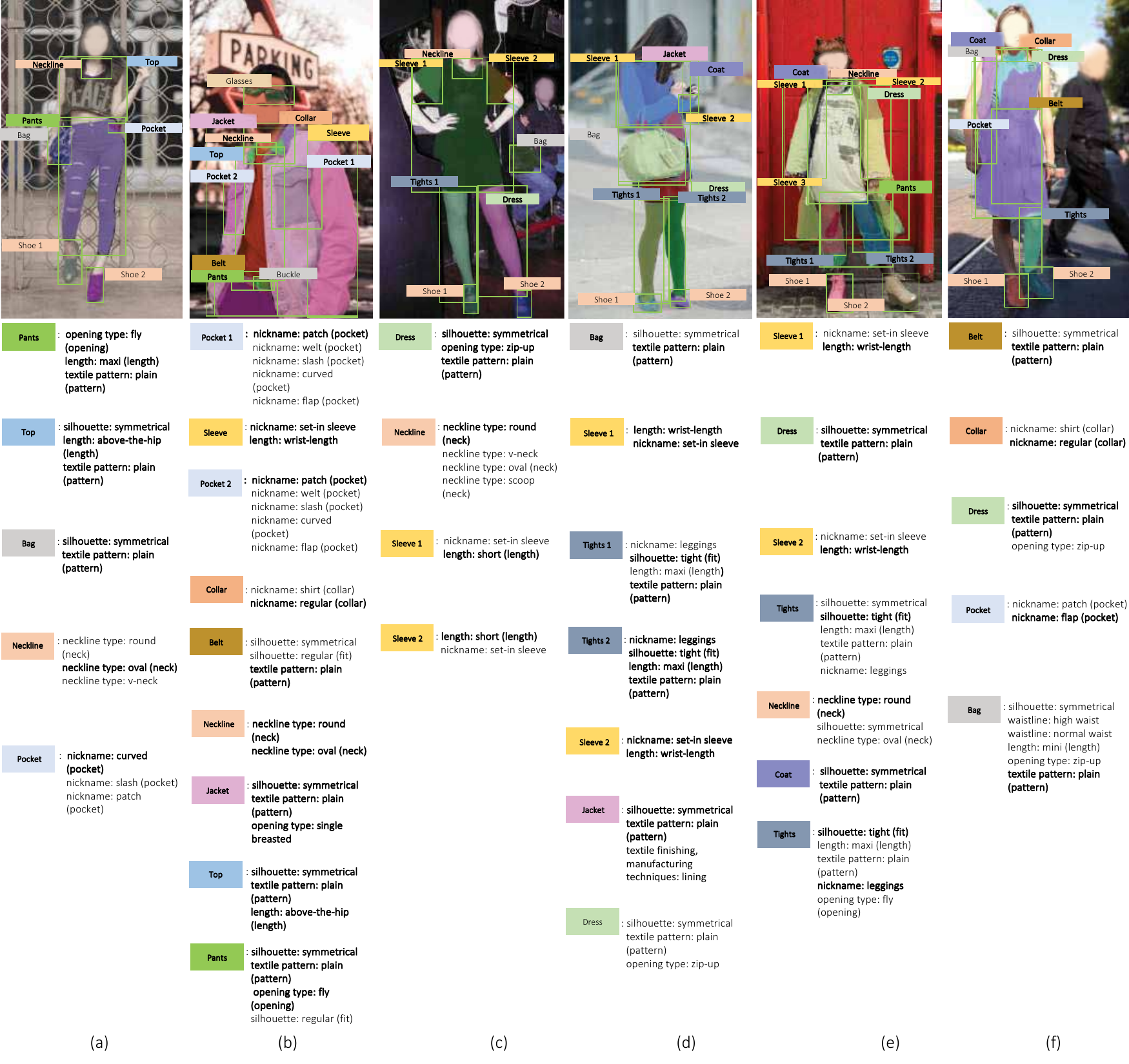}
  \end{center}
  \caption{Attribute-Mask R-CNN results on the Fashionpedia validation set. Masks, bounding boxes, and apparel categories (category score $> 0.6$) are shown. The localized attributes from the top 5 masks (that contain attributes) on each image are also shown. Correctly predicted categories and localized attributes are bolded}
  \label{fig:results_example}
\end{figure*}

\textbf{Prediction visualization.} Baseline outputs (with both segmentation masks and localized attributes) are also visualized in Figure~\ref{fig:results_example}.
Our Attribute-Mask R-CNN achieves good results even for small objects like shoes and glasses.
Model can correctly predict fine-grained attributes for some masks on one hand (e.g., Figure~\ref{fig:results_example} second image at top row~\includegraphics[height=.025\textwidth]{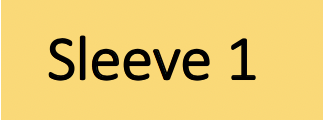}).
On the other hand, it also incorrectly predict the wrong nickname (welt) to pocket (e.g., Figure~\ref{fig:results_example} third image at top row~\includegraphics[height=.025\textwidth]{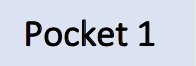}).
These results show that there is headroom remaining for future development of more advanced computer vision models on this task (see supplementary material for more details of the baseline analysis).

\section{Conclusion}
In this work, we focus on a new task that unifies instance segmentation and attribute recognition.
To solve challenging problems entailed in this task, we introduced the Fashionpedia ontology and dataset.
To the best of our knowledge, Fashionpedia is the first dataset that combines part-level segmentation masks with fine-grained attributes. 
We presented Attribute-Mask R-CNN, a novel model for this task, along with a novel evaluation metric. 
We expect models trained on Fashionpedia can be applied to many applications including better product recommendation in online shopping, enhanced visual search results, and resolving ambiguous fashion-related words for text queries. 
We hope Fashionpedia will contribute to the advances in fine-grained image understanding in the fashion domain.

\section{Acknowledgements}
\label{sec:sup_acklo}
This research was partially supported by a Google Faculty Research Award.
We thank Kavita Bala, Carla Gomes, Dustin Hwang, Rohun Tripathi, Omid Poursaeed, Hector Liu, and Nayanathara Palanivel, Konstantin Lopuhin for their helpful feedback and discussion in the development of Fashionpedia dataset. We also thank Zeqi Gu, Fisher Yu, Wenqi Xian, Chao Suo, Junwen Bai, Paul Upchurch, Anmol Kabra, and Brendan Rappazzo for their help developing the fine-grained attribute annotation tool.

\section{Supplementary Material}
\label{sec:sup}

In our work we presented the new task of instance segmentation with attribute localization.
We introduced a new ontology and dataset, Fashionpedia, to further describe the various aspects of this task.
We also proposed a novel evaluation metric and Attribute-Mask R-CNN model for this task.
In the supplemental material, we provide the following items that shed further insight on these contributions:
\begin{itemize}
    \item More comprehensive experimental results (\S~\ref{sec:supp_model})
    \item An extended discussion of Fashionpedia ontology and potential knowledge graph applications (\S~\ref{sec:supp_ontology})
    \item More details of dataset analysis (\S~\ref{sec:supp_data})
    \item Additional information of annotation process (\S~\ref{sec:supp_anno})
    \item Discussion (\S~\ref{sec:supp_mis})
\end{itemize}

\subsection{Attribute-Mask R-CNN}
\label{sec:supp_model}

\begin{figure}
  \begin{center}
  \includegraphics[width=\textwidth]{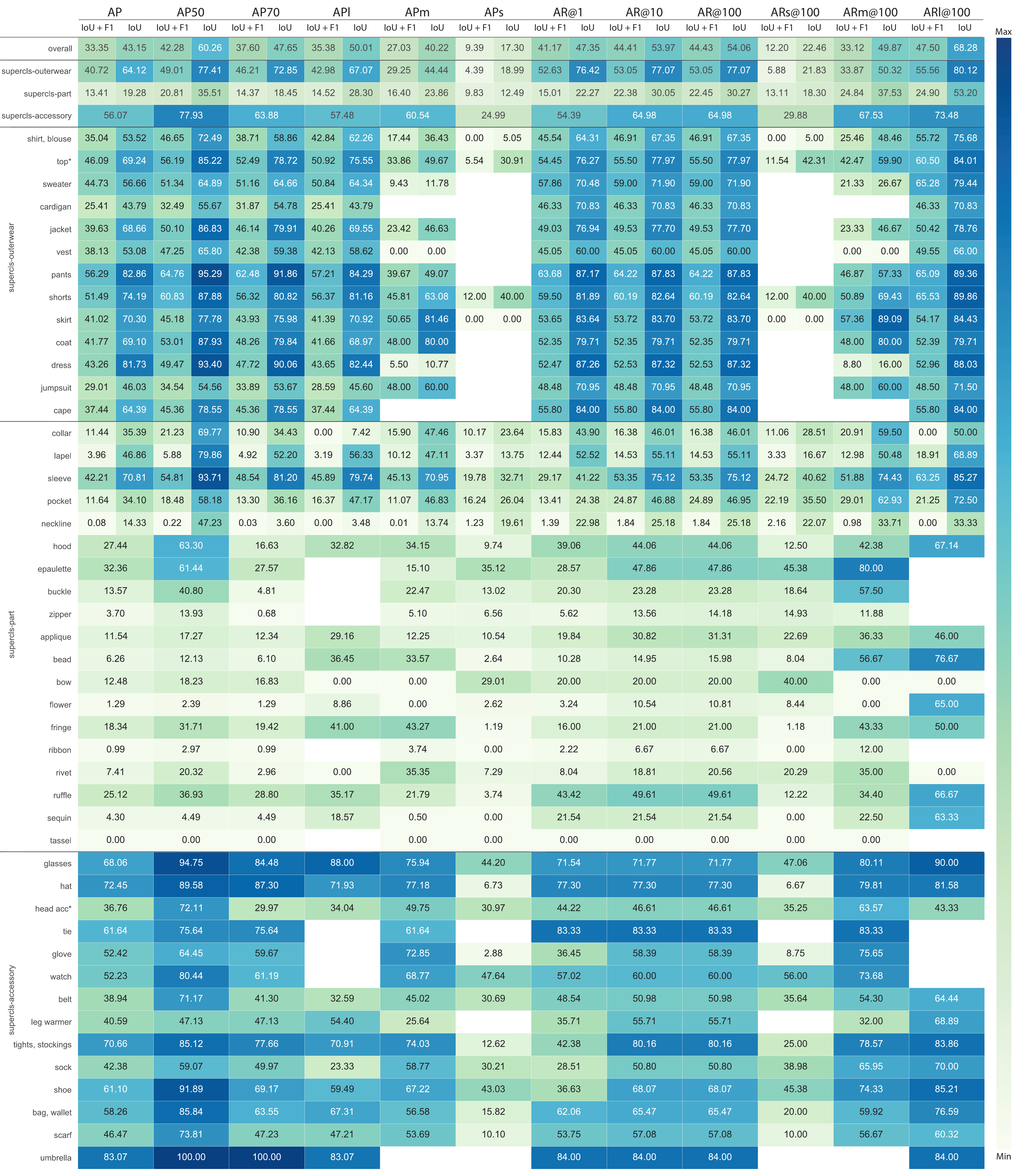}
  \end{center}
  \caption{Detailed results (for masks) using Mask R-CNN with SpineNet-143 backbone. We present the same metrics as COCO leaderboard for overall categories, three super categories for apparel objects, and 46 fine-grained apparel categories. We use both constraints (for example, AP$_{\text{IoU}}$ and AP$_{\text{IoU}+\text{F}_1}$) if possible. For categories without attributes, the value represents AP$_{\text{IoU}}$ or AR$_{\text{IoU}}$. ``top\*'' is short for ``top, t-shirt, sweatshirt''. ``head acc\*'' is short for ``headband, head covering, hair accessory''}
  \label{supfig:ap1}
\end{figure}

\textbf{Per-class evaluation.} 
Fig.~\ref{supfig:ap1} presents detailed evaluation results per supercategory and per category.
In Fig.~\ref{supfig:ap1}, we  follow  the  same  metrics from  COCO  leaderboard (AP,  AP50,  AP75,  APl,  APm,  APs, AR@1, AR@10, AR@100, ARs@100, ARm@100, ARl@100), 
with $\tau_{IoU}$ and $\tau_{F1}$ if possible.
Fig.~\ref{supfig:ap1} shows that metrics considering both constraint $\tau_{IoU}$ and $\tau_{F1}$ are always lower than using $\tau_{IoU}$ alone across all the supercategories and categories. 
This further demonstrates the challenging aspect of our proposed task.

In general, categories belong to ``garment parts'' have a lower AP and AR, comparing with ``outerwear'' and ``accessories''.

\begin{figure}
\centering
\subfigure[\scriptsize{Super category: outerwear}]{
    \includegraphics[scale=0.2, clip=true, trim = 15mm 60mm 25mm 60mm]{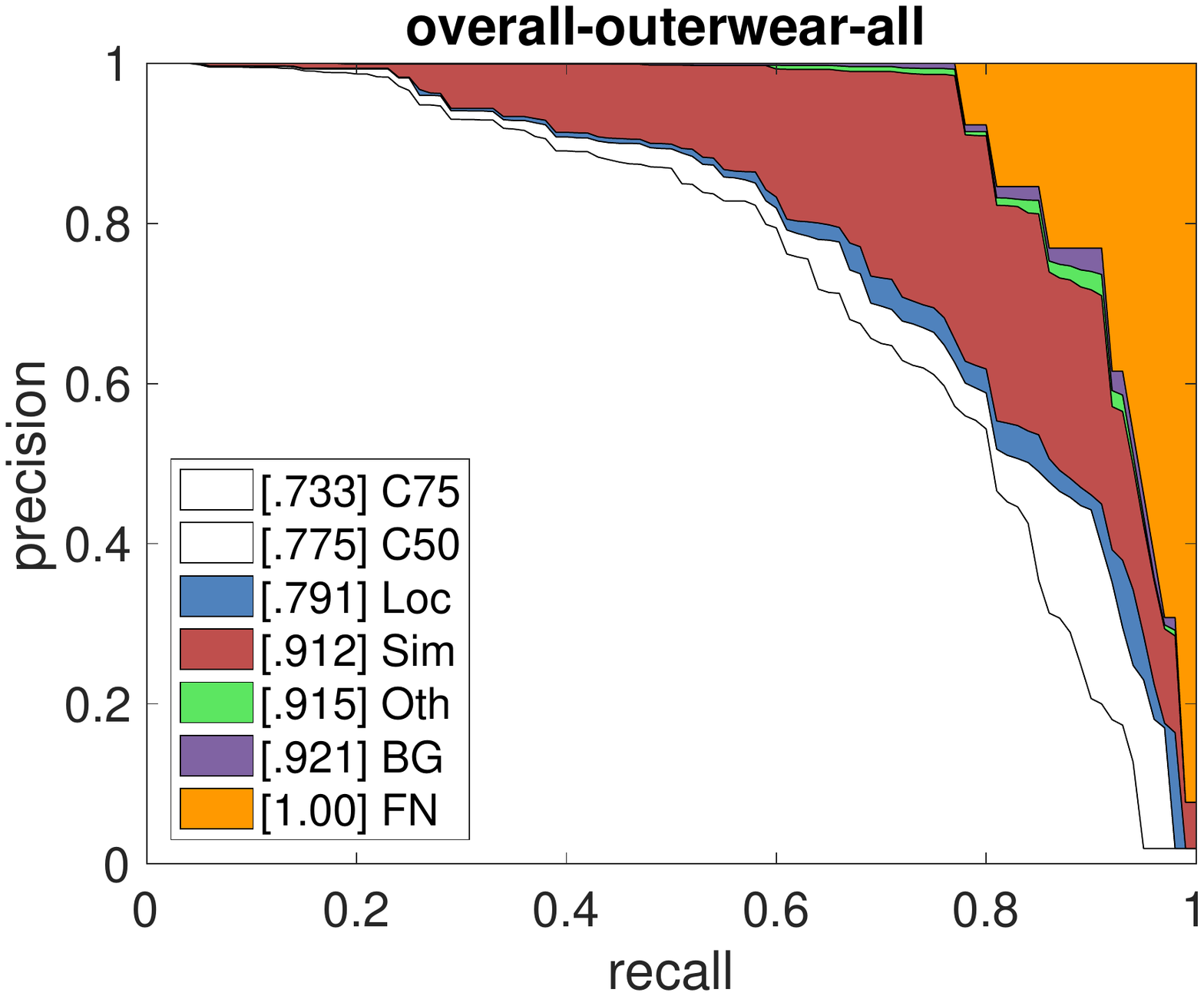}
    \label{supfig:outer_all}
}
\hfill
\subfigure[\scriptsize{Super category: parts}]{
    \includegraphics[scale=0.2, clip=true, trim = 15mm 60mm 25mm 60mm]{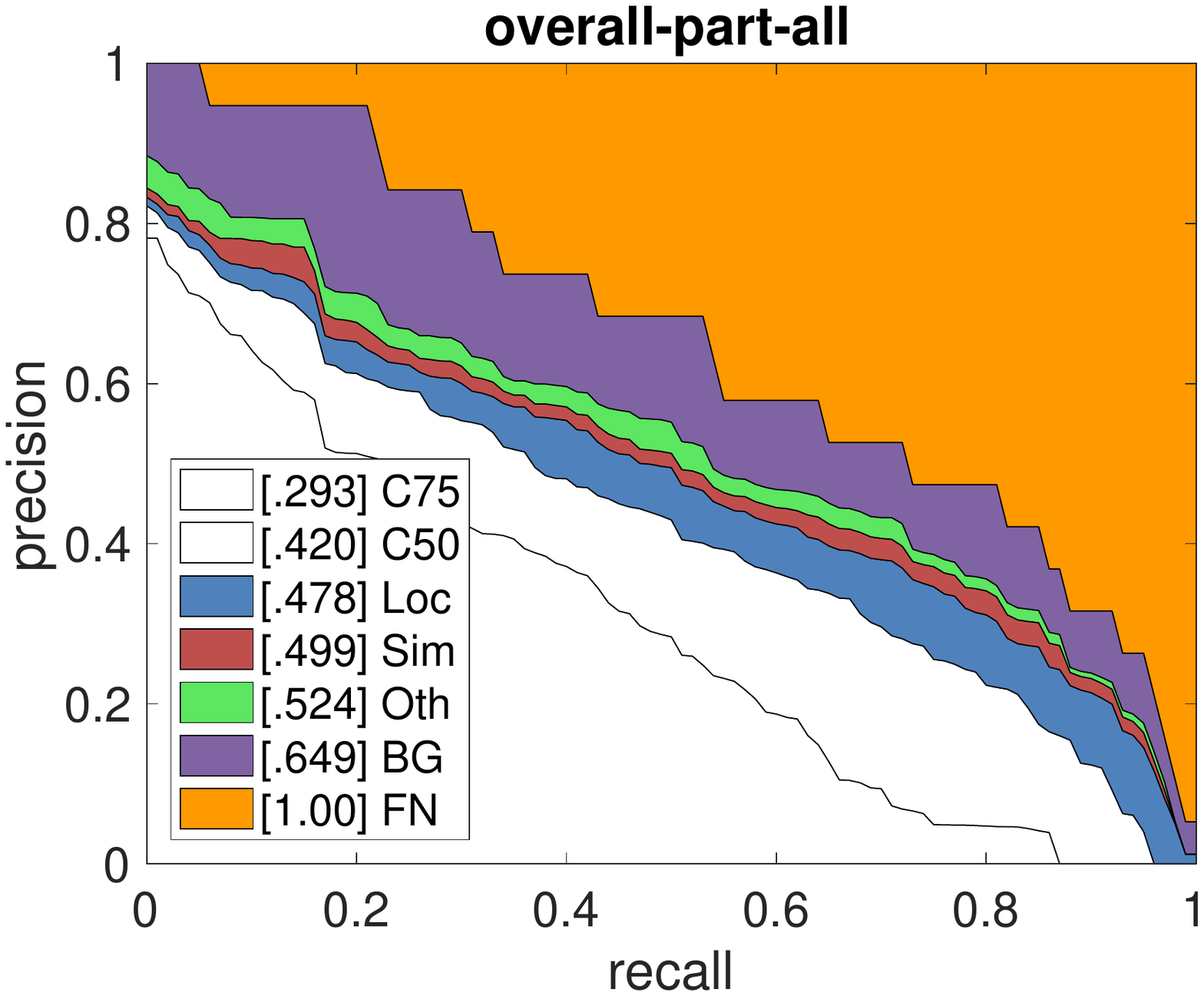}
    \label{supfig:part_all}
}
\hfill
\subfigure[\scriptsize{Super category: accessory}]{
    \includegraphics[scale=0.2, clip=true, trim = 15mm 60mm 25mm 60mm]{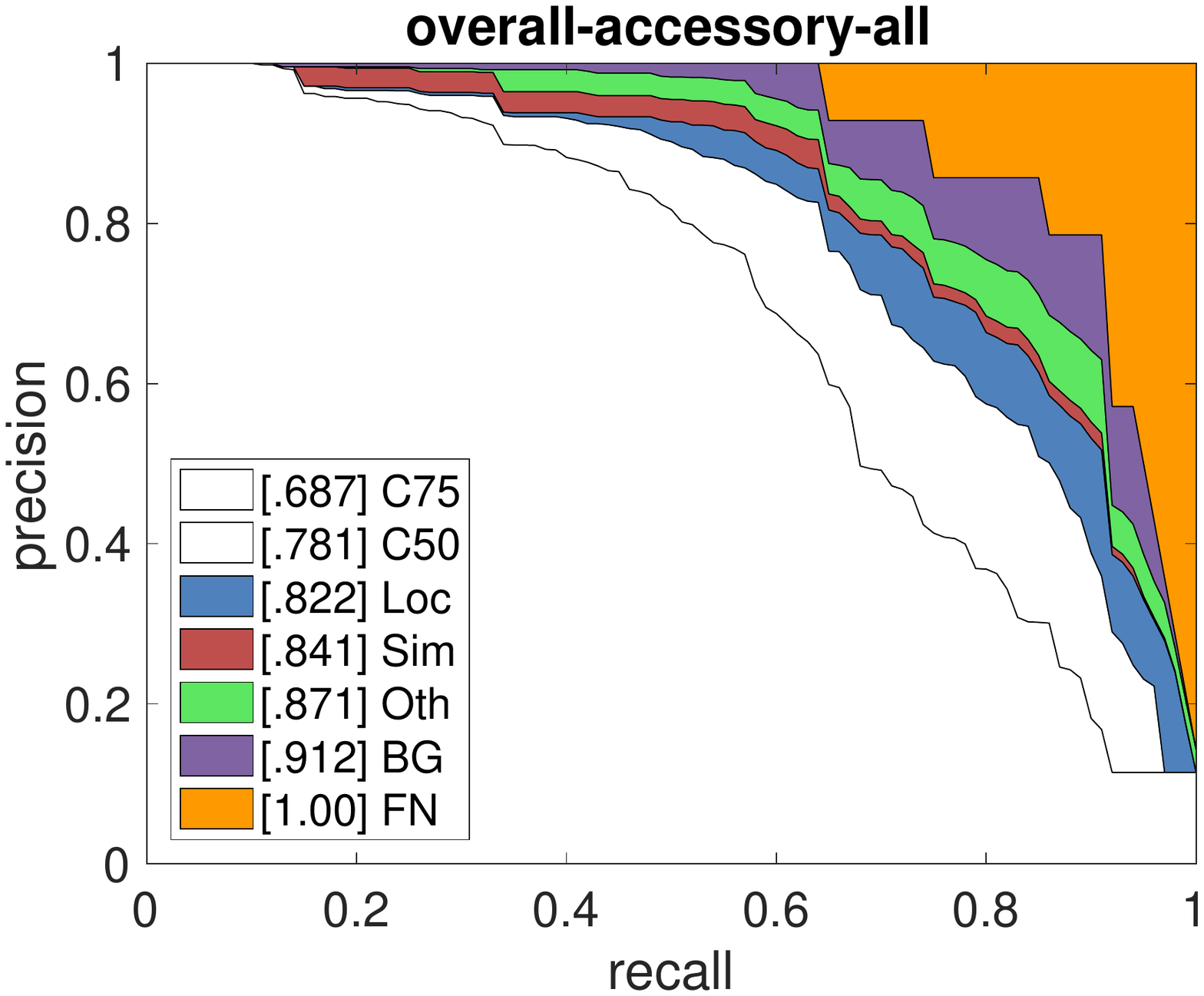}
    \label{supfig:acc_all}
}
\hfill
\subfigure[\scriptsize{Outerwear: skirt}]{
    \includegraphics[scale=0.2, clip=true, trim = 15mm 60mm 25mm 60mm]{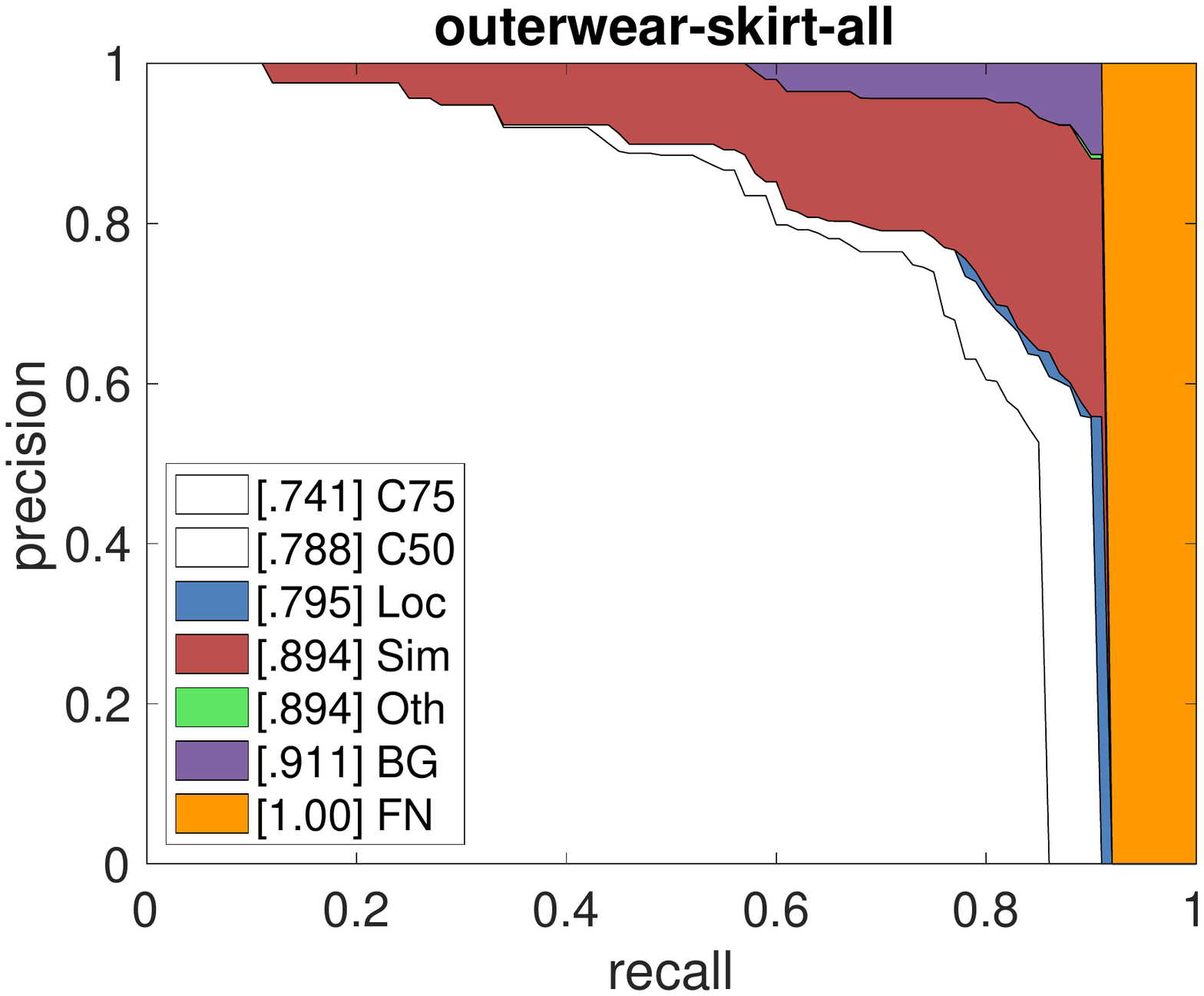}
    \label{supfig:outer_skirt}
}
\hfill
\subfigure[\scriptsize{Parts: pocket}]{
    \includegraphics[scale=0.2, clip=true, trim = 15mm 60mm 25mm 60mm]{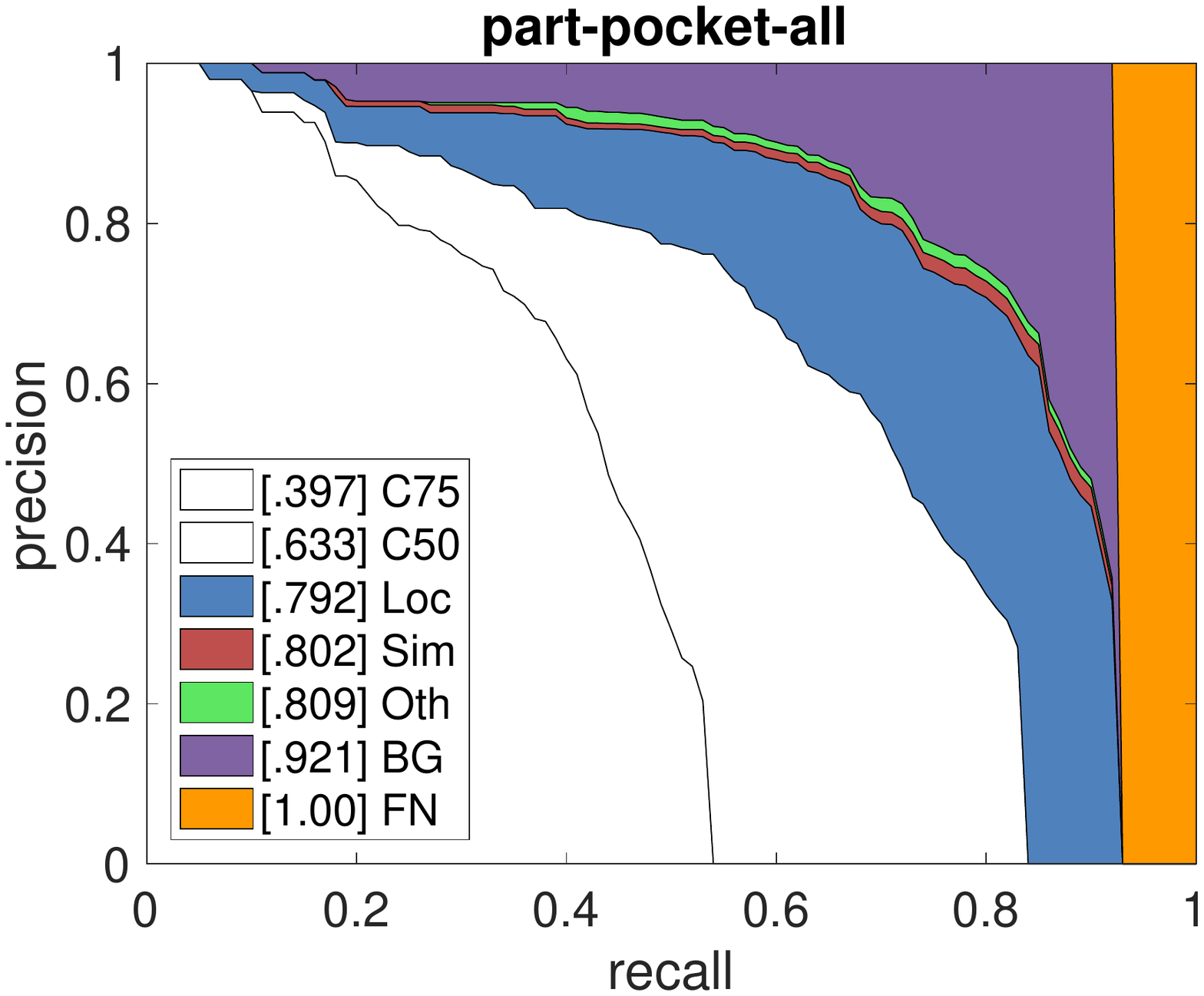}
    \label{supfig:part_zipper}
}
\hfill
\subfigure[\scriptsize{Accessory: tights, stockings}]{
    \includegraphics[scale=0.2, clip=true, trim = 15mm 60mm 25mm 60mm]{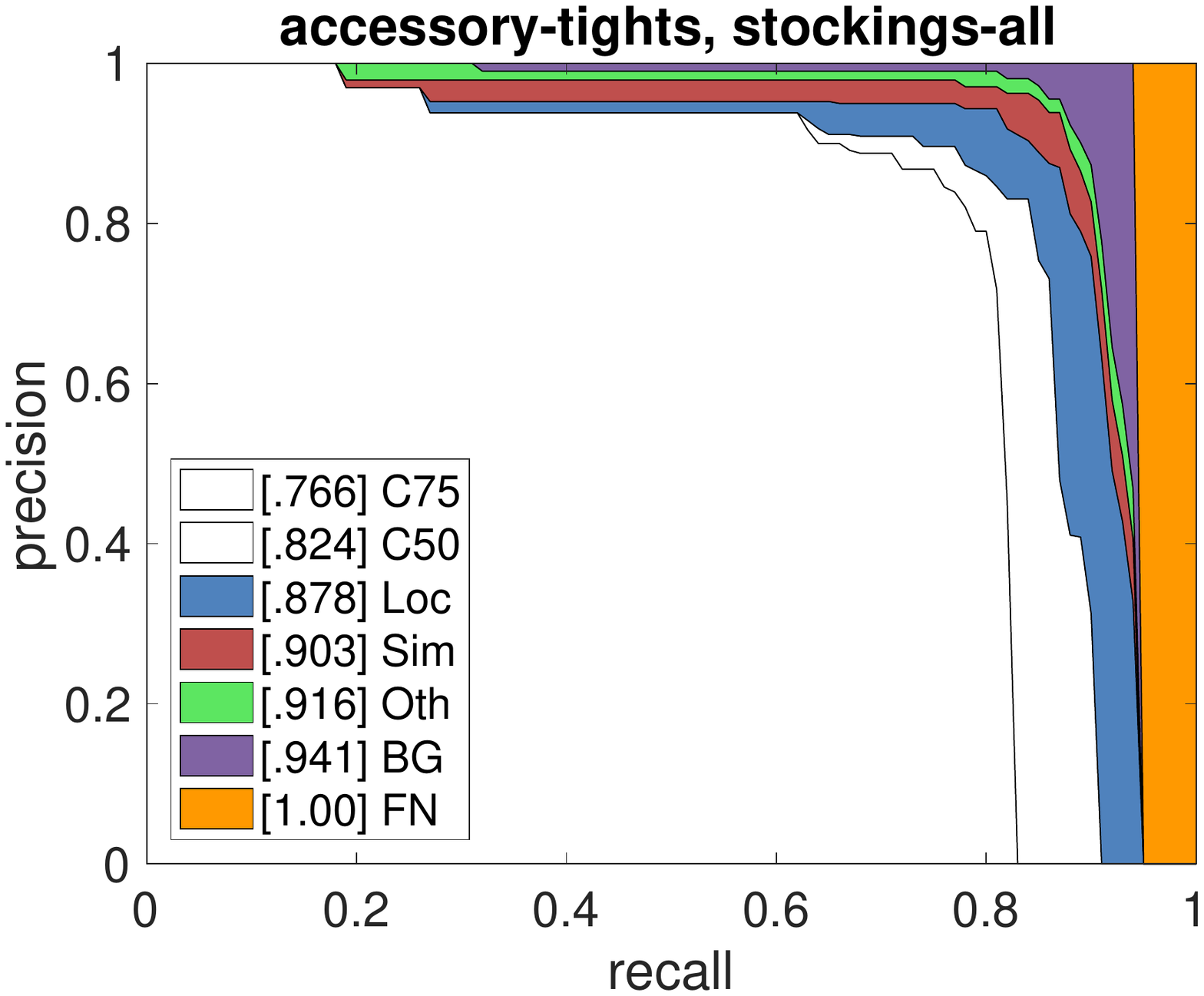}
    \label{supfig:acc_tights}
}
\caption{
Main apparel detectors analysis.
Each plot shows 7 precision recall curves where each evaluation setting is more permissive than the previous.
Specifically,
\textbf{C75}: strict IoU ($\tau_{\text{IoU}}=0.75$);
\textbf{C50}: PASCAL IoU ($\tau_{\text{IoU}}=0.5$);
\textbf{Loc}: localization errors ignored ($\tau_{\text{IoU}}=0.1$);
\textbf{Sim}: supercategory False Positives (FPs) removed;
\textbf{Oth}: category FPs removed;
\textbf{BG}: background (and class confusion) FPs removed;
\textbf{FN}: False Negatives are removed.
The first row \textit{(overall-[supercategory]-[size])}
contains results for three supercategories in Fashionpedia; the second row \textit{([supercategory]-[category]-[size])} shows results for three fine-grained categories (one per supercategory). Legends present the area under each curve (corresponds to AP metric) in brackets as well}
\label{fig:supp_detectors}
\end{figure}

A detailed breakdown of detection errors is presented in Fig.~\ref{fig:supp_detectors} for supercategories and three main categories.
In terms of supercategories in Fashionpedia, ``outerwear'' errors are dominated by within supercategory class confusions (Fig.~\ref{supfig:outer_all}).
Within this supercategory class, ignoring  localization  errors  would  only  raise  AP  slightly  from  $77.5$  to  $79.1$  ($+1.6$).
A  similar  trend  can  be  observed in class ``skirt'',  which belongs to ``outerwear'' (Fig.~\ref{supfig:outer_skirt}).
Detection errors of ``part'' (Fig.~\ref{supfig:part_all}~\ref{supfig:part_zipper}) and ``accessory''(Fig.~\ref{supfig:acc_all}~\ref{supfig:acc_tights}) on the other hand, are dominated by both background confusion and localization. ``part'' also has a lower AP in general, compared with other two super-categories.
A possible reason is that objects belong to ``part'' usually have smaller sizes and lower counts.

\begin{figure}
  \begin{center}
  \includegraphics[width=0.6\textwidth]{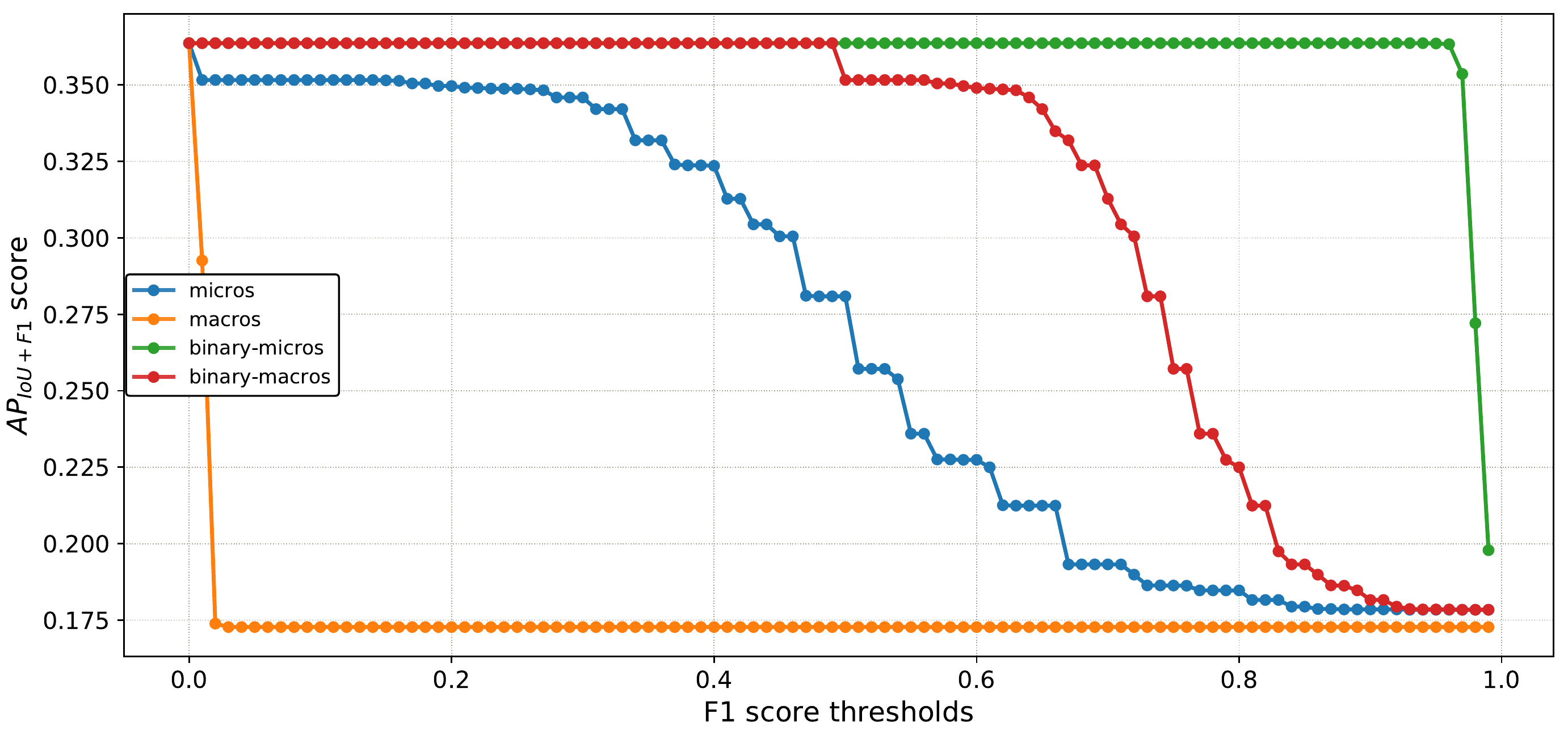}
  \end{center}
  \caption{$AP_{IoU+F1}^{F1=\tau_{F1}}$ score with different $\tau_{F1}$. The value presented are average over $\tau_{IoU} \in [0.5:0.05:0.95]$. We use ``binary-macro'' as our main metric}
  \label{supfig:f1_types}
\end{figure}

\textbf{F1 score calculation.} 
Since we measure the f1 score of predicted attributes and groundtruth attributes per mask, we consider the both options of multi-label multi-class classification with 294 classes for one instance, and binary classification for 294 instances.
Multi-label multi-class classification is a straightforward task, as it is a common setting for most of the fine-grained classification tasks.
In binary classification scenario, we consider the $1$ and $0$ of the multi-hot encoding of both results and ground-truth labels as the positive and negative classes respectively. 
There are also two averaging choices: ``micro'' and ``macro''.
``Micro'' averaging calculates the score globally by counting the total true positives, false negatives and false positives.
``Macro'' averaging calculates the metrics for each attribute class and reports the unweighted mean.
In sum, there are four options of f1-score averaging methods: 1) ``micro'', 2) ``macro'', 3) ``binary-micro'', 4) ``binary-macro''.

As shown in Fig.~\ref{supfig:f1_types}, we present the $AP_{IoU+F1}^{F1=\tau_{F1}}$, with $\tau_{IoU}$ averaged in the range of $[0.5:0.05:0.95]$. $\tau_{F1}$ is increased from $0.0$ to $1.0$ with a increment of $0.01$.
Fig.~\ref{supfig:f1_types} illustrates that as the value of $\tau_{F1}$ increases, $AP_{IoU+F1}^{F1=\tau_{F1}}$ decreases in different rates given different choices of f1 score calculation.
There are 294 attributes in total, and an average of 3.7 attributes per mask in Fashionpedia training data.
It's not surprising to observe that``Binary-micro'' produces high f1-scores in general (higher than $0.97$), as the $AP_{IoU+F1}$ score only decreases if the $\tau_{F1} \geq 0.97$.
On the other hand, ``macro'' averaging in multi-label multi-class classification scenario gives us extremely low f1-scores ($0.01 - 0.03$). This further demonstrates the room for improvement for localized attribute classification task.
We used ``binary-macros'' as our main metric.

\begin{figure}[t]
  \begin{center}
  \includegraphics[width=0.9\textwidth]{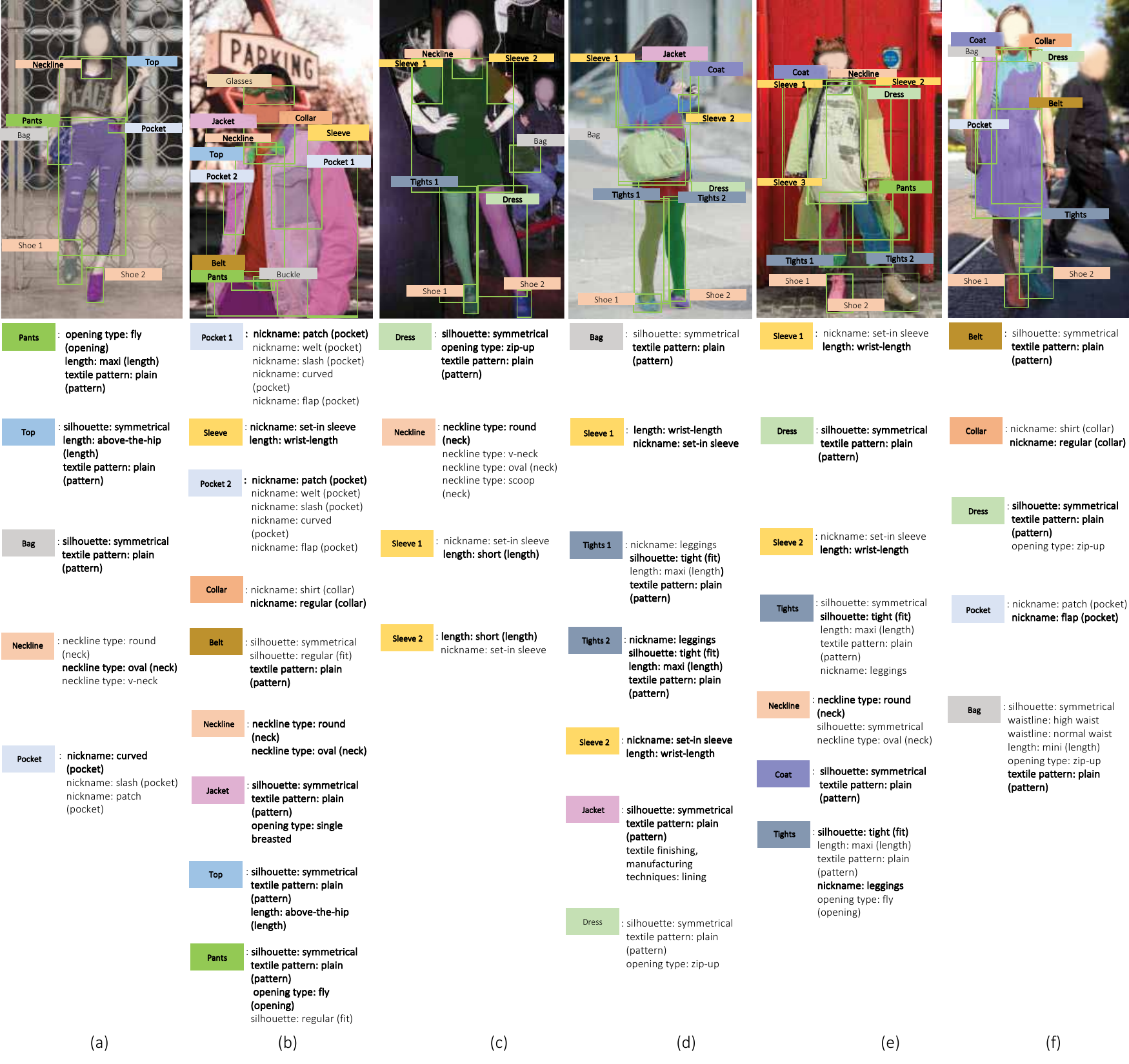}
  \end{center}
  \caption{Baseline results on the Fashionpedia validation set. Masks, bounding boxes, and apparel categories (category score $> 0.6$) are shown. Attributes from top 10 masks (that contain attributes) from each image are also shown. Correct predictions of objects and attributes are bolded}
  \label{supfig:predictionexamples}
\end{figure}

\textbf{Result visualization.}
Figure~\ref{supfig:predictionexamples} shows that our simple baseline model can detect most of the apparel categories correctly.
However, it also produces false positives sometimes.
For example, it segments legs as tights and stockings (Figure~\ref{supfig:predictionexamples}(f)~\includegraphics[height=.025\textwidth]{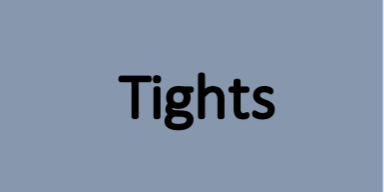}).
A possible reason is that both objects have the same shape and stockings are worn on the legs.

Predicting fine-grained attributes, on the other hand, is a more challenging problem for the baseline model.
We summarize several issues:
(1) predict more attributes than needed: (Figure~\ref{supfig:predictionexamples}(a)~\includegraphics[height=.025\textwidth]{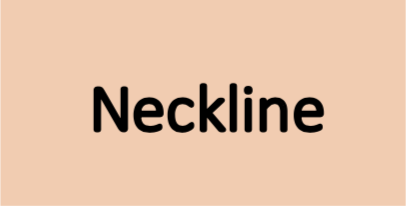}, (b)~\includegraphics[height=.025\textwidth]{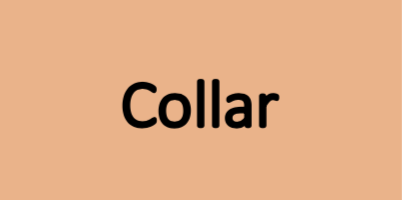}, (c)~\includegraphics[height=.025\textwidth]{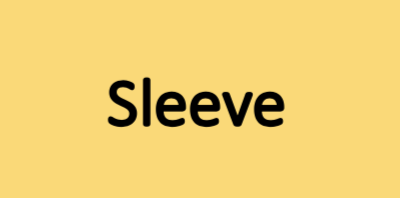});
(2) fail to distinguish among fine-grained attributes: for example, dropped-shoulder sleeve (ground truth) v.s. set-in sleeve (predicted) (Figure~\ref{supfig:predictionexamples}(e)~\includegraphics[height=.025\textwidth]{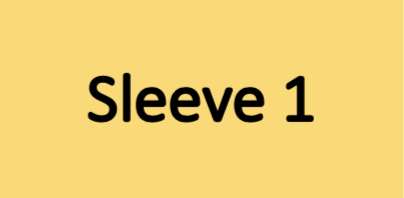}); 
(3) other false positives: Figure~\ref{supfig:predictionexamples}(e)~\includegraphics[height=.025\textwidth]{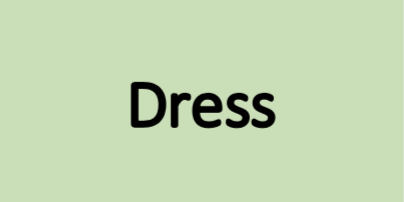} has a double-breasted opening, yet the model predicted it as the zip opening.

These results further show that there are rooms for improvement and future development of more advanced computer vision models on this instance segmentation with attribute localization task.

\begin{figure}
  \begin{center}
  \includegraphics[width=0.9\textwidth]{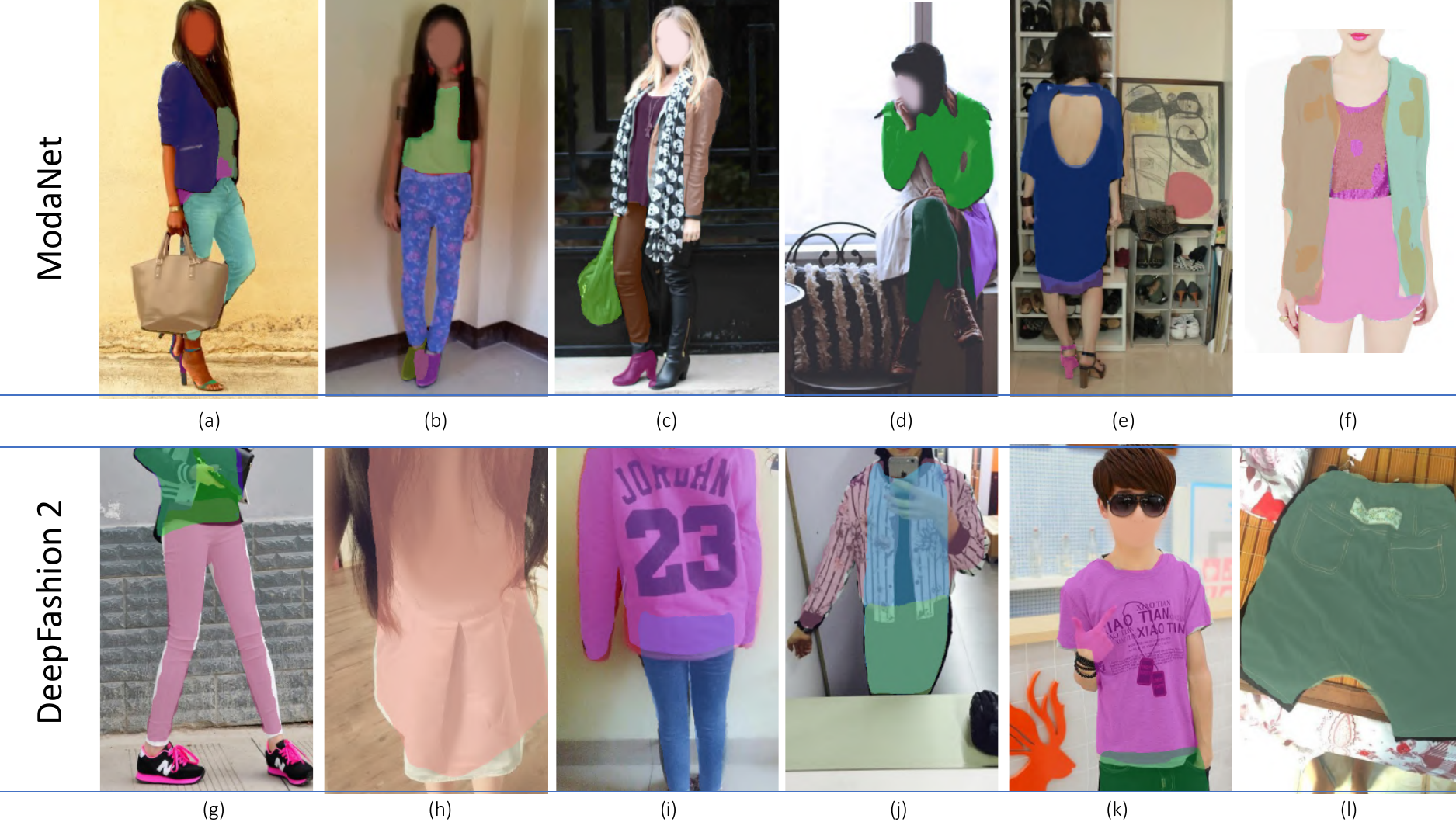}
  \end{center}
  \caption{Baseline results on ModaNet and DeepFashion2 validation set}
  \label{supfig:modanet_df2}
\end{figure}

\begin{figure}
  \begin{center}
  \includegraphics[width=0.8\textwidth]{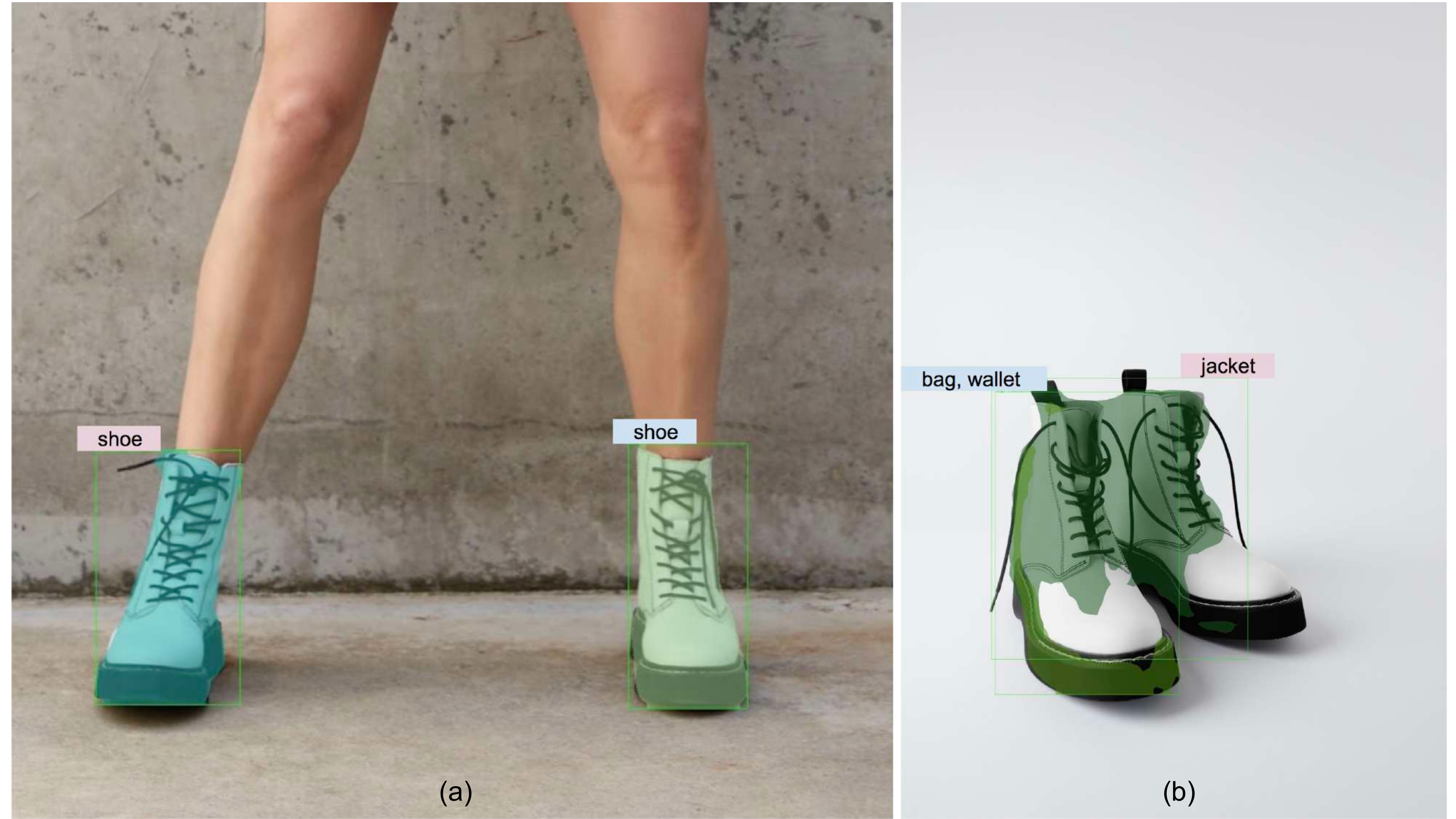}
  \end{center}
  \caption{Generated masks on online-shopping images~\cite{zaraimages}. (a) and (b) show the same types of shoes in different settings. Our model correctly detects and categorizes the pair of shoes worn by a fashion model, yet mistakenly detects shoes as jacket and a bag in (b)}
  \label{supfig:image_tested_on_no_human_body}
\end{figure}

\textbf{Result visualization on other datasets.}
Other fashion datasets such as ModaNet and DeepFashion2 also contain instance segmentation masks.
Aside from the results presented in the main paper (see Table 3 of the main paper) on Fashionpedia, we present the qualitative analysis on the segmentation masks generated among 
Fashionpedia(Fig.~\ref{supfig:predictionexamples}), 
ModaNet(Fig.~\ref{supfig:modanet_df2}(a-f)) and DeepFashion2(Fig.~\ref{supfig:modanet_df2}(g-l)) datasets.
Photos of the first row in Fig.~\ref{supfig:modanet_df2} are from ModaNet.
They show that the quality of the generated masks on ModaNet is fairly good and comparable to Fashionpedia in general (Fig.~\ref{supfig:modanet_df2}(a)).
We also have a couple of observations of the failure cases:
(1) fail to detect apparel objects: for example, the shoe from Fig.~\ref{supfig:modanet_df2}(c) is not detected. Parts of the pants (Fig.~\ref{supfig:modanet_df2}(c)) and coat (Fig.~\ref{supfig:modanet_df2}(d)) are not detected;
(2) fail to detect some categories: Fig.~\ref{supfig:modanet_df2}(e) shows that the shoes on the shoe rack and right foot are not detected, possibly due to a lack of such instances in the ModaNet training dataset.
Similar to Fashionpedia, ModaNet mostly consist of street style images. See Fig.~\ref{supfig:image_tested_on_no_human_body}(b) for example predictions from model trained on Fashionpedia;
3) close-up images: ModaNet contains mostly full-body images.
This might be the possible reason to the decreased quality of predicted masks on close-up shot like  Fig.~\ref{supfig:modanet_df2}(f).

For DeepFashion 2 (Fig.~\ref{supfig:modanet_df2}(g,h,k)), the generated segmentation masks tends to not tightly follow the contours of garments in the images.
The main reason possibly is that the average number of vertices per polygon is $14.7$ for Deepfashion2, which is lower than Fashionpedia and ModaNet (see Table 2 in the main text).
Our qualitative analysis also shows that: 1) the model will generate the segmentation masks of pants (Fig.~\ref{supfig:modanet_df2}(i)) and tops (Fig.~\ref{supfig:modanet_df2}(j)) that are not visible in the images. Both of them are covered by a jacket. And we find that in DeepFashion 2, some part of the garments which is covered by other objects are indeed annotated with segmentation masks; 
2) better performance on objects that are not on human body (Fig.~\ref{supfig:modanet_df2}(l)): DeepFashion 2 contains many commercial-customer image pairs (both images with and without human body) in the training dataset. In contrast, both Fashionpedia and ModaNet contain more images with human body than images without human body in the training datasets.

\textbf{Generalization to the other image domains.}
For Fashionpedia, we also inference on images found in online shopping websites, which usually displays a single apparel category, with or without a fashion model. We found out that the learned model works reasonably well if the apparel item is worn by a model (Fig.~\ref{supfig:image_tested_on_no_human_body}).

\subsection{Fashionpedia Ontology and Knowledge Graph}
\label{sec:supp_ontology}

\begin{figure}
\centering
\subfigure[Categories\label{fig:taxo_obj}]{\includegraphics[scale=0.2, clip=true, trim = 0mm 56mm 0mm 10mm]{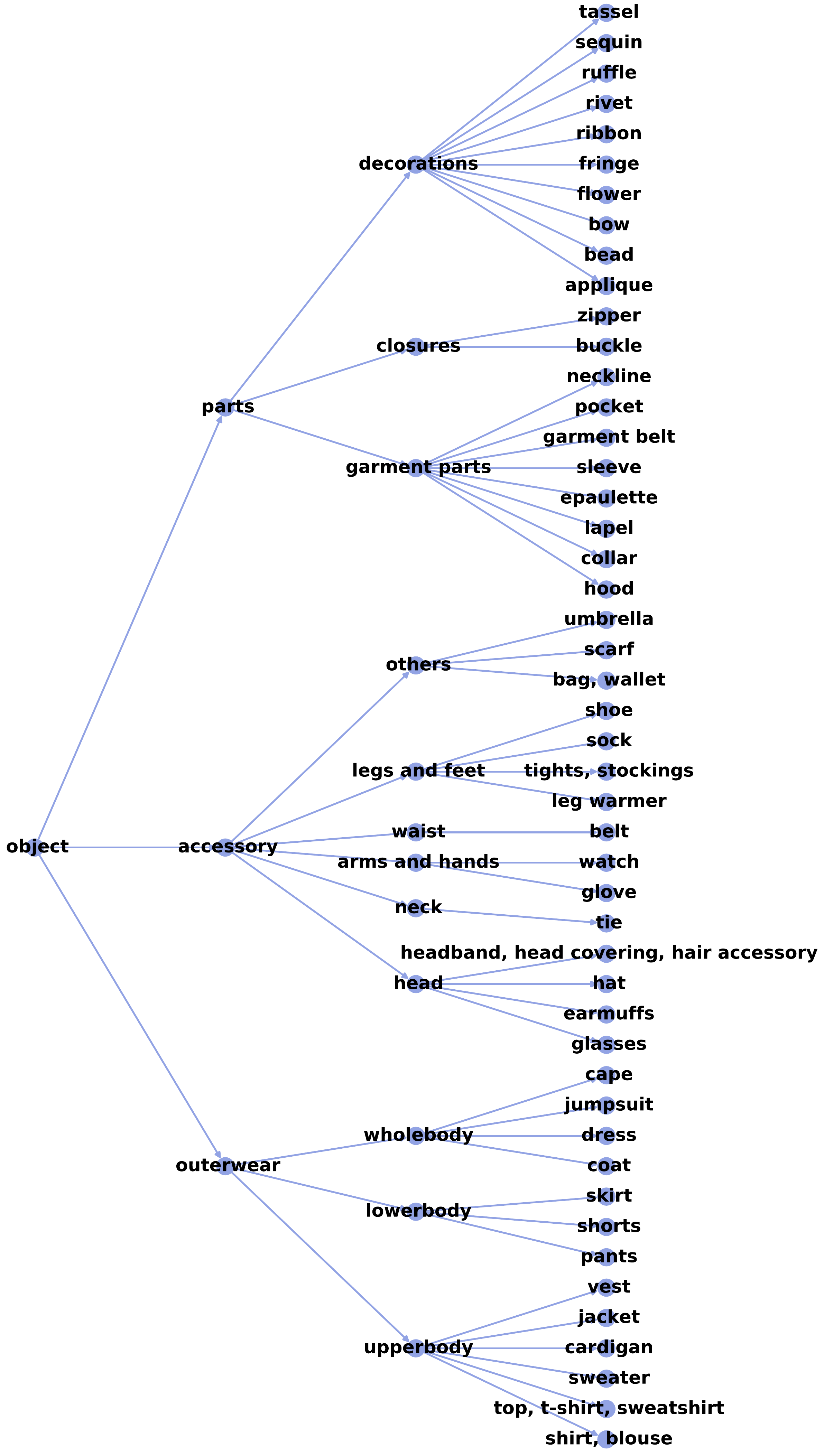}}
\subfigure[Attributes (partially)\label{fig:taxo_att}]{\includegraphics[scale=0.4, clip=true, trim = 0mm 150mm 40mm 180mm]{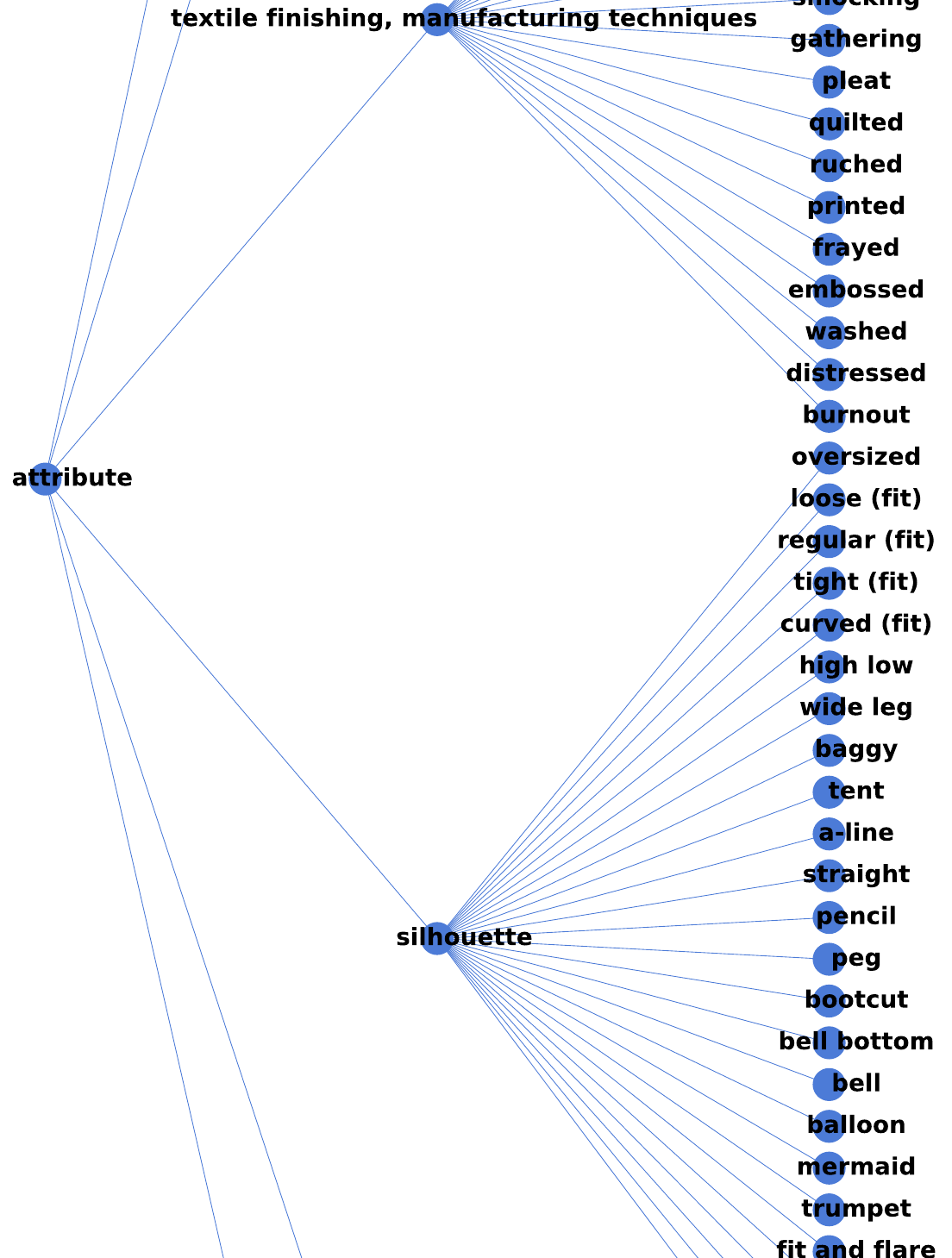}}
\caption{Apparel categories (a) and fine-grained attributes (b) hierarchy in Fashionpedia}
\label{supfig:taxo}
\end{figure}

\begin{figure}
  \begin{center}
  \includegraphics[width=0.99\textwidth]{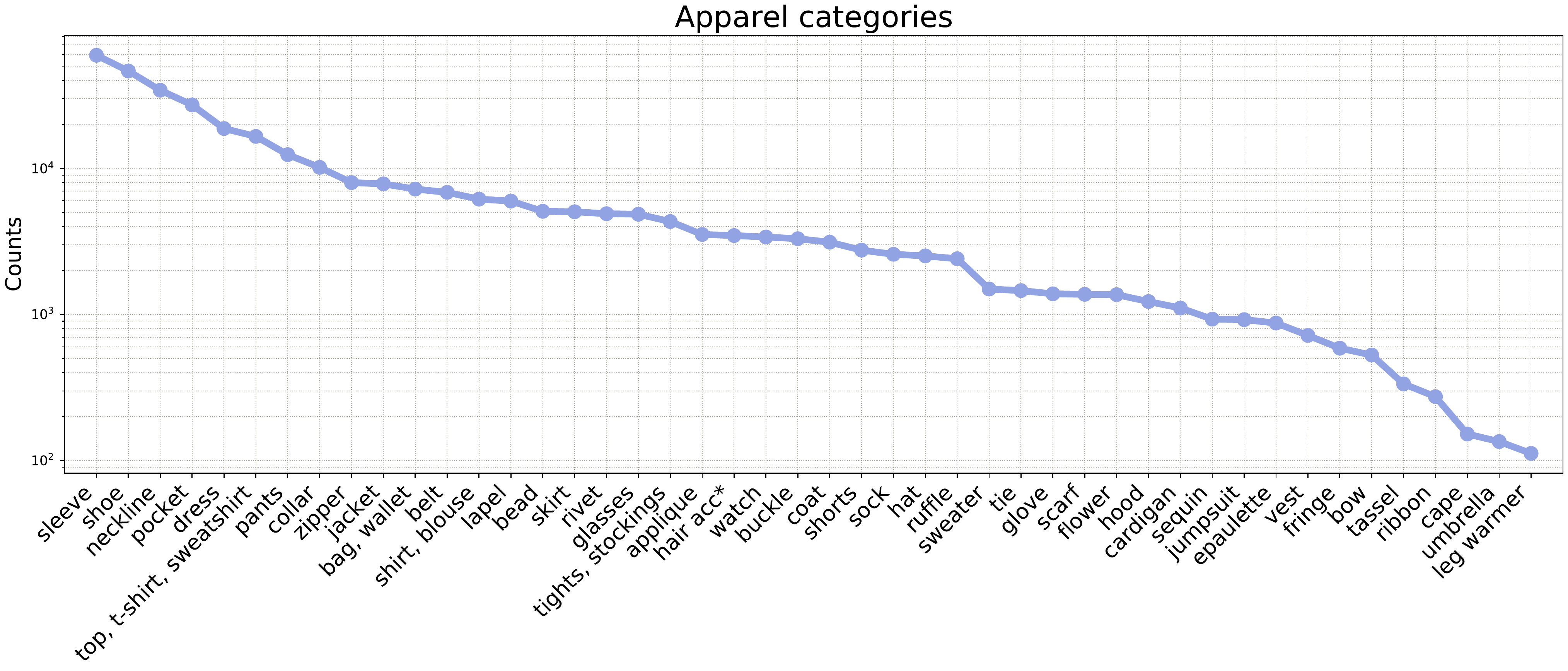}
  \end{center}
  \caption{Mask counts per apparel categories in training data. ``head acc\*'' is short for ``headband, head covering, hair accessory''}
  \label{suppfig:cat}
\end{figure}

\begin{figure}
\centering
\includegraphics[width=0.9\columnwidth]{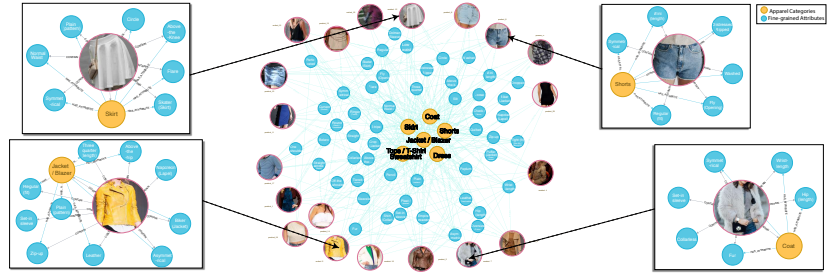}
\caption{Fashionpedia Knowledge Graph: we present a subset of the Fashionpedia knowledge graph by aggregating 20 annotated products.
The knowledge graph can be used as a tool for generating structural information
}
\label{suppfig:kg}
\end{figure}

Fig.~\ref{supfig:taxo} presents our Fashionpedia ontology in detail.
Fig.~\ref{suppfig:cat} and~\ref{suppfig:att-counts} displays the training data mask counts per category and per attributes.
Utilizing the proposed ontology and the image dataset, 
a large-scale fashion knowledge graph can be constructed to represent the fashion world in the product level.
Fig.~\ref{suppfig:kg} illustrates a subset of the Fashionpedia knowledge graph.

\begin{figure}[!b]
\centering
\subfigure[]{
    \includegraphics[width=0.99\textwidth]{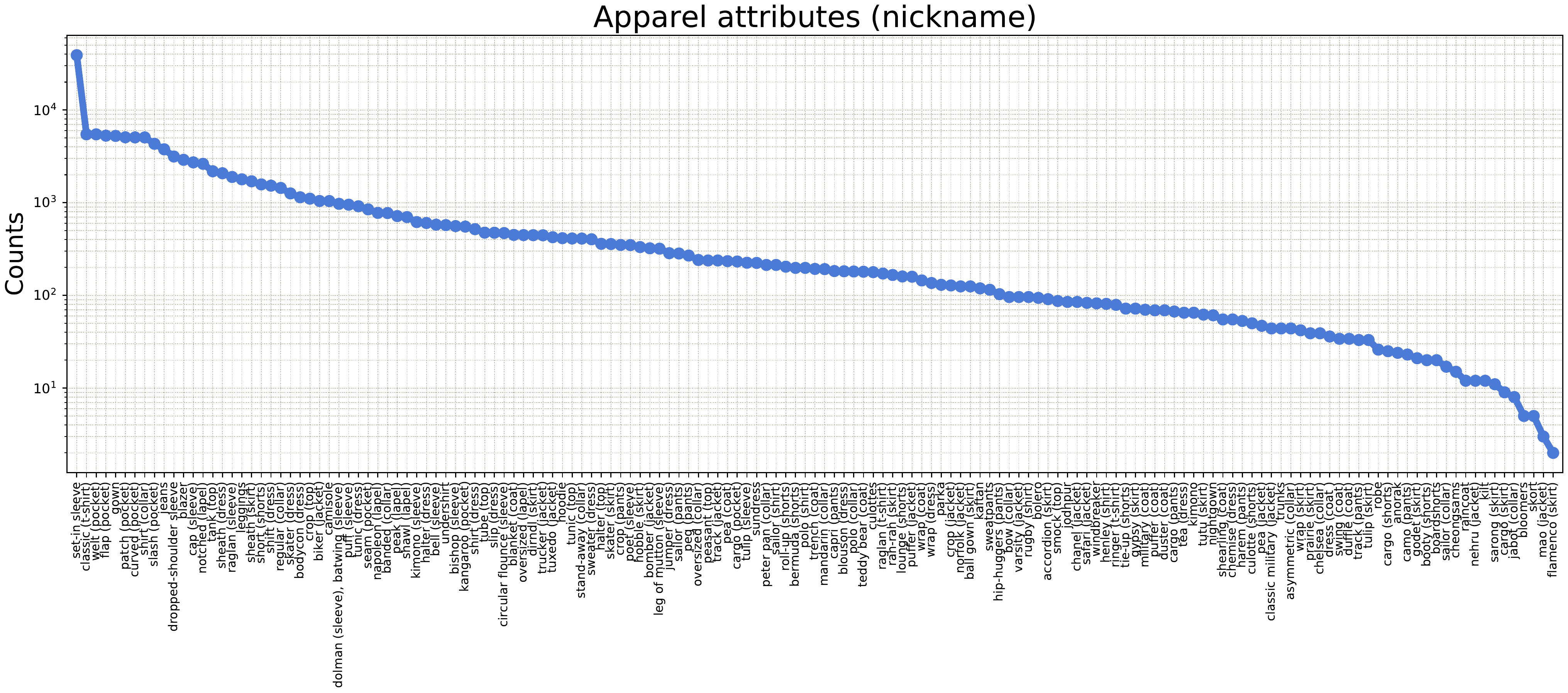}
    \label{suppfig:nickname}
}
\caption{Mask counts per attributes in training data, grouped by super categories. Best viewed digitally}
\end{figure}%

\begin{figure}\ContinuedFloat
\subfigure[]{
    \includegraphics[width=0.46\textwidth]{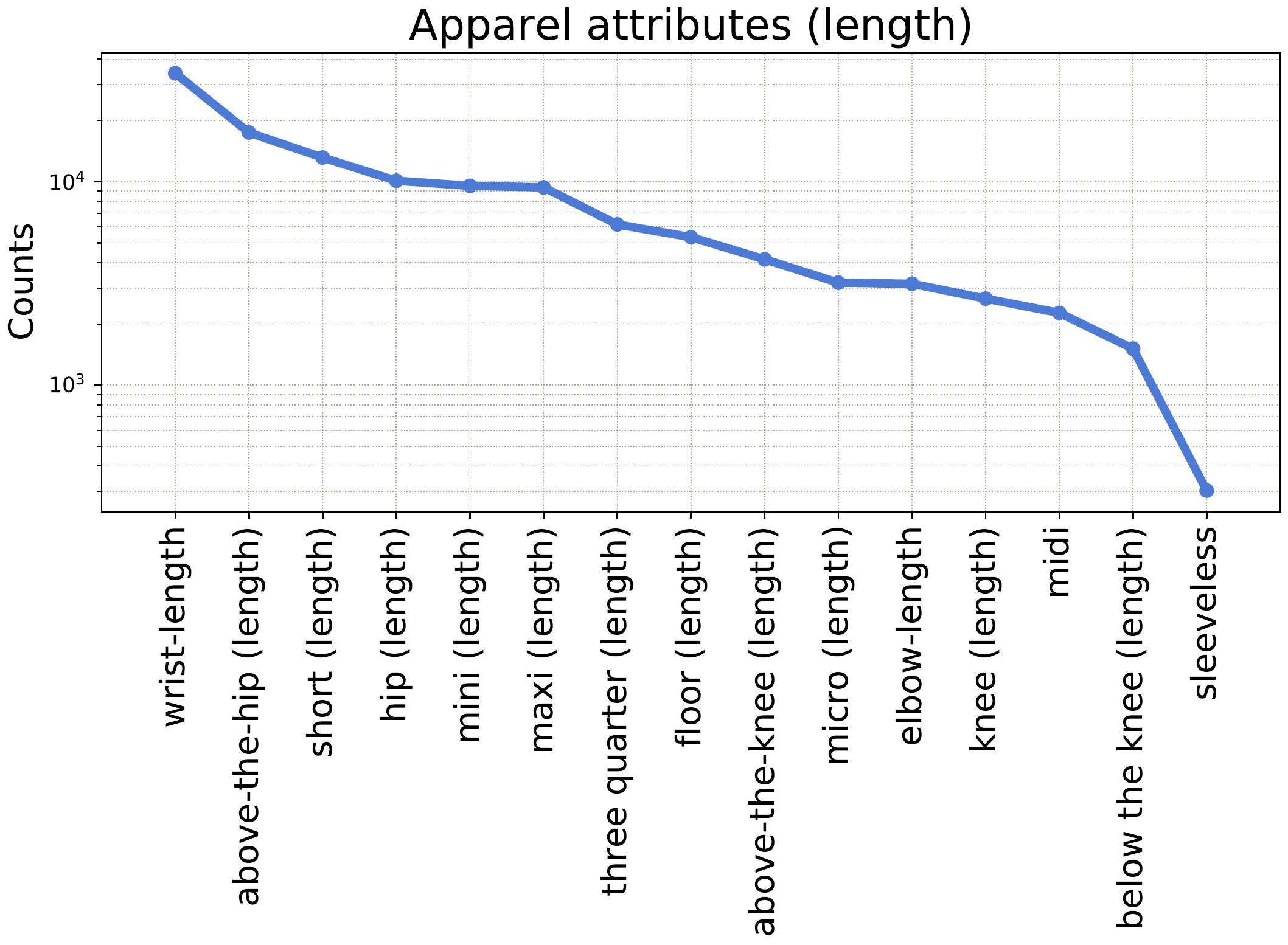}
    \label{suppfig:length}
}
\hfill
\subfigure[]{
    \includegraphics[width=0.46\textwidth]{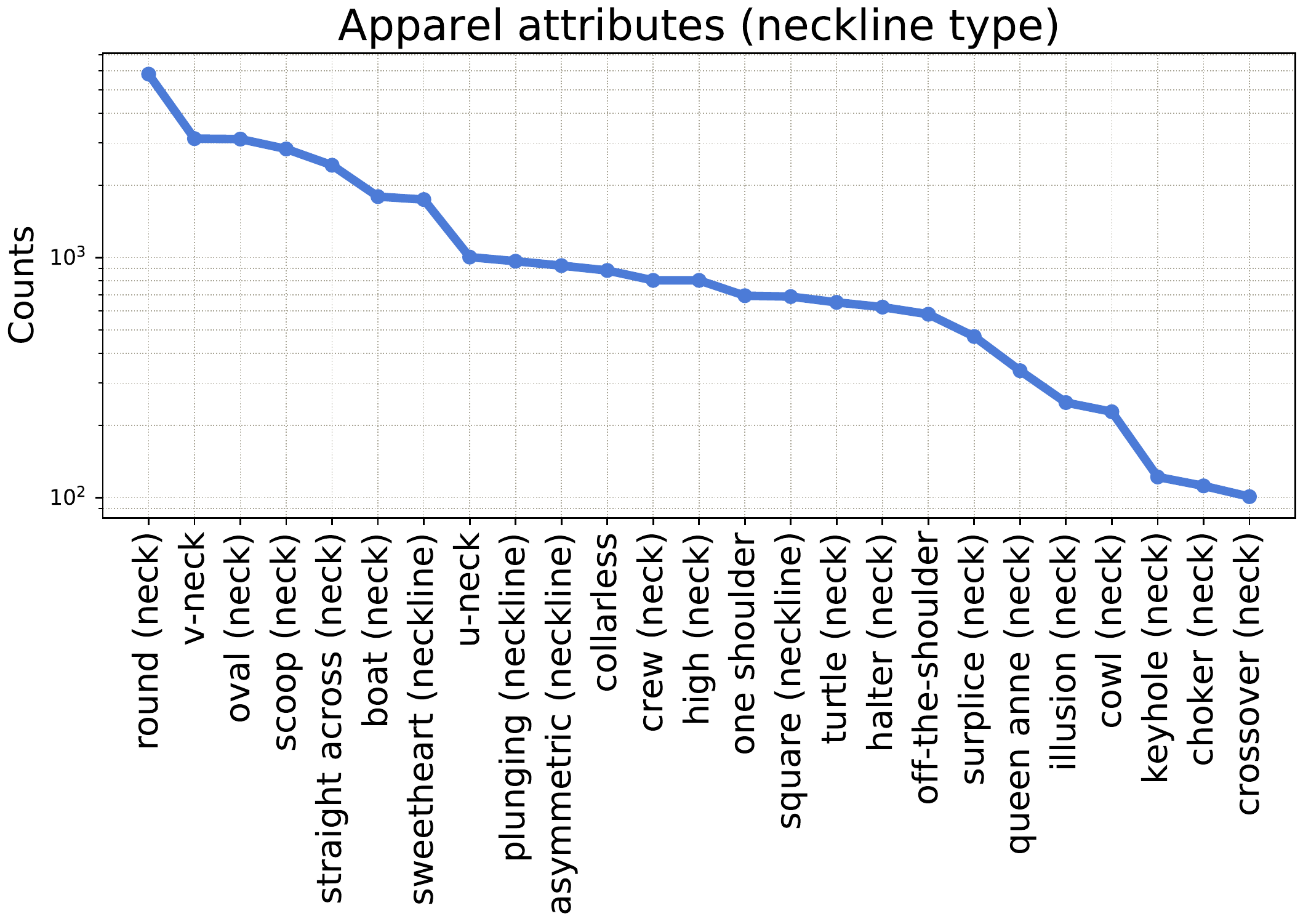}
    \label{suppfig:neck}
}
\subfigure[]{
    \includegraphics[width=0.46\textwidth]{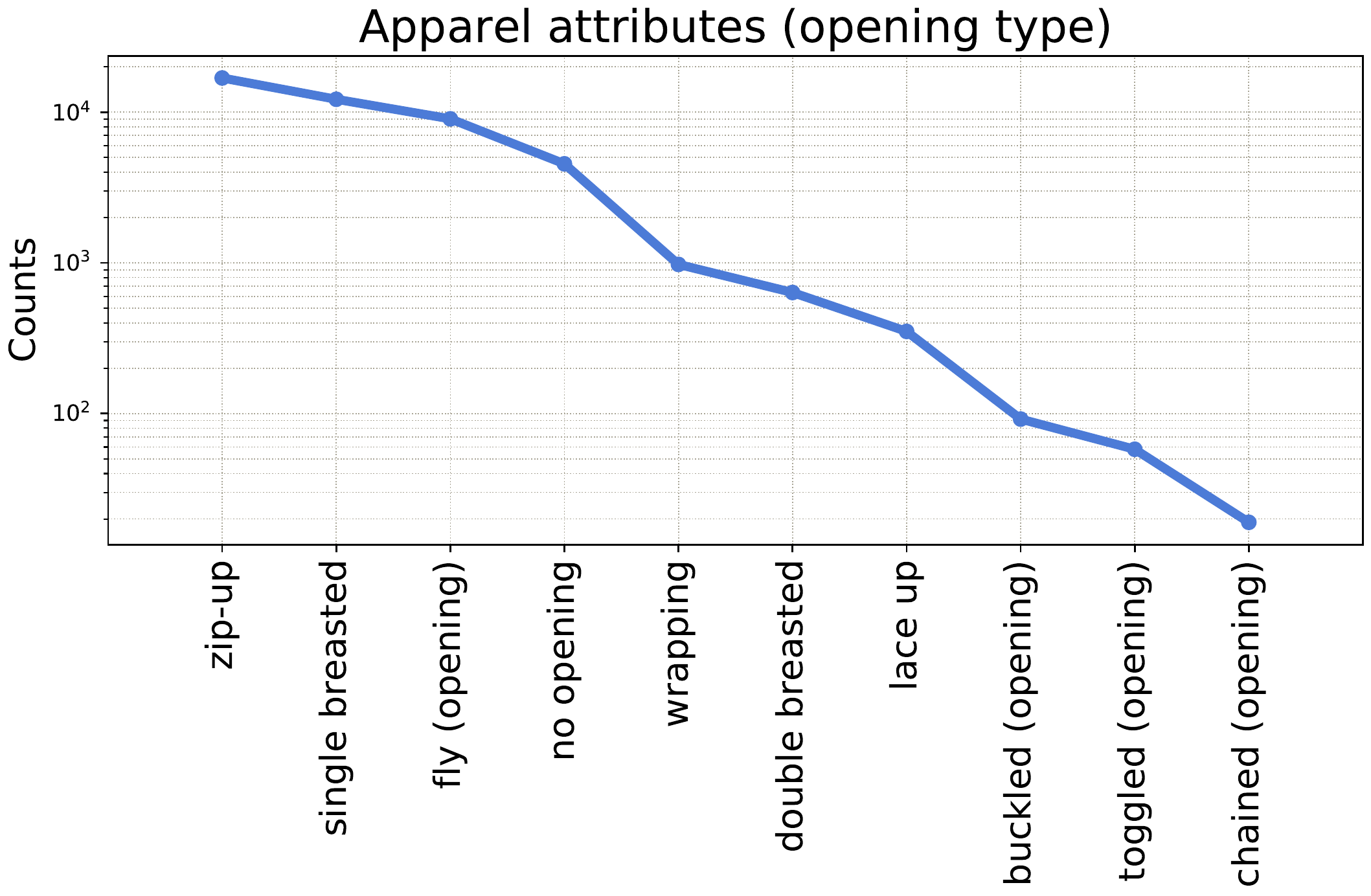}
    \label{suppfig:open}
}
\subfigure[]{
    \includegraphics[width=0.46\textwidth]{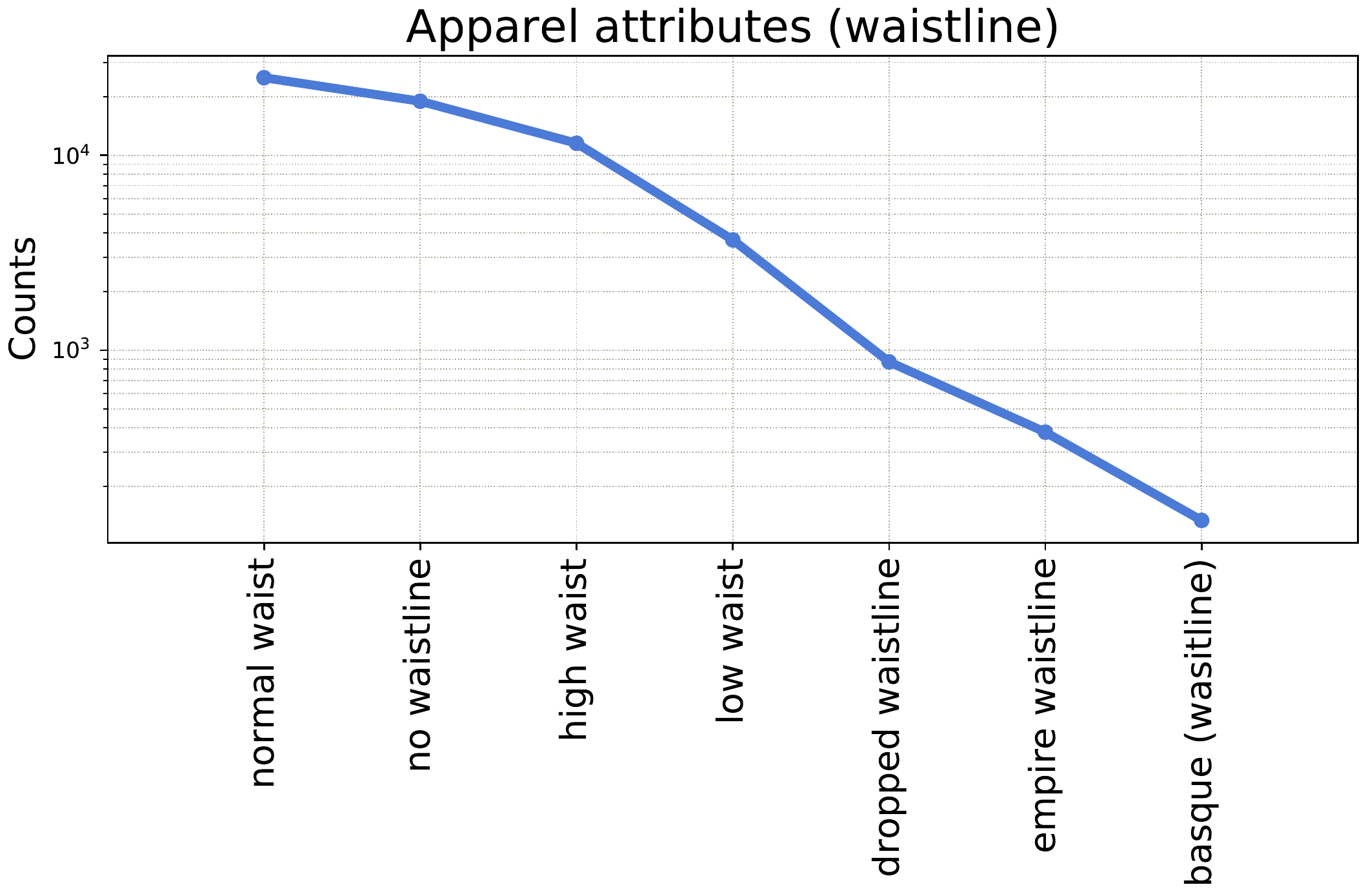}
    \label{suppfig:waist}
}
\subfigure[]{
    \includegraphics[width=0.46\textwidth]{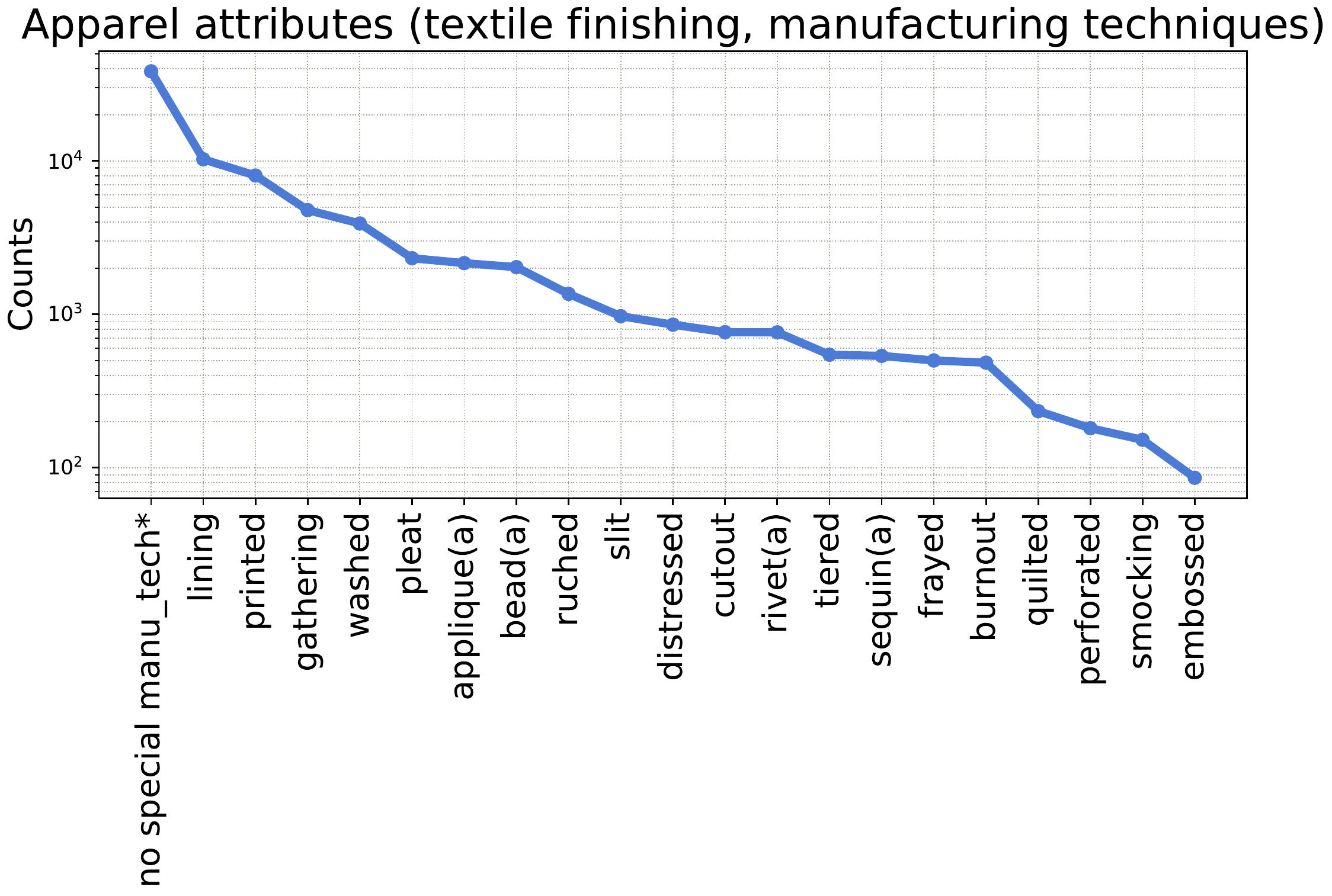}
    \label{suppfig:finish}
}
\subfigure[]{
    \includegraphics[width=0.46\textwidth]{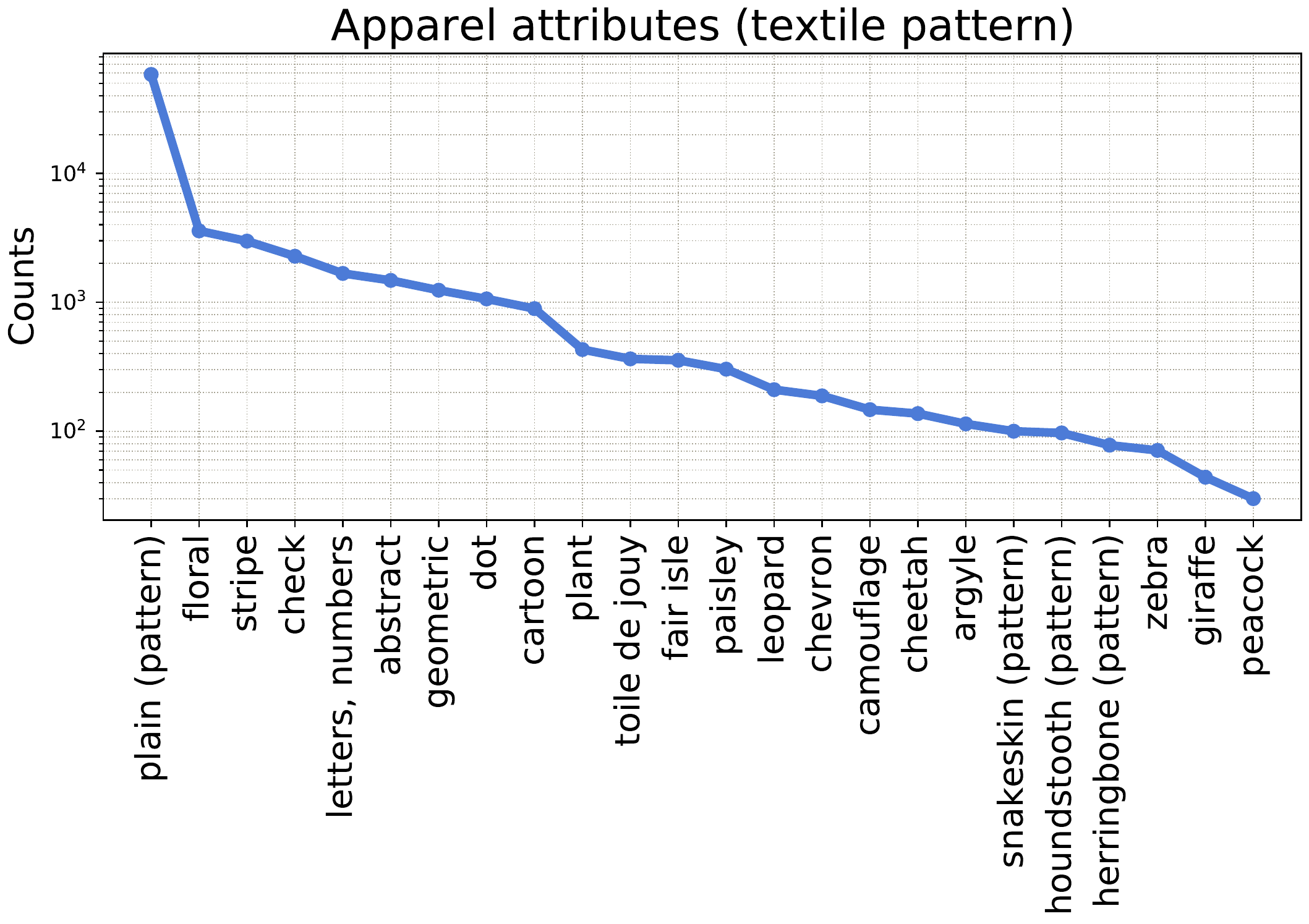}
    \label{suppfig:pattern}
}
\subfigure[]{
    \includegraphics[width=0.46\textwidth]{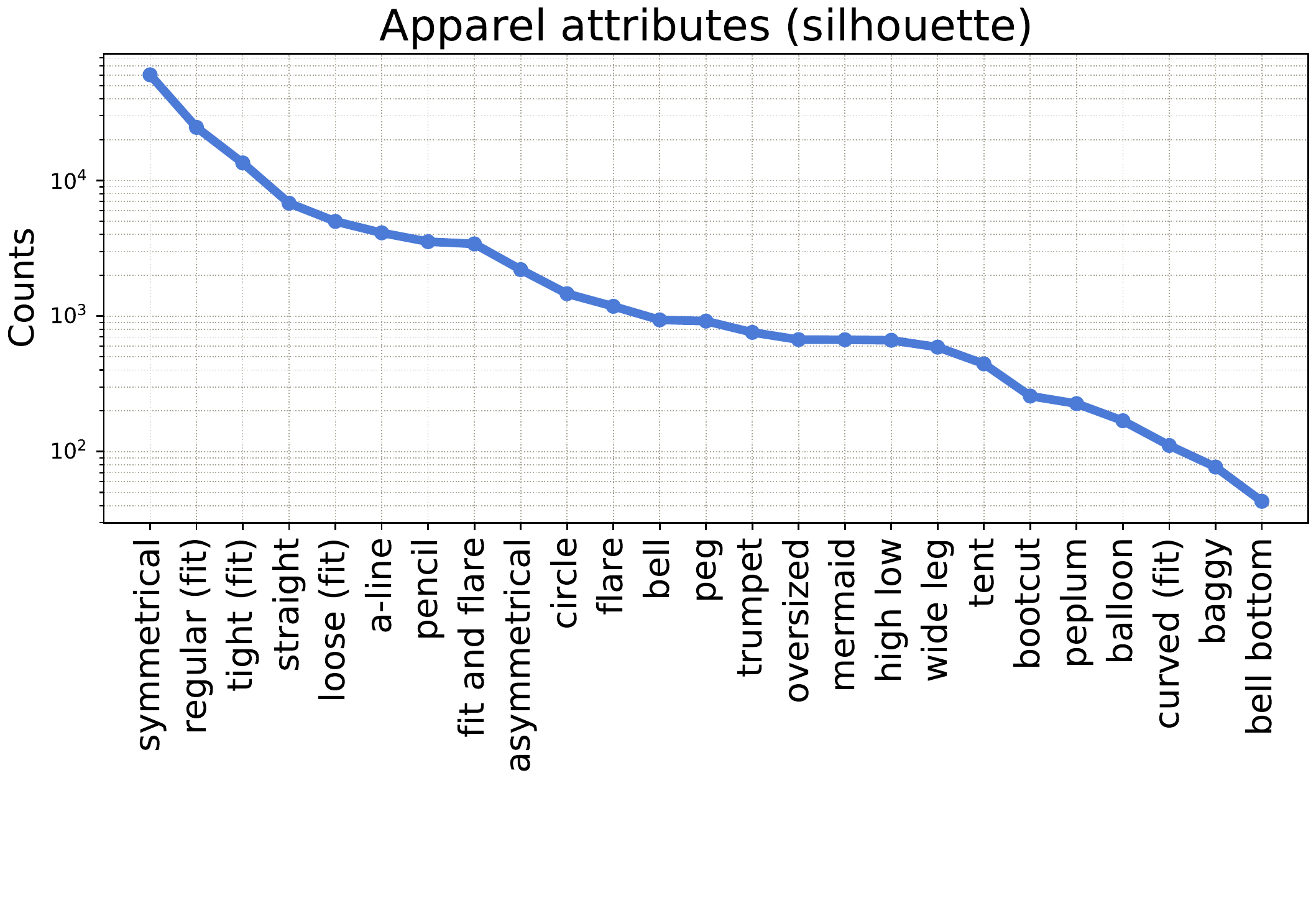}, 
    \label{suppfig:silhou}
}
\subfigure[]{
    \includegraphics[width=0.46\textwidth]{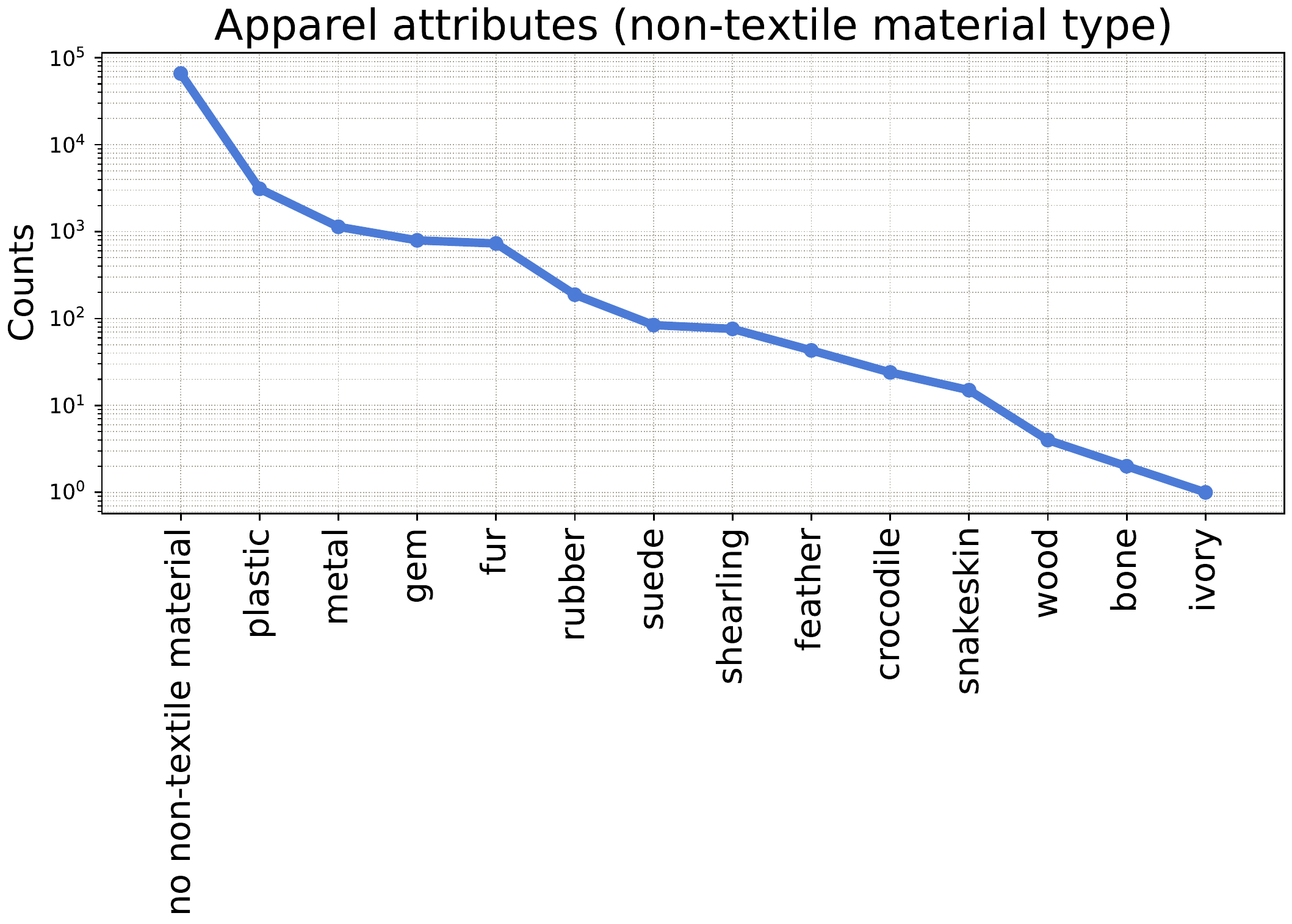}
    \label{suppfig:non-tex}
}
\caption{Mask counts per attributes in training data, grouped by super categories (cont.). Best viewed digitally}
\label{suppfig:att-counts}
\end{figure}

\textbf{Apparel graphs.}
Integrating the main garments, garment parts, attributes, and relationships presented in one outfit ensemble, we can create an apparel graph representation for each outfit in an image.
Each apparel graph is a structured representation of an outfit ensemble, containing certain types of garments.
Nodes in the graph represent the main garments, garment parts, and attributes.
Main garments and garment parts are linked to their respective attributes through different types of relationships. Figure~\ref{supfig:examples} shows more image examples with apparel graphs. 

\textbf{Fashionpedia knowledge graph.}
While apparel graphs are localized representations of certain outfit ensembles in fashion images, we can also create a single Fashionpedia knowledge graph (Figure~\ref{suppfig:kg}).
The Fashionpedia knowledge graph is the union of all apparel graphs and includes entire main garments, garment parts, attributes, and relationships in the dataset.
In this way, we are able to represent and understand fashion images in a more structured way.

We expect our Fashionpedia knowledge graph and the database to have applicability to extending the existing knowledge graph (such as WikiData~\cite{vrandevcic2014wikidata}) with novel domain-specific knowledge, improving the underlying fashion product recommendation system, enhancing search engine's results for fashion visual search, resolving ambiguous fashion-related words for text search, and more.

\subsection{Dataset Analysis}
\label{sec:supp_data}

\begin{table}[!b]
\small
\scriptsize
\begin{center}
\caption{Percentage of attributes in Fashionpedia broken down by super-class. ``Tex finish, manu-tech.'' is short for ``Textile finishing, Manufacturing techniques''. Summaries of ``not sure'' and ``not on the list'' during attributes annotations are also presented. It was calculated by the counts divided by the total masks with attributes. ``not sure'' is mainly due to occlusion inside the images, which cause some super-classes (such as waistline, opening type, and length) are unidentifiable in the images. The percentage of ``not on the list'' is less than $15\%$. This demonstrates the comprehensiveness of our Fashionpedia ontology}
\label{tab:supp_attanno}
\resizebox{0.8\textwidth}{!}{%
\begin{tabular}{ l | r | r | r }
\Xhline{1.0pt}
\textbf{Super-category} & \textbf{class count}  & \textbf{\textit{not sure}} & \textbf{\textit{not on the list}}
\\
\Xhline{1.0pt}
Length       &15    &12.79\% &0.01\% \\
Nickname     &153   &9.15 \% &12.76\%  \\
Opening Type &10   &32.69\% &3.90\%  \\
Silhouettes  &25  &2.90\%  &0.27\% \\
Tex finish, manu-tech 
             &21 &4.47\%  &1.34\% \\
Textile Pattern &24  &2.18\%  &5.30\% \\
None-Textile Type &14  &4.90\% &4.07\% \\
Neckline    &25   &9.57\%   &3.38\% \\
Waistline    &7   &30.46\% &0.17\% 
\\
\Xhline{1.0pt}
\end{tabular} %
}
\end{center}
\end{table}

\begin{figure}
  \begin{center}
  \includegraphics[scale=0.4]{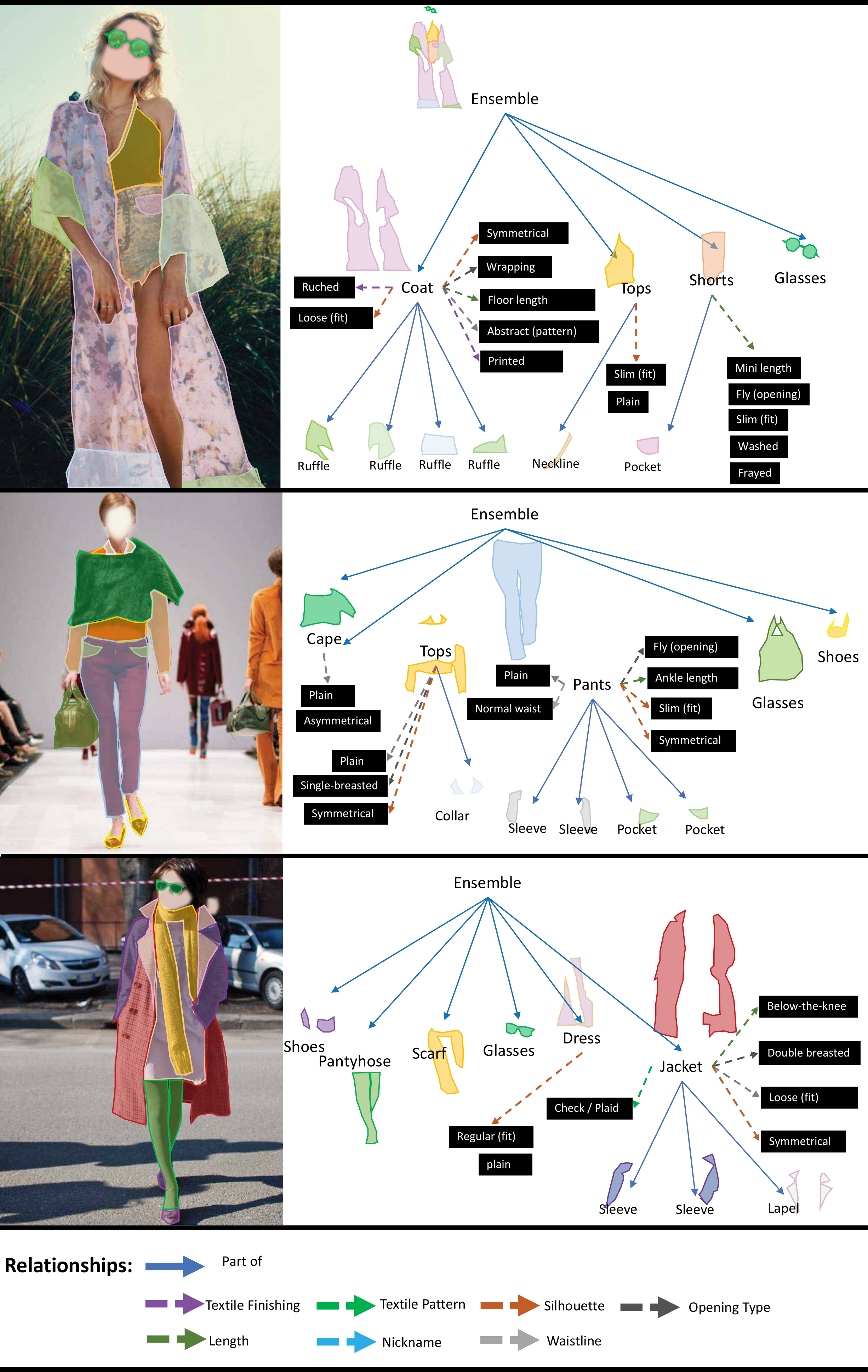}  
  \end{center}
  \caption{Example images and annotations from our dataset: the images are annotated with both instance segmentation masks and fine-grained attributes (black boxes)}
  \label{supfig:examples}
\end{figure}

\begin{figure}
  \begin{center}
  \includegraphics[width=0.9\textwidth, clip=true, trim = 0 50mm 0  50mm]{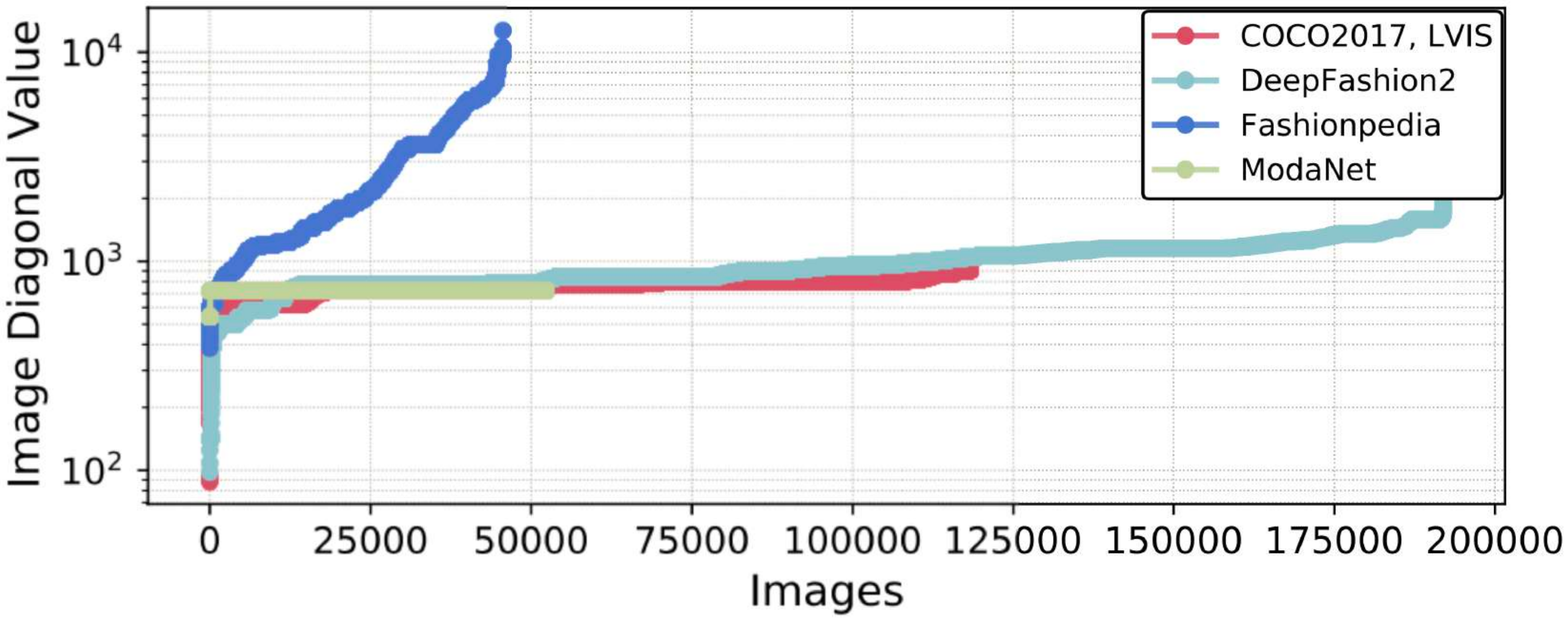}
  \end{center}
  \caption{Image size comparison among Fashionpedia, ModeNet, DeepFashion2, and COCO2017, LVIS. Only training images are shown. The Fashionpedia images has the most diverse resolutions. Note that COCO2017 and LVIS have higher resolution images for annotation. The distribution presented here are the publicly available photos}
  \label{supfig:img_comp}
\end{figure}

\begin{figure}
  \begin{center}
  \includegraphics[width=0.9\textwidth]{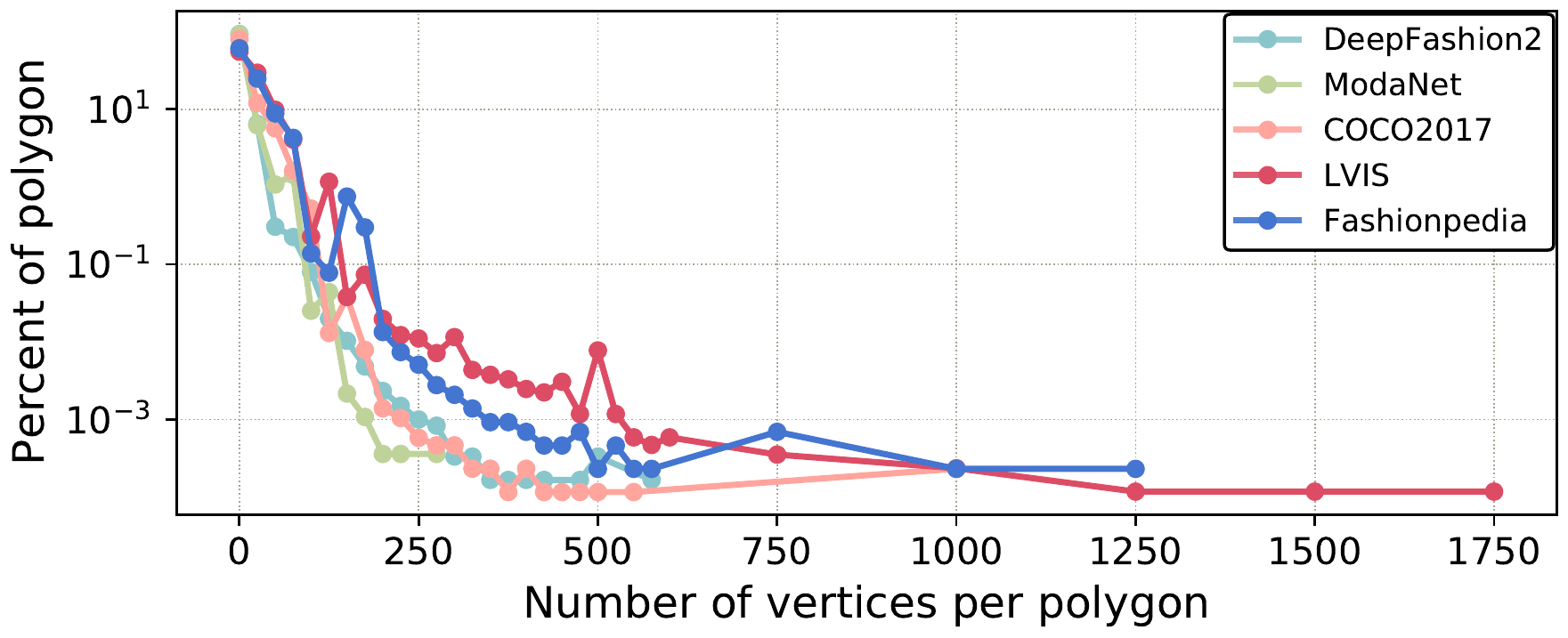}
  \end{center}
  \caption{The number of vertices per polygon. This represents the quality of masks and the efforts of annotators. Values in the x-axis were discretized for better visual effect. Y-axis is on log scale. Fashionpedia has the second widest range, next to LVIS}
  \label{supfig:ver_per_poly}
\end{figure}

Fig.~\ref{supfig:examples} shows more annotation examples, represented in the exploded views of annotation diagrams.
Table~\ref{tab:supp_attanno} displays the details about ``not sure'' and ``not on the list'' results during attribute annotation process.
We present the result per super-categories of attributes.
Label ``not sure'' means the expert annotator is uncertain about the choice given the segmentation mask.
``Not on the list'' means the annotator is certain that the given mask presents another attributes that is not presented in the Fashionpedia ontology.
Other than ``nicknames'' (which is the specific name for a certain apparel category), less than 6$\%$ of the total masks account for the ``not on the list'' category.

Fig.~\ref{supfig:img_comp} and~\ref{supfig:ver_per_poly} also compare Fashionpedia and other images datasets in terms of image size and vertices per polygons. 

We compare image resolutions between Fashionpedia and four other segmentation datasets (COCO and LVIS share the same images). 
Fig.~\ref{supfig:img_comp} shows that images in Fashionpedia have the most diverse image width and height. 
While ModaNet has the most consistent resolutions of images.
Note that high resolution images will burden the data loading process of training. With that in mind, we will release our dataset in both the resized and the original versions.

We also report the distribution of number of vertices per polygons in Fig.~\ref{supfig:ver_per_poly}.
This measures the annotation effort in mask annotation. Fashionpedia has the second-widest range, next to LVIS.

\subsection{Fashionpedia dataset creation details}
\label{sec:supp_anno}

\textbf{Image collection.}
To avoid photo bias, all the images are randomly collected from Flickr and free license stock photo websites (Unsplash, Burst by Shopify, Free stocks, Kaboompics, and Pexels). 
The collected images are further verified manually by two fashion experts.
Specifically, they check the scenes’ diversity and make sure the clothing items were visible and annotatable in the images.
The estimated image type breakdown is listed as follows: street style images ($30\%$ of the full dataset), celebrity events images ($30\%$), runway show images ($30\%$), and online shopping product images ($10\%$). For gender distribution, the gender in $80\%$ of images are female, and $20\%$ of images are male.

\begin{figure*}
  \begin{center}
  \includegraphics[width=0.9\textwidth]{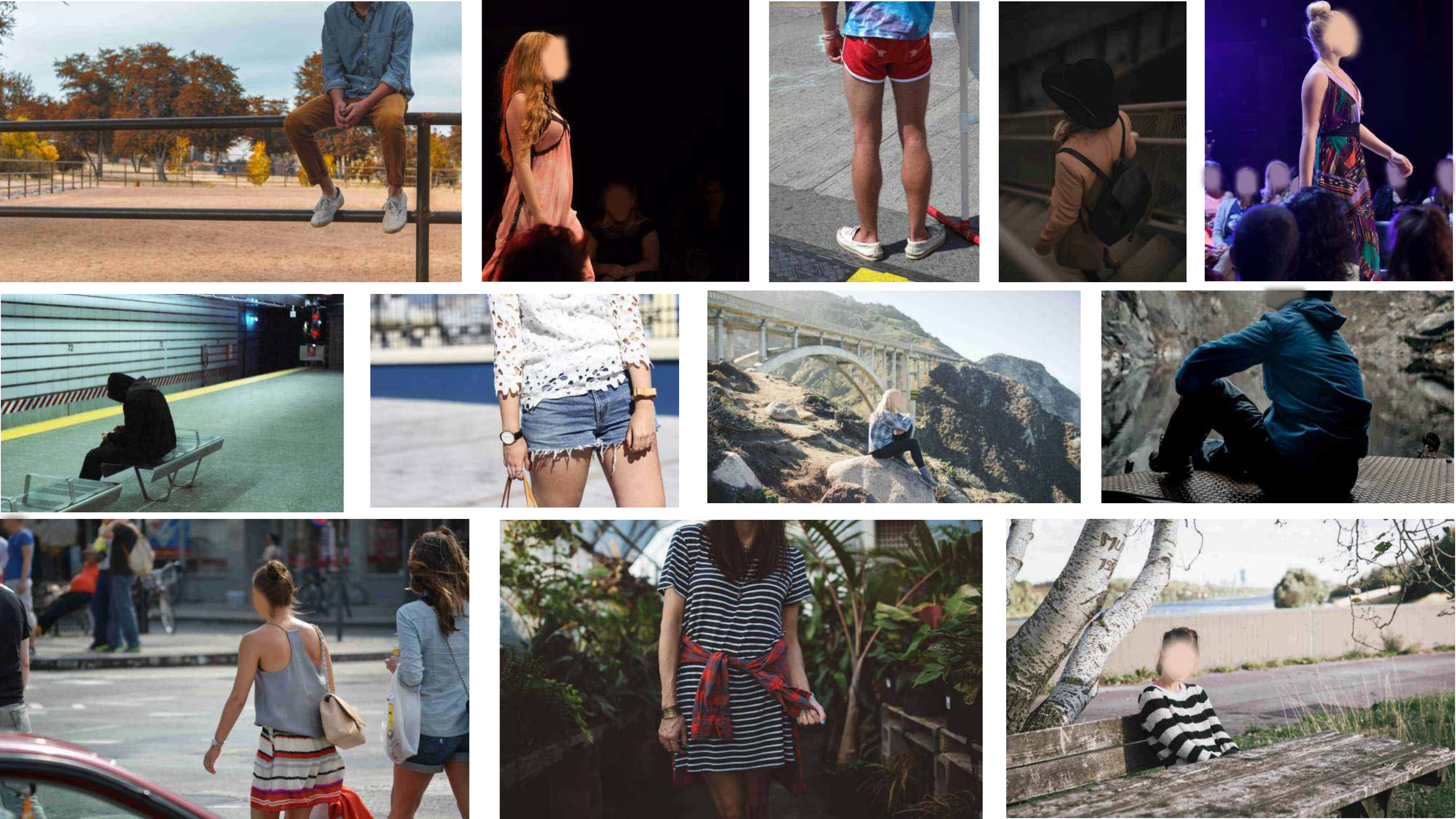}
  \end{center}
  \caption{Example of different type (people position/gesture, full/half shot, occlusion, scenes, garment types, etc.) of images in Fashionpedia dataset}
  \label{supfig:biased_image}
\end{figure*}

We did aim to address the issue of photographic bias in the image collection process.
Our dataset includes the images that are not centered, not full shot and with occlusion (see examples in Fig.~\ref{supfig:biased_image}).
Furthermore, our focus is to identify clothing items, not to identify people during the image collection process.

\begin{figure*}[t]
  \begin{center}
  \includegraphics[width=0.9\textwidth]{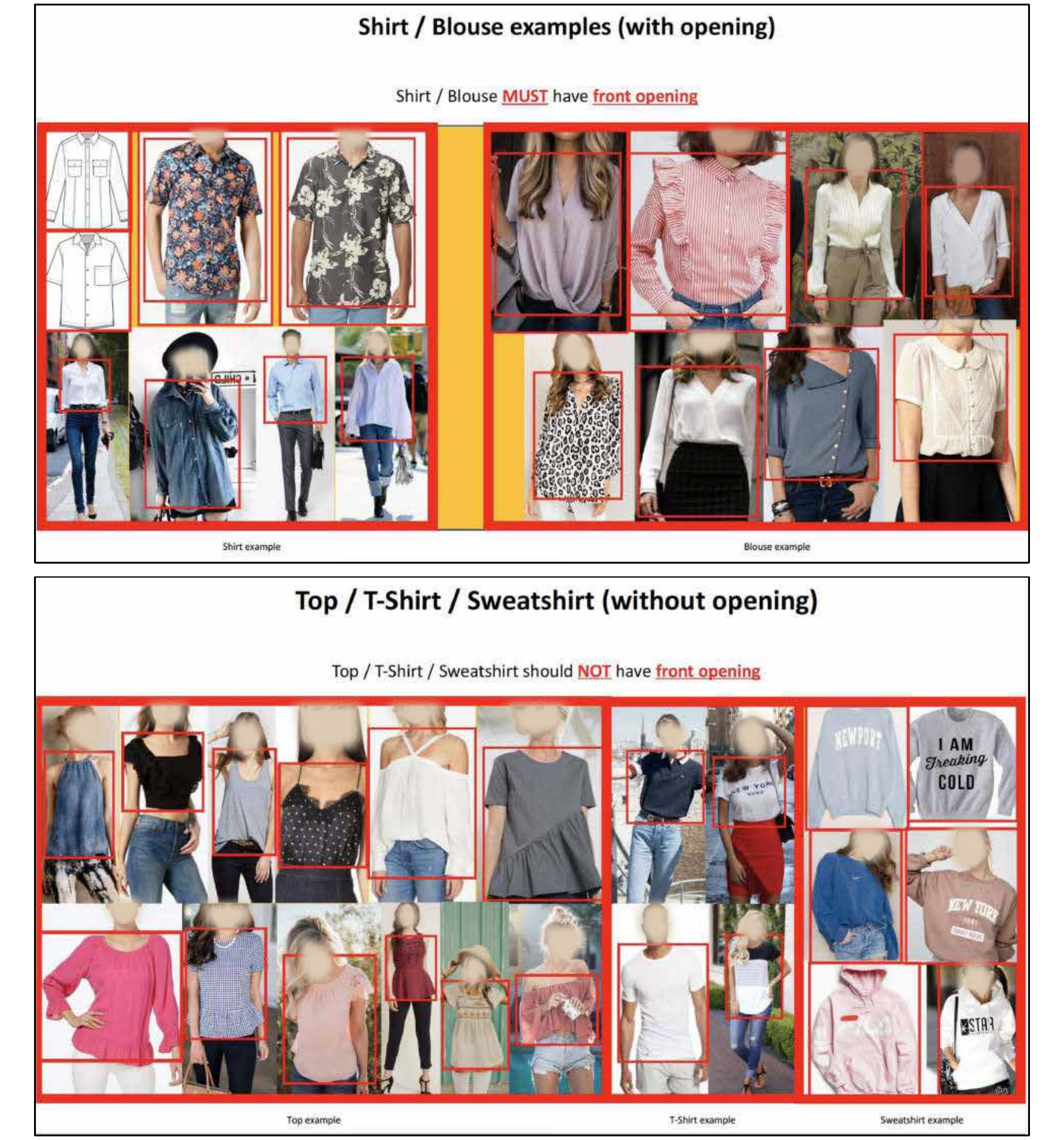}
  \end{center}
  \caption{Annotation tutorial example for shirt and top}
  \label{supfig:shirt_top}
\end{figure*}

\begin{figure*}[t]
  \begin{center}
  \includegraphics[width=0.9\textwidth]{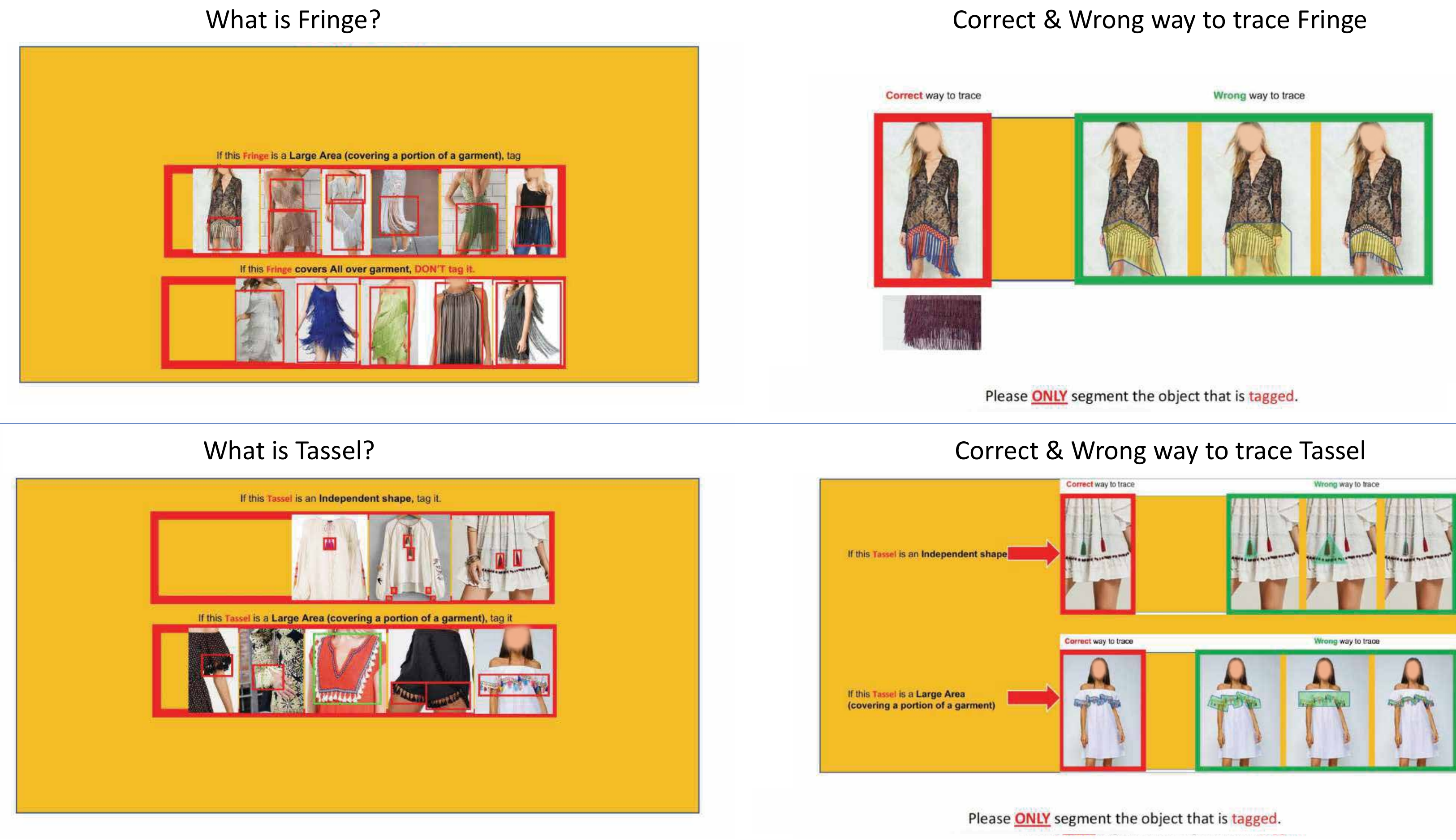}
  \end{center}
  \caption{Annotation tutorial for fringe and tassel}
  \label{supfig:fring_tassel}
\end{figure*}

\begin{figure*}[t]
  \begin{center}
  \includegraphics[width=0.9\textwidth]{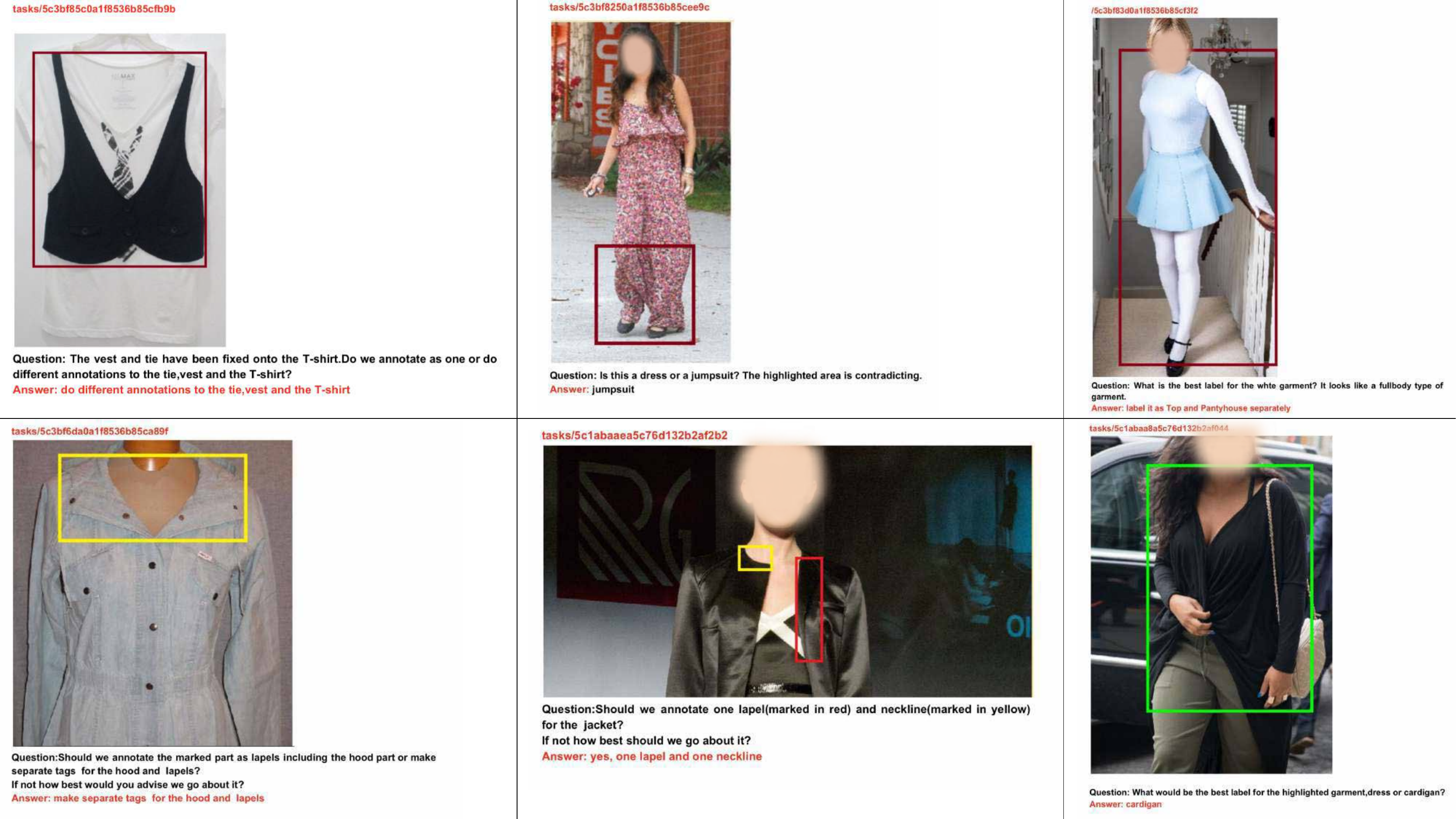}
  \end{center}
  \caption{Example of debatable fashion items in Fashionpedia dataset. The questions are asked by the crowdworkers. The answers are provided by two fashion experts}
  \label{supfig:debatalbe_item}
\end{figure*}

\textbf{Crowd workers and 10-day training for mask annotation.}
In the spirit of sharing the same apparel vocabulary for all the annotators, we prepared a detailed tutorial (with text descriptions and image examples) for each category and attributes in the Fashionpedia ontology (see Fig.~\ref{supfig:shirt_top} for an example).
Before the official annotation process, we spent 10 days on training the 28 crowd workers for the following three main reasons.

First, some apparel categories are commonly referred as other names in general. For example, ``top'' is a general term for ``shirt'', ``sweater'', ``t-shirt'', ``sweatshirt''. 
Some annotators can mistakenly annotate a ``shirt'' as a ``top''.
We need to train these workers so they have the same understanding of the proposed Fashionpedia ontology.
Utilizing the prepared tutorials (see Fig.~\ref{supfig:shirt_top} for an example), we trained and educated annotators on how to distinguish among different apparel categories. 

Second, there are fine-grained differences among apparel categories.
For example, we observed that some workers initially had difficulty in understanding the difference among different garment parts, such as ‘tassel’ and ‘fringe’.
To help them understand the difference of these objects, we ask them to practice and identify more sample images before the annotation process.
Fig.~\ref{supfig:fring_tassel} shows our tutorials for these two categories. We specifically shows some correct and wrong examples of annotations.

Third, we ask for the quality of annotations. In particular, we ask the annotators to follow the contours of garments in the images as closely as possible. The polygon annotation process is monitored and verified for a few days before the workers started the actual annotation process.

\textbf{Quality control of debatable apparel categories.} 
During the annotation process, we allow annotators to ask questions about the uncertain categories. Two fashion experts monitored the annotation process by answering questions, checking the annotation quality, and providing weekly feedback to annotators.

Instead of asking annotators to rate their confidence level of each segmentation mask, we asked them to send back all the uncertain masks to us during the annotation. The same two fashion experts made the final judgement and gave the feedback to the workers on these debatable or unsure fashion categories. Some examples of debatable or fuzzy fashion items that we have documented can be found in Figure~\ref{supfig:debatalbe_item}.

\subsection{Discussion}
\label{sec:supp_mis}

\textbf{Does this dataset include the images or labels of previous datasets?}
We only include the previous datasets for comparison. Our dataset doesn’t intentionally use any images or labels from previous datasets. All the images and labels from Fashionpedia are newly collected and annotated.

\textbf{Who were the fashion experts annotating localized attributes in Fashionpedia dataset?} The fashion experts are the 15 fashion graduate students that we recruited from one of the top fashion design institutes. For double-blind policy, we cannot mention the name of the university. But we will release the name of this university and the collaborators from this university in the final version of this paper.

\begin{figure*}[t]
  \begin{center}
  \includegraphics[width=0.9\textwidth]{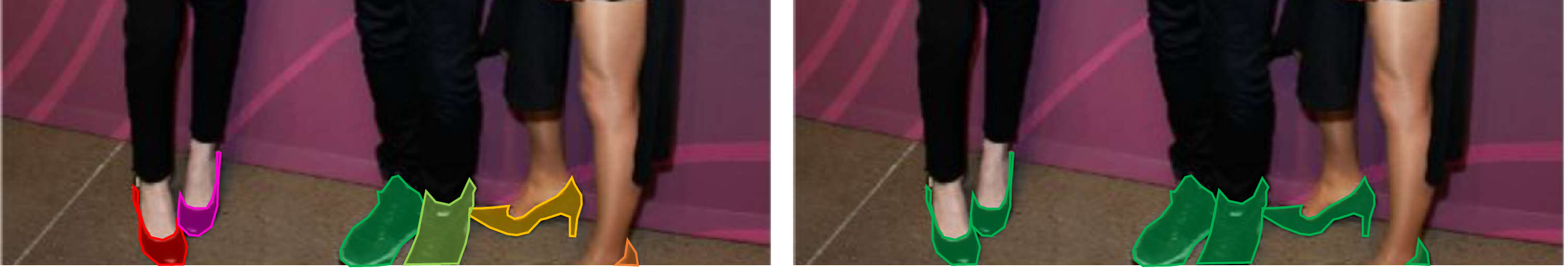}
  \end{center}
  \caption{Instance segmentation (left) and semantic segmentation (right)}
  \label{supfig:semantic_segmentationn}
\end{figure*}

\begin{figure*}[!b]
  \begin{center}
  \includegraphics[width=0.8\textwidth]{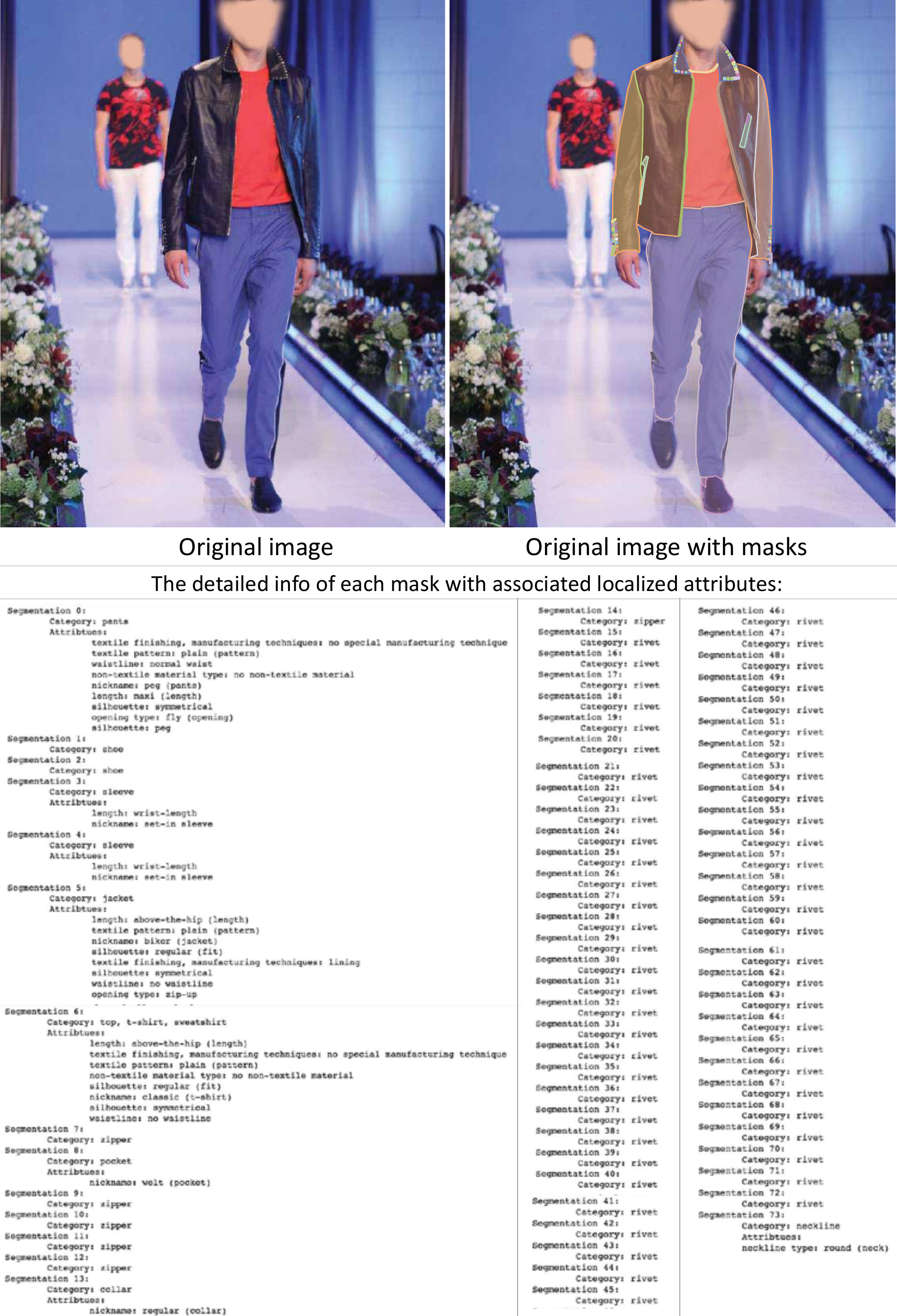}
  \end{center}
  \caption{The image with 74 masks in Fashionpedia dataset}
  \label{supfig:mask74}
\end{figure*}

\textbf{Instance segmentation v.s. semantic segmentation.} We didn't conduct semantic segmentation experiments on our dataset for the following two reasons: 1) Although semantic segmentation is a useful task, we believe instance segmentation is more meaningful for fashion images. 
For example, if we need to distinguish the different shoe style of a fashion image containing 3 pair of different shoes, instance segmentation (Figure~\ref{supfig:semantic_segmentationn}(a)) can help us distinguish each shoe separately. However, semantic segmentation (Figure~\ref{supfig:semantic_segmentationn}(b)) will mix all the shoe instances together. 2) Semantic segmentation is the sub-problem of instance segmentation. If we merge the same detected object class from our instance segmentation experimental result, it yields the results for semantic segmentation.

\textbf{Which image has the most annotated masks?}
In Fashionpedia dataset, the maximum number of segmentation masks in an image is $74$ (Fig.~\ref{supfig:mask74}). We find that most of the masks are belonging to ``rivets'' (garment parts).

\textbf{What's the difference between Fashionpedia and other fine-grained datasets like CUB-200?} We propose to localize fine-grained attributes within segmentation masks of images. This is a novel task with real-world application to the best of our knowledge. The differences between Fashionpedia and CUB are as follows: 1) CUB uses keypoints as annotation to indicate different locations on birds, while Fashionpedia has segmentation masks of garments, garment parts, and accessories; 2) Fashionpedia attributes are associated with garment or garment part instances in images, whereas CUB provides global attributes, not associated with any specific keypoints.

\subsubsection{iMat-Fashion Kaggle challenges}
\label{sec:sup_kaggle}

To advance state-of-the-art of visual analysis of clothing, we hosted two kaggle challenges (imaterialist-fashion) on Kaggle in 2019~\footnote{\url{https://www.kaggle.com/c/imaterialist-fashion-2019-FGVC6}} and 2020~\footnote{\url{https://www.kaggle.com/c/imaterialist-fashion-2020-fgvc7}} respectively.

\newpage
%
%
\bibliographystyle{splncs04}
\bibliography{egbib}

\end{document}